\documentclass[runningheads]{llncs}

\usepackage{graphicx}
\usepackage[utf8]{inputenc} 
\usepackage{booktabs}       
\usepackage{amsfonts}       
\usepackage[]{todonotes}
\usepackage{multirow}
\usepackage{amsmath}
\usepackage{array}
\usepackage[export]{adjustbox}
\usepackage{hyperref}

\begin{document}
%
\title{Modeling continuous-time stochastic processes using $\mathcal{N}$-Curve mixtures}

\titlerunning{Modeling stochastic processes using $\mathcal{N}$-Curve mixtures}

%
%
\author{Ronny Hug\orcidID{0000-0001-6104-710X} \and
	Wolfgang Hübner\orcidID{0000-0001-5634-6324} \and
	Michael Arens\orcidID{0000-0002-7857-0332}}

\authorrunning{R. Hug et al.}
%
\institute{Fraunhofer Institute of Optronics, System Technologies, and Image Exploitation (IOSB)\thanks{Fraunhofer IOSB is a member of the Fraunhofer Center for Machine Learning.}, Gutleuthausstr. 1, 76275 Ettlingen, Germany
	\email{\{ronny.hug|wolfgang.huebner|michael.arens\}@iosb.fraunhofer.de}}

\maketitle              
\begin{abstract} 
	Representations of sequential data are commonly based on the assumption that observed sequences are realizations of an unknown underlying stochastic process, where the learning problem includes determination of the model parameters.
	In this context the model must be able to capture the multi-modal nature of the data, without blurring between modes.
	This property is essential for applications like trajectory prediction or human motion modeling.	
	Towards this end, a neural network model for continuous-time stochastic processes usable for sequence prediction is proposed.
	The model is based on Mixture Density Networks using Bézier curves with Gaussian random variables as control points (abbrev.: $\mathcal{N}$-Curves).
	Key advantages of the model include the ability of generating smooth multi-mode predictions in a single inference step which reduces the need for Monte Carlo simulation, as required in many multi-step prediction models, based on state-of-the-art neural networks.
	Essential properties of the proposed approach are illustrated by several toy examples and the task of multi-step sequence prediction.
	Further, the model performance is evaluated on two real world use-cases, i.e. human trajectory prediction and human motion modeling, outperforming different state-of-the-art models.
	
\end{abstract}
\section{Introduction}
Probabilistic models of sequential data have a broad range of applications related to representation learning, sequence generation and prediction.
Autoregressive models, like the Gaussian Process regression model or autoregressive moving average (ARMA) models, have been applied in this domain \cite{roberts2013gaussian}\cite{montgomery2015introduction}. 
More recently, neural network-based approaches have gained popularity with the advent of deep learning \cite{langkvist2014review}\cite{fawaz2019deep}.
An objective common to most approaches is to adapt the respective model to the data given for a specific task.
Following this, (deep) learning approaches, and especially sequence learning approaches, are concerned with learning a model of a presumed underlying stochastic process $\{X_t\}_{t \in T}$, where $X_t$ is a random variable and $T$ is an index set.
The model is then learned from given realizations (sample sequences).

This paper focuses on the subtask of multi-step sequence prediction, where the evolution of a probability density function for $n$ discrete time steps should be inferred, given $m$ observations (subsequent values) of a sample sequence.
Multi-step prediction approaches are applied to a variety of applications, like stock market prediction \cite{atsalakis2009surveying}\cite{soni2011applications}, trajectory prediction \cite{rudenko2019human}\cite{lefevre2014survey} or human motion modeling \cite{wang2006gaussian}\cite{taylor2007modeling}\cite{martinez2017human}.
Such multi-step predictions can be relevant in applications that require ahead-of-time planning, like robot navigation or autonomous driving \cite{lefevre2014survey}.
Common approaches for tackling sequence prediction include for example specialized stochastic process models and (stochastic) neural network models \cite{rudenko2019human}\cite{lefevre2014survey}\cite{soni2011applications}\cite{bishop1995neural}\cite{bishop2006pattern}.
While being able to generate (multi-modal) sequence predictions, a drawback shared by many of these models is a computationally intensive inference scheme for generating predictions.

Towards this end, a model of a continuous-time stochastic process is proposed, which extends on Mixture Density Networks and can be learned from time-discrete realizations.
For an initial proof of concept, the index set is given by $t \in [0, 1]$, thus focuses on fixed length sequences.
Set into the context of $n$-step prediction, generating multi-modal predictions should be possible without the need of extensive Monte Carlo simulation, thus performing $n$-step inference with minimum overhead in a single step.
To achieve this, conditional sampling steps are moved into the training phase and the modes of the modeled stochastic process are described in terms of probabilistic, parametric curves.  
These curves, termed $\mathcal{N}$-Curves, are based on Bézier curves, replacing deterministic control points by Gaussian control points, and constrain inference in order to generate sufficiently smooth predictions.
As a consequence, every point on the $\mathcal{N}$-Curve is a Gaussian random variable, thus resembling the model of a continuous-time stochastic process.
For generating multi-modal predictions (multiple sequences), a stochastic process consisting of Gaussian mixture random variables can be defined by combining multiple $\mathcal{N}$-Curves into a mixture of $\mathcal{N}$-Curves.
Further, by generating an entire curve, $n$-step prediction is performed in one single inference step using the proposed model.
Finally, by basing the approach on Bézier curves, data of arbitrary dimensionality can be modeled by choosing the control point dimensionality accordingly.

The paper is structured as follows: At first, an overview over related work is given in section \ref{sec:rw}.
Afterwards, the derivations of $\mathcal{N}$-Curve and $\mathcal{N}$-Curve mixture models are given in section \ref{sec:n-curves}, where essential properties of the model are illustrated by toy examples.
Then, a neural network-based approach for learning the parameters of an $\mathcal{N}$-Curve mixture from sequence data is presented in section \ref{sec:ncdn}.
Finally, experiments are conducted on real world data in order to showcase the models capabilities in the context of (multi-modal) $n$-step sequence prediction (section \ref{sec:experiments}).
The experimental section also provides examples to support results of the toy examples given in section \ref{sec:n-curves}.
At last, conclusions and future directions of research are given in section \ref{sec:conclusions}.

\section{Related Work}  
\label{sec:rw}
For structuring the related work section, approaches for $1$-step prediction, which are a special case of $n$-step prediction, will be presented briefly in \ref{ss:rw_single}.
Next, approaches for $n$-step prediction, which are extensions of $1$-step prediction approaches, are presented in \ref{ss:rw_multi}.
Overall, the following discussion addresses approaches that model probability distributions, or stochastic processes, explicitly.
Approaches learning distributions in their latent space, like Generative Adversarial Networks (\cite{goodfellow2014generative}\cite{gupta2018social}\cite{sadeghian2019sophie}) or (conditional) Variational Autoencoders (\cite{kingma2014vae}\cite{sohn2015learning}\cite{schmerling2018multimodal}), are not considered here.
Lastly, this section is restricted to approaches capable of modeling multi-modal distributions, as many real world problems require multi-modal representations. 

In accordance with these criteria, sections \ref{ss:rw_single} and \ref{ss:rw_multi} focus on recent extensions of Gaussian Processes and different neural network models that generate probabilistic output, namely Bayesian Neuronal Networks and Mixture Density Networks.

\subsection{One-step prediction} 
\label{ss:rw_single}
In $1$-step prediction, the task is to infer the next value in a sequence, given the last $m$ preceding values of the same sequence.
It is oftentimes used as a building block of $n$-step prediction by performing $1$-step predictions iteratively.
Besides that, $1$-step prediction is also a fundamental component in Bayesian filtering for performing its \emph{prediction} step \cite{barshalom2002tracking}\cite{saerkkae2013filtering}.

\textbf{Gaussian Process} (\emph{GP}) regression \cite{rasmussen2003gaussian} is a popular model used in $1$-step prediction.
Given a collection of sample points of a non-linear function $f(\cdot): \mathbb{R}^m \rightarrow \mathbb{R}$, a mean function $m(\cdot)$ and a covariance function $k(\cdot, \cdot)$ (oftentimes also called \emph{kernel}) the GP yields a multivariate Gaussian prior probability distribution over functions intersecting given sample points.
This Gaussian distribution can then be used to calculate a conditional predictive distribution over the next value in the sequence given observed previous values \cite{ellis2009modelling}\cite{chandorkar2017probabilistic}.

\textbf{Deep Gaussian Processes} \cite{damianou2013deep} extend on the GP framework in order to constitute non-Gaussian, and therefore more complex models.
A deep GP is a hierarchy of multiple GPs using non-linear mappings between each layer of the hierarchy.
However, the resulting probability densities are intractable and thus require an approximate solution, which can be achieved e.g. by variational approximation \cite{campbell2015bayesian}.

\textbf{Bayesian Neural Networks} (\cite{bishop1995neural}, \emph{BNN}) treat all weights and biases as random variables.
Then, Bayesian inference is applied during training in order to determine the posterior distribution of the network parameters (sometimes referred to as \emph{Bayesian Backpropagation}).
Due to intractable probability distributions, either Monte Carlo methods \cite{neal1992bayesian} or approximate inference has to be applied.
Common techniques used for approximate inference include variational inference \cite{blundell2015weight}, inference based on expectation propagation \cite{hernandez2015probabilistic} and Monte Carlo dropout \cite{gal2016dropout}. 

\textbf{Mixture Density Networks} (\cite{bishop2006pattern}, \emph{MDN}) provide another neural network-based approach to probabilistic inference.
MDNs are deterministic neural networks, mapping the output of the last layer onto the parameters of a Gaussian (mixture) distribution.
Compared to BNNs, being a deterministic model, these networks are much simpler in terms of inference and computational cost, while still generating probabilistic output.
On the downside, MDNs do not allow to make assumptions about model uncertainty in a direct way.

\subsection{Multi-step prediction}
\label{ss:rw_multi}
The models presented in this section build upon the deep GP, BNN and MDN models from the previous section.
Each model is extended into a recurrent model by feeding its $1$-step prediction back into the model for generating the next prediction.
Similar to the $1$-step prediction setting, the task is to infer the next $n$ subsequent values of a sequence of interest given the last $m$ preceding values.

\textbf{Recurrent Gaussian Processes.} 
With respect to GP regression models, the standard GP model can be used for $n$-step prediction by embedding its $1$-step prediction model into a sequential Monte Carlo simulation \cite{ellis2009modelling}.
For the more complex deep GP, a recurrent extension has been proposed in \cite{mattos2015recurrent}, incorporating a recurrent variational approximation scheme for a deep GP using a state space model-based approach.
State space models are a way of representing the evolution of a stateful system through time by using non-linear mappings for the systems transition function \cite{hangos2006analysis}.
Being based on the deep GP model, the recurrent deep GP model requires an approximate inference approach for generating predictions.
Besides having a computation intensive inference scheme, GP-based approaches grant good control over generated predictions, by explicitly modeling the kernel functions, thus controlling the prior over functions representable by the model. 
This gives an advantage over most competing neural network-based approaches that generate sequences iteratively in a mostly unconstrained fashion.
In comparison, the model proposed in this paper also optimizes in function space in order to constrain generated predictions, but grants less control over the generated function than GP-based approaches as no explicit prior is given.
Due to these similarities, the recurrent deep GP model is used for comparison in the experimental section \ref{subsec:motion_modeling}.

\textbf{Bayesian Recurrent Neural Networks} (\emph{BRNN}) have been proposed in \cite{fortunato2017bayesian}, where the variational Bayesian Backpropagation scheme \cite{blundell2015weight} is adapted for backpropagation through time.
BNNs, and BRNNs respectively, offer robustness to over-fitting, allow probabilistic predictions and provide information about model uncertainty.
As a drawback, such models are difficult to train, due to the requirement of approximate inference making the training more computational intensive and potentially less stable.
Further, the need for approximate inference also yields a significant computational overhead when generating predictions.
Consequently, these models are less widely used in real world applications.

\textbf{Recurrent Mixture Density Networks} (\emph{RMDN}) are most commonly an adaptation of the model proposed in \cite{graves2013generating}, where an MDN is stacked on top of an LSTM.
In these models, the recurrent structure is used for encoding the observed sequence as well as for generating predictions.
It should be noted, that sometimes a temporal convolutional network is used instead of an LSTM for encoding observations due its less complex structure \cite{bai2018empirical}\cite{rudenko2019human}.
Compared to BRNNs, RMDNs have a simpler structure and thus are easier to train and control, which is the reason why these models are widely used for sequence prediction \cite{rudenko2019human}.
It has to be noted though, that many of these approaches (e.g. \cite{hasan2019forecasting}\cite{bartoli2018context}\cite{ha2018world}\cite{alahi2016social}) are only used for unimodal, maximum likelihood predictions.
This most likely stems from the fact that in this kind of model unimodal $n$-step prediction is cheap computation-wise, while multi-modal $n$-step prediction requires Monte Carlo simulation and thus additional computational cost \cite{hug2018particle}.
With regards to RMDN models, there are only few approaches actually targeting multi-modal predictions.
Two recent approaches are given in \cite{bhattacharyya2018accurate} and \cite{hug2018particle}.
Due to the focus on multi-modal sequence prediction, these approaches are used for comparison in the experimental section \ref{subsec:traj_pred}.

\section{Bézier curves with Gaussian control points} 
\label{sec:n-curves}
In this section, a Bézier curve capable of describing a (conditional) stochastic process $\mathcal{G}_T = \{X_t\}_{t \in T}$ with Gaussian random variables $X_t \sim \mathcal{N}(\mu_t, \Sigma_t)$ and index set $T = [0, 1]$ by using Gaussian control points is proposed.
Later on, this concept is extended for modeling random variables following a Gaussian mixture distribution. 
Throughout this section, several simple experiments are conducted to investigate different properties of the proposed model.
For these experiments, the proposed model is trained using the approach presented in section \ref{sec:ncdn}. 

\subsection{Modeling stochastic processes using Bézier curves}
\label{subsec:modeling}
A Bézier curve of degree $N$
\begin{align}
\label{eq:bezier-curve}
B(t, \mathcal{P}) = \sum_{i=0}^{N} b_{i,N}(t) P_i
\end{align}
is a polynomial curve constructed by a linear combination of $N+1$ $d$-dimensional \emph{control points} $\mathcal{P} = \{P_0, P_1, ... P_N\}$ using the Bernstein polynomials
\begin{align}
b_{i,N}(t) = \binom{N}{i} (1 - t)^{N - i} t^i
\end{align}
for weighting.
A curve point is determined by the curve parameter $t \in [0, 1]$.

When modeling a stochastic process $\mathcal{G}_T$ using such a parametric curve, each curve point needs to represent a Gaussian distribution.
Therefore, a "Gaussian" Bézier curve $\psi$ (abbrev.: $\mathcal{N}$-Curve), of degree $N$ is proposed.
The $\mathcal{N}$-Curve is an extension of equation \ref{eq:bezier-curve}, where the control points are replaced by $N+1$ Gaussian control points $\mathcal{P}_\mathcal{N} = \{P_0, P_1, ... P_N\}$ with $P_i \sim \mathcal{N}(\mu_i, \Sigma_i)~\forall P_i \in \mathcal{P}_\mathcal{N}$.
The set of mean vectors is denoted as $\mu_\mathcal{P} = \{\mu_0, \mu_1, ..., \mu_N\}$ and the set of covariance matrices $\Sigma_\mathcal{P} = \{\Sigma_0, \Sigma_1, ..., \Sigma_N\}$ respectively.
Thus the $\mathcal{N}$-Curve is defined by a tuple $\psi = (\mu_\mathcal{P}, \Sigma_\mathcal{P})$.
Given $\mathcal{P}_\mathcal{N}$, the stochasticity is inherited from the control points to the curve points $B_\mathcal{N}(t, \psi)~\forall t \in [0, 1]$ by the linear combination of the Gaussian control points, which again follows a Gaussian distribution.
Each curve point then defines the parameters of a (multivariate) Gaussian probability distribution
\begin{align}
	B_\mathcal{N}(t, \psi) = (\mu^\psi(t), \Sigma^\psi(t))
\label{eq:BN}
\end{align}
with
\begin{align}
\begin{split}
	\mu^\psi(t) &= \sum_{i=0}^{N} b_{i,N}(t) \mu_i,~\text{and} \\
	\Sigma^\psi(t) &= \sum_{i=0}^{N} \left(b_{i,N}(t)\right)^2 \Sigma_i.
\end{split}
\end{align}
Following $Ax + By + c \sim \mathcal{N}(A\mu_x + B\mu_y + c, A\Sigma_xA^T + B\Sigma_yB^T)$\footnote{Following the definition as provided in \emph{The Matrix Cookbook}\cite{petersen2008matrix}.} for $x \sim \mathcal{N}(\mu_x, \Sigma_x)$ and $y \sim \mathcal{N}(\mu_y, \Sigma_y)$, it can directly be seen that $B_\mathcal{N}(t, \psi)$ induces the Gaussian probability density
\begin{align}
	p^\psi_t(x) = p(x|\mu^\psi(t), \Sigma^\psi(t)) = \mathcal{N}(x|\mu^\psi(t), \Sigma^\psi(t))
\end{align}
at index $t$.

With respect to the stochastic process $\mathcal{G}_T = \{X_t\}_{t \in T}$, Gaussian probability distributions at discrete points in time $\{X_1, ..., X_n\}$ can now be described with $B_\mathcal{N}(t, \psi)$ using $n$ equally distributed values for $t$, yielding a discrete subset
\begin{align}
	T_* = \{\frac{v}{n-1}|v \in \{0, ..., n-1\}\} = \{t_1, ..., t_n\}
\label{eq:T}
\end{align}
of the index set $T$.
Thus, each process index (curve parameter) $t_i \in T_*$ corresponds to its respective sequence index at time $i \in \{1, ..., n\}$.
Following this, the Gaussian random variable $X_{t_i}$ at time $i$ is given by
\begin{align}
	X_{t_i} \sim \mathcal{N}(B_\mathcal{N}(t_i, \psi)) = \mathcal{N}(\mu^\psi(t_i), \Sigma^\psi(t_i))
\end{align}
with $P_0$ and $P_n$ as exact start and end conditions.

\begin{figure}[htb]
	\begin{center}
		\includegraphics[width=0.325\textwidth]{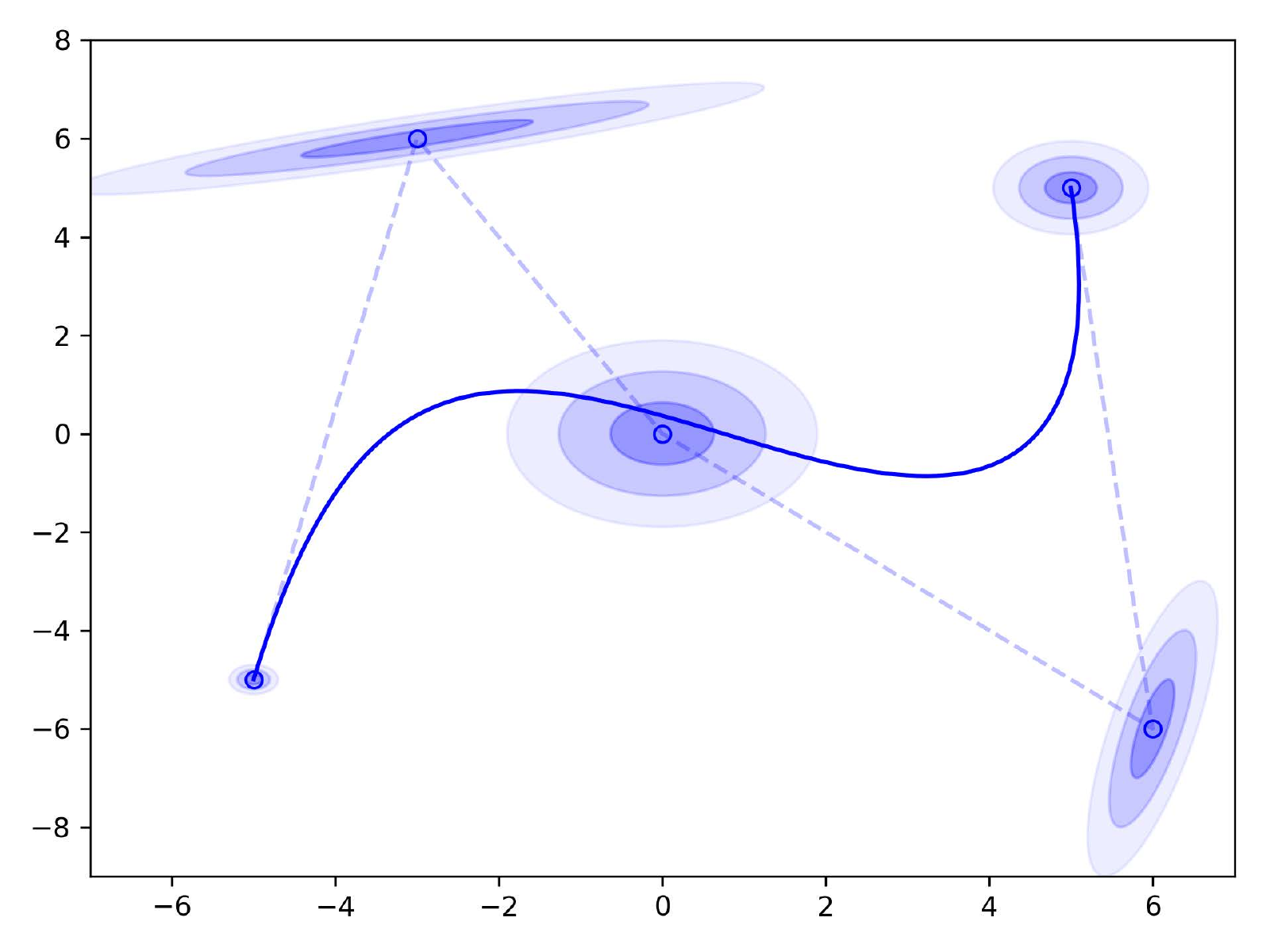}
		\includegraphics[width=0.325\textwidth]{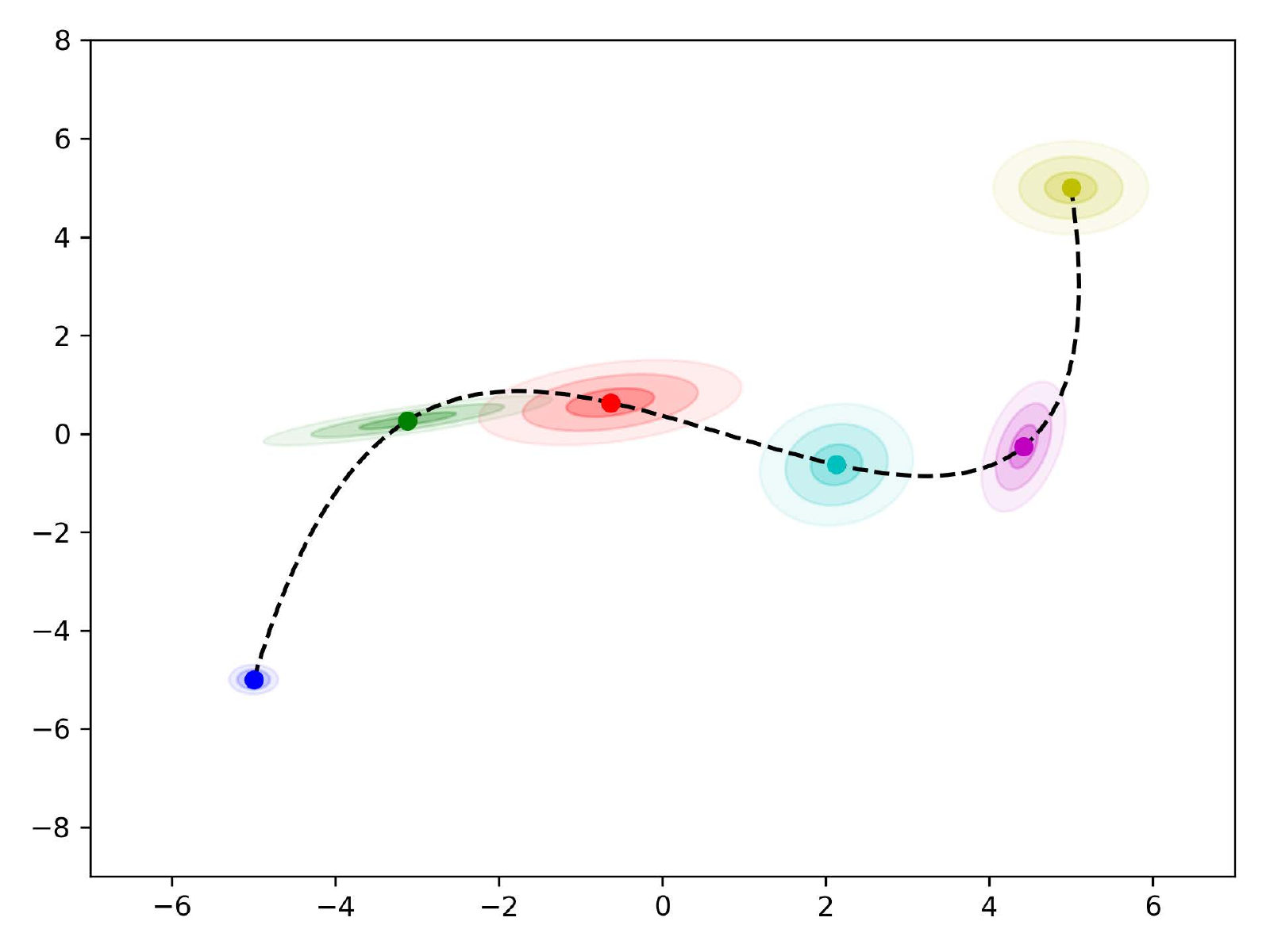}  	
		\includegraphics[width=0.325\textwidth]{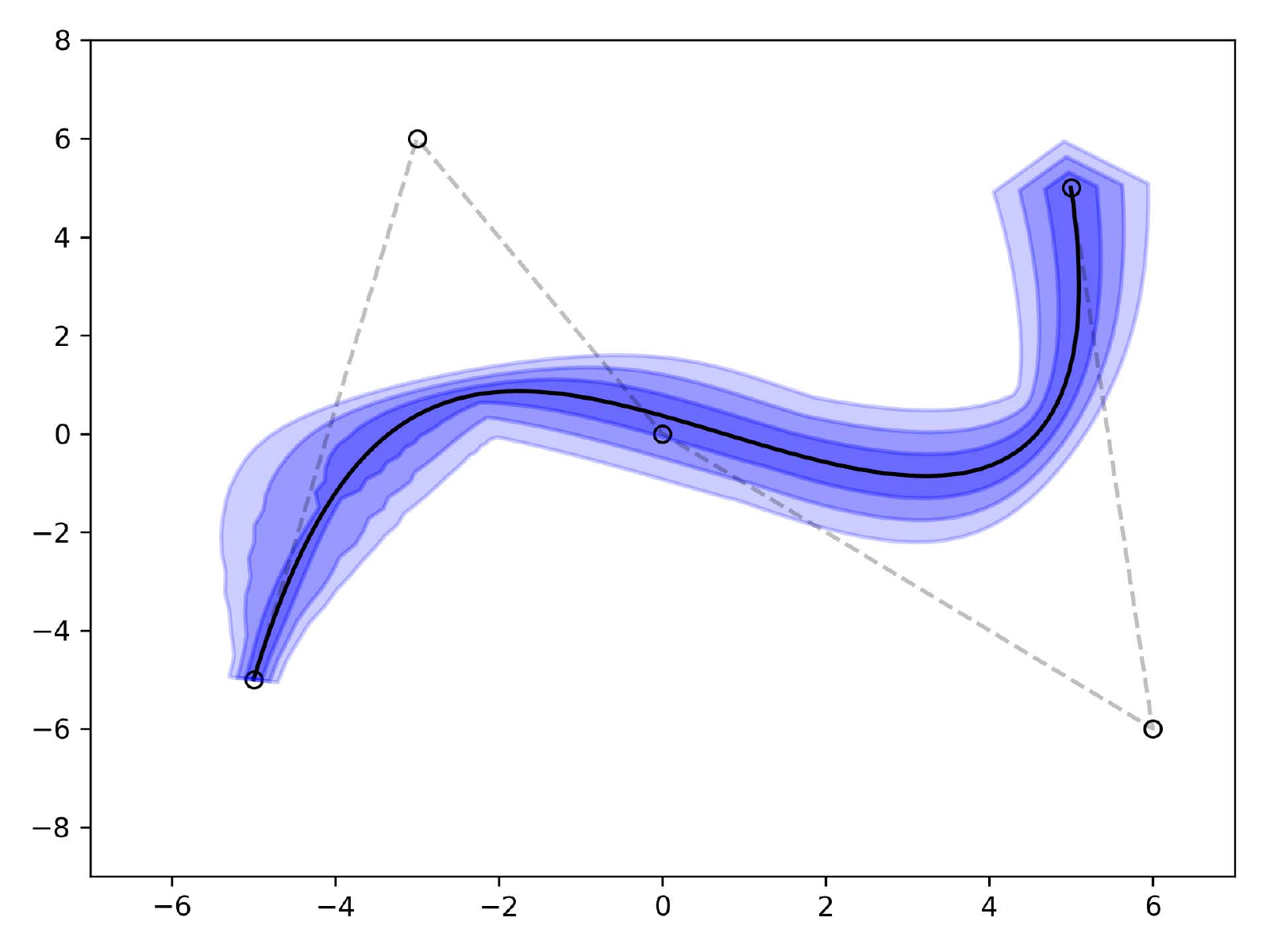}	
	\end{center}
	\caption{Left to right: Exemplary $2$-dimensional $\mathcal{N}$-Curve (mean curve as given by Gaussian control points), Gaussian distributions along the $\mathcal{N}$-Curve for different $t$ and the  $3\sigma$-region in which generated sample sequences lie.}
	\label{fig:splines}
\end{figure}

$\psi$ can be used as a generative model by either generating samples $X_t$ at specific points in time $t$ or by generating (continuous) realizations of the stochastic process. 
The latter is achieved by sampling a set of Bézier curve control points from $\psi$.
Figure \ref{fig:splines} depicts a $2$-dimensional example for a $\mathcal{N}$-Curve.
Here, a $\mathcal{N}$-Curve with 5 control points with respective covariance ellipses is shown (left).
Gaussian random variables $X_t$ along the $\mathcal{N}$-Curve given different values for the curve parameter $t$ are illustrated on the center image.
Note that the parametric curve interpolates the mean vectors of all Gaussian distributions through time.
Using the $\mathcal{N}$-Curve as a generative model, (almost) all realizations lie within the contour illustrated in the right image. 
This contour is approximated by overlaying the variances of random variables along the curve.

\subsubsection{Toy example 1: Approximating unimodal processes with varying noise.}
\label{sss:approx_uni}
\begin{figure}[htb]
	\begin{center}
		\includegraphics[width=0.4\textwidth]{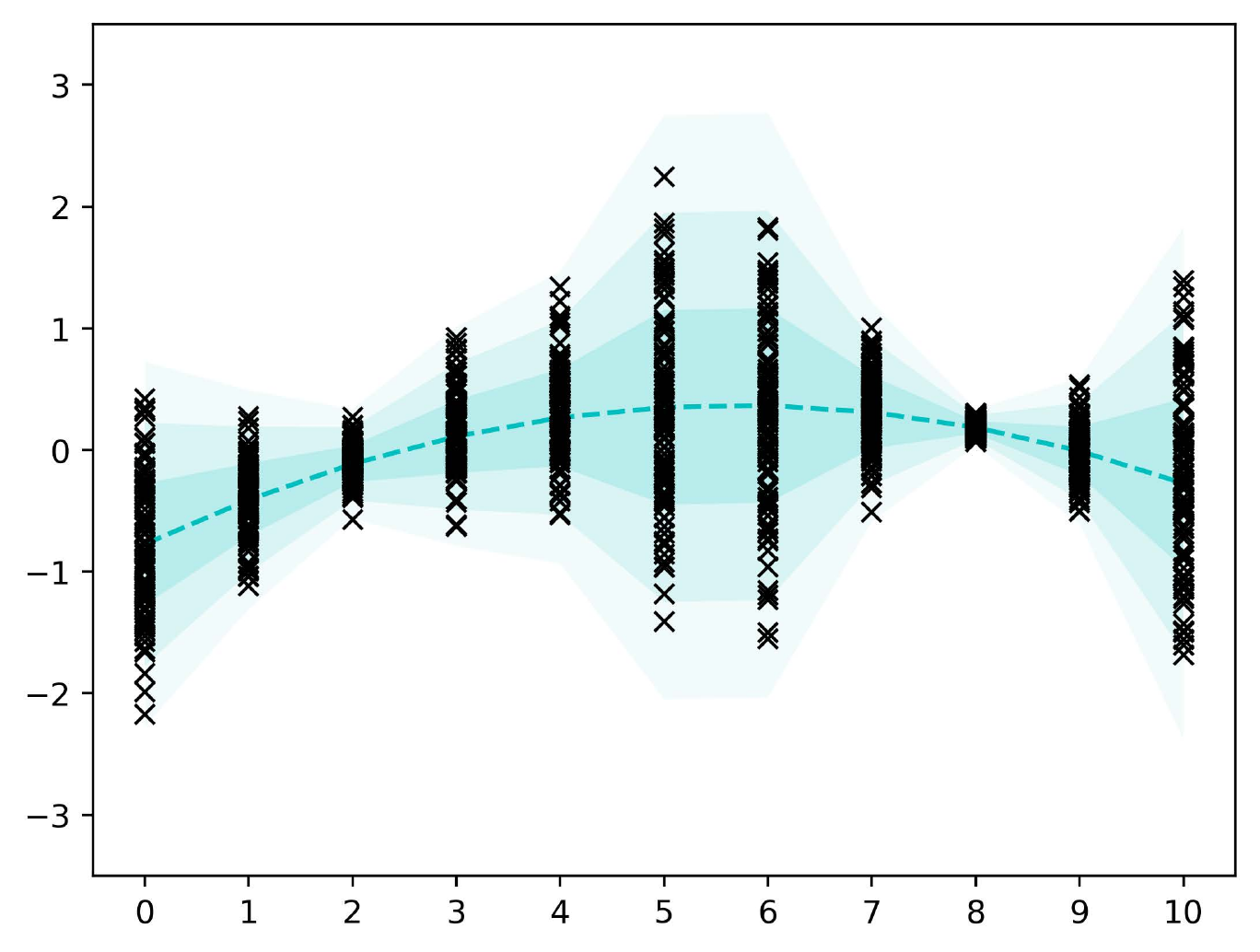}
		\includegraphics[width=0.4\textwidth]{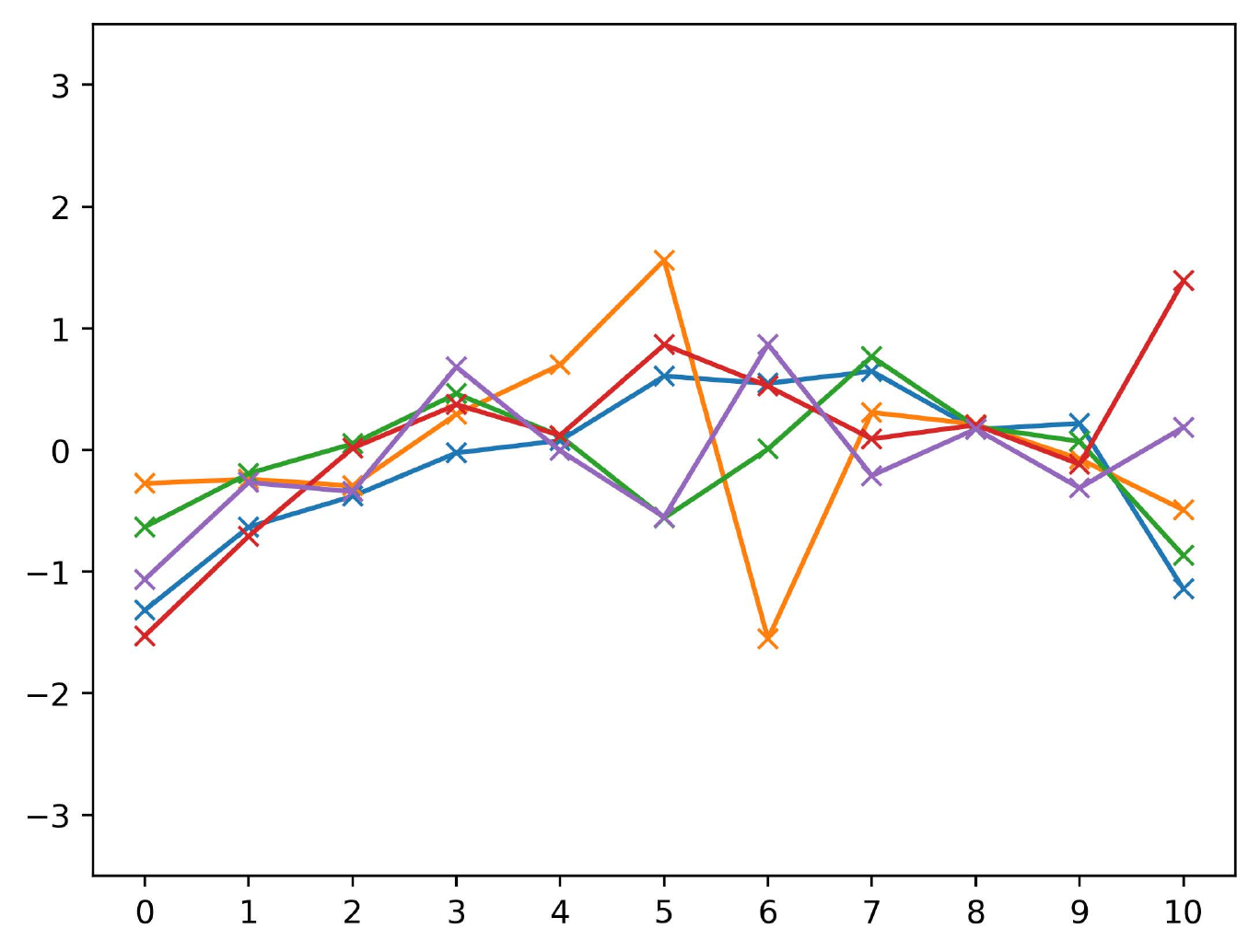}
	\end{center}
	\caption{Left: Ground truth mean values, standard deviations and sample points taken from a stochastic process at different points in time.
			Right: Sample realizations of the process.}
	\label{fig:te_ap_gt}
\end{figure}

In this example, the capabilities of the $\mathcal{N}$-Curve model as a representation for stochastic processes is illustrated.
As a simple experiment, an unimodal stochastic process with mean values moving along a curve is conducted.
At $11$ points in time samples have been taken around the respective mean values under varying standard deviations to show the impact of the number of control points used.
Figure \ref{fig:te_ap_gt} shows the mean curve, standard deviations and samples, as well as sample sequences used to learn the parameters of an $\mathcal{N}$-Curve. 

\begin{figure}[htb]
	\begin{center}
		\includegraphics[width=0.325\textwidth]{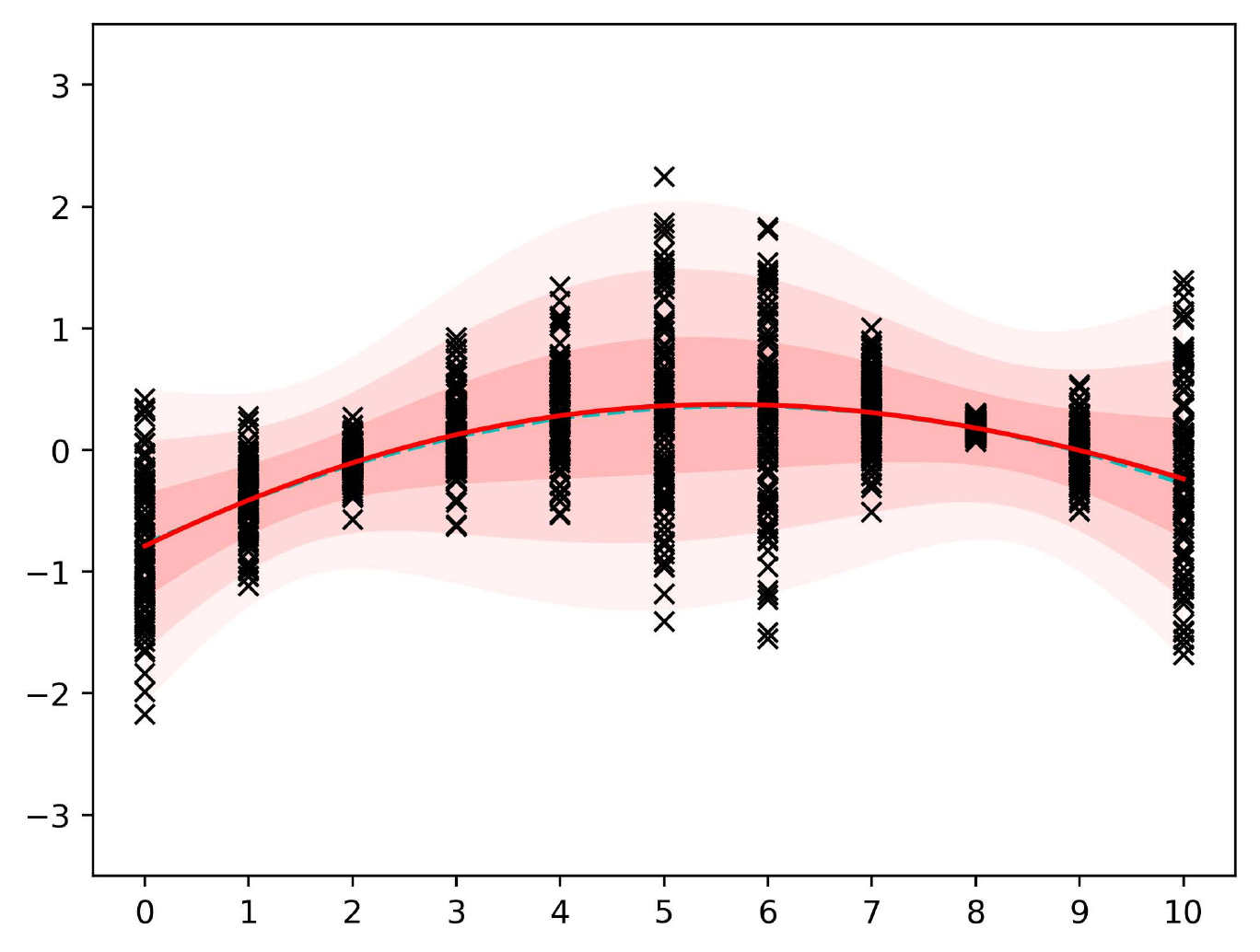}
		\includegraphics[width=0.325\textwidth]{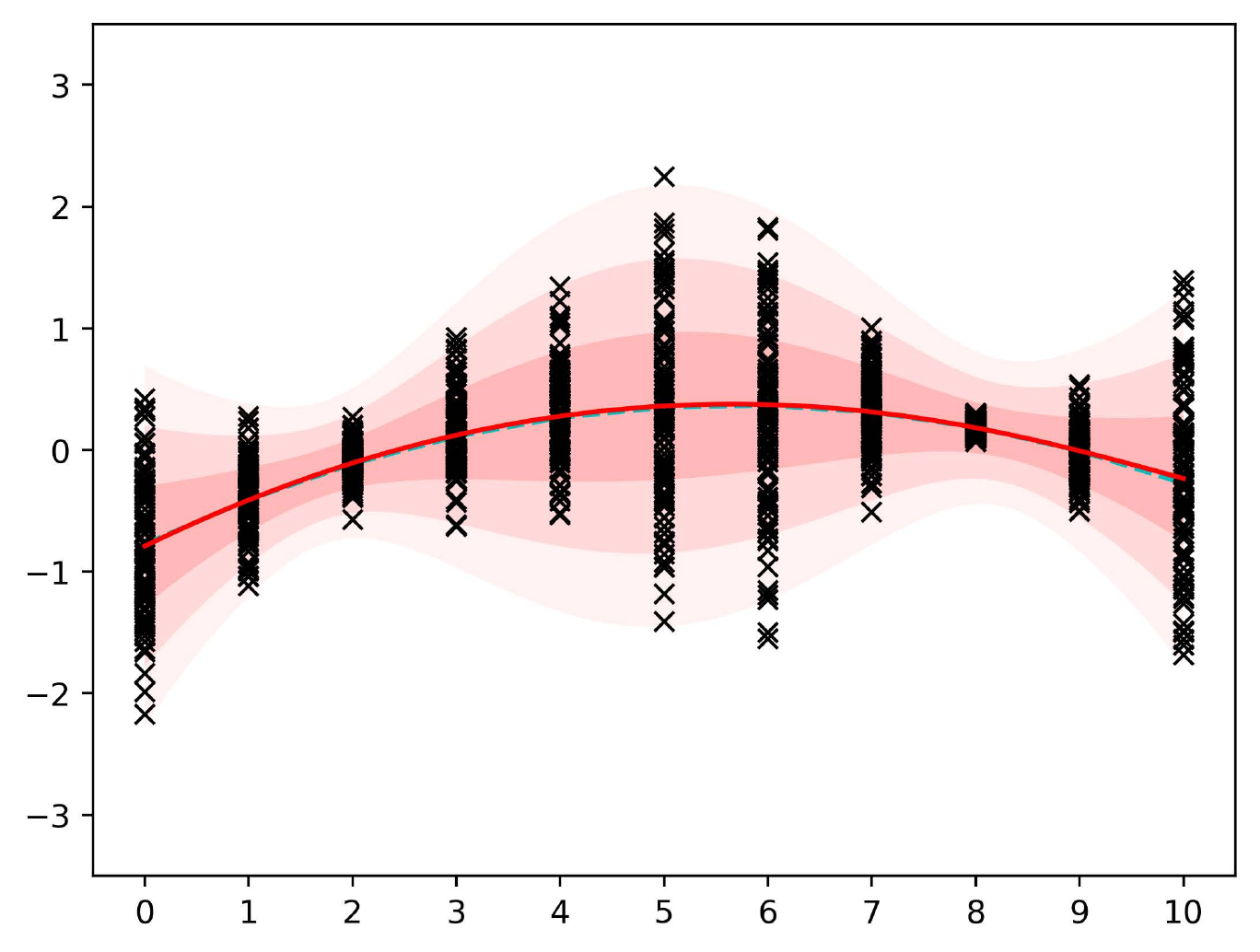}
		\includegraphics[width=0.325\textwidth]{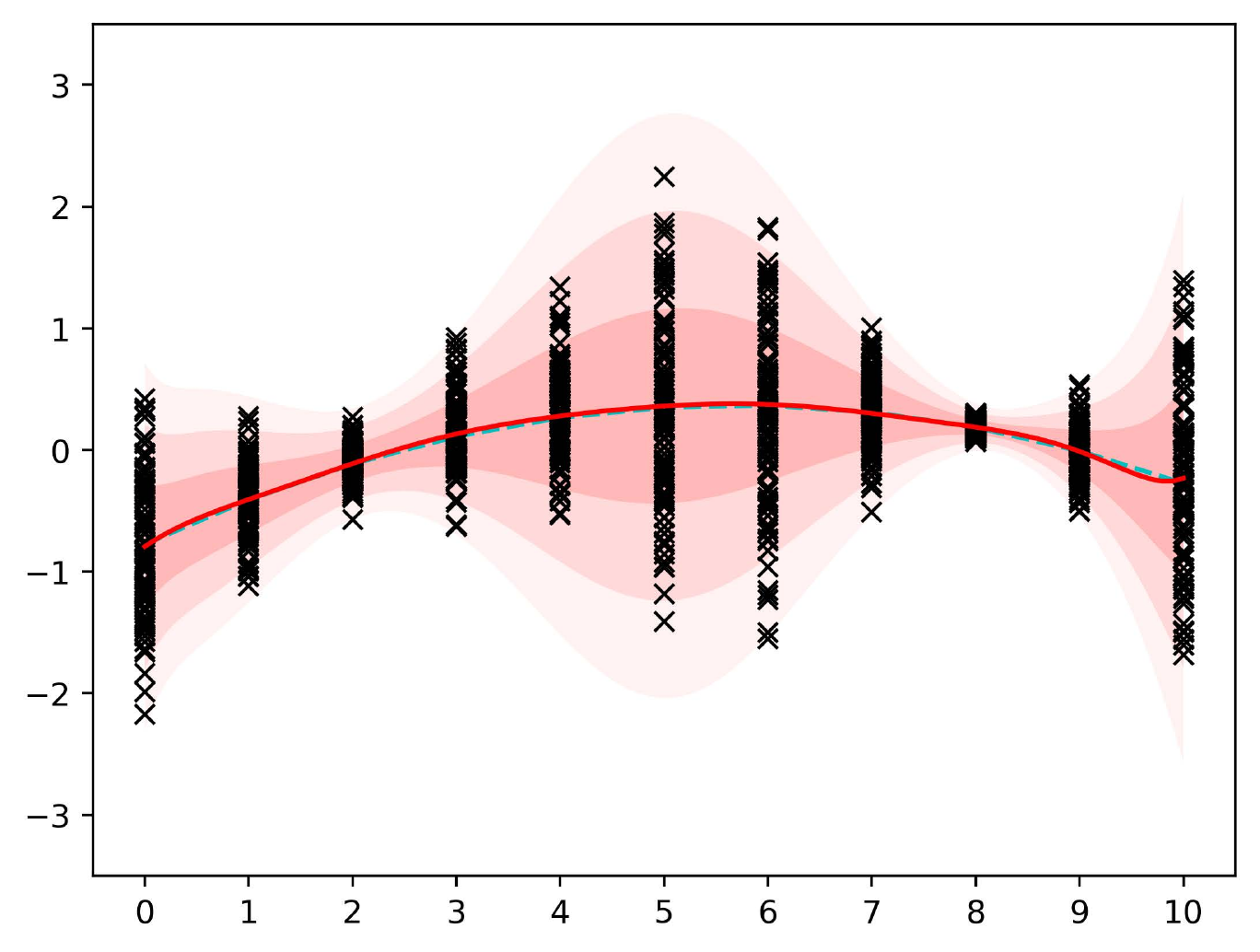}
	\end{center}
	\caption{Approximations of given stochastic process using $\mathcal{N}$-Curves with $5$, $7$ and $15$ control points (left to right).}
	\label{fig:te_ap_pred}
\end{figure}

Resulting $\mathcal{N}$-Curves with $5$, $7$ and $15$ control points are depicted in figure \ref{fig:te_ap_pred}.
It can be seen that the $\mathcal{N}$-Curve model learns a smooth mean curve and compensates noise using the variance of the control points.
Further, with an increasing number of control points, more variation in input noise can be compensated.

\subsubsection{Toy example 2: $\mathcal{N}$-Curve instant inference vs. SMC inference.}
\label{sss:smc}
One of the targeted properties of the $\mathcal{N}$-Curve is to move inference into the training phase, thus allowing for instant prediction of several time steps.
This is opposed to sequential monte carlo (\emph{SMC}) approaches, which learn the transition of subsequent time steps and perform iterative inference after training.
In this experiment, the performance of the $\mathcal{N}$-Curve model is compared to an SMC approach.
For comparison an LSTM-MDN model generating a prediction for one step at a time embedded into a particle filter cycle is used (comparable to \cite{hug2018particle}).

\begin{figure}[htb]
	\begin{center}
		\includegraphics[width=0.325\textwidth]{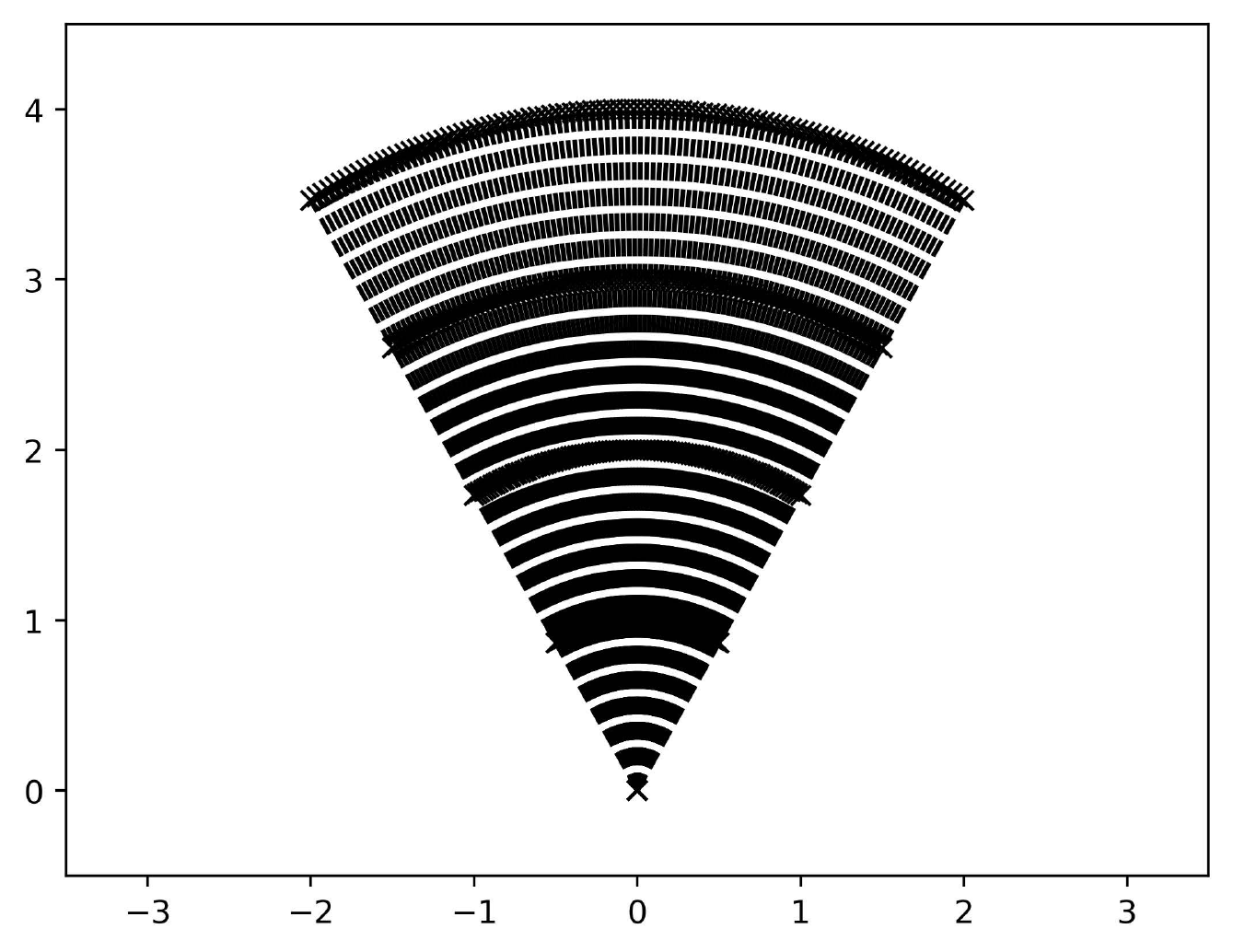}
		\includegraphics[width=0.325\textwidth]{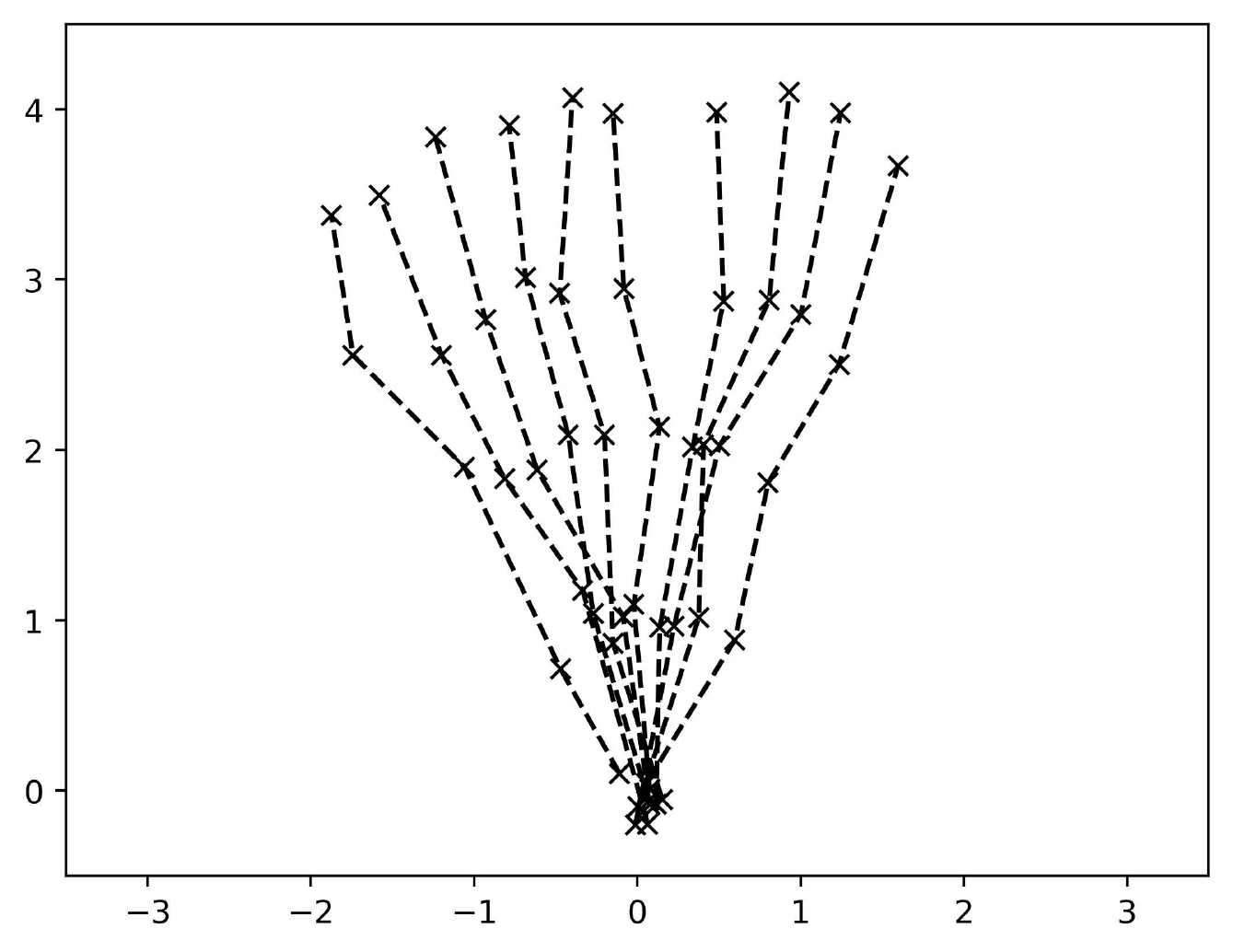}
		\includegraphics[width=0.325\textwidth]{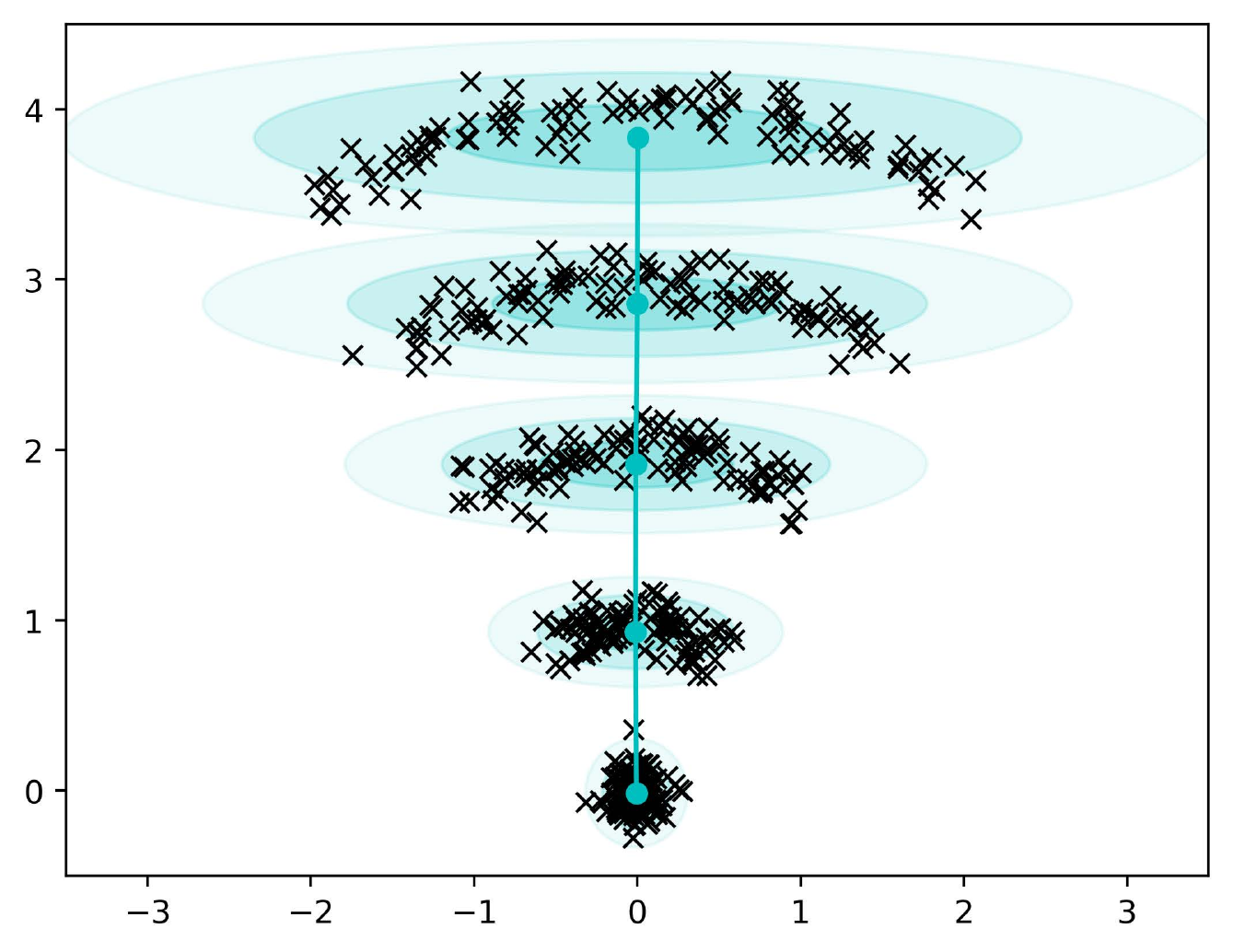}
	\end{center}
	\caption{Left to right: Noise-free trajectories, noisy training samples and expected Gaussian distributions for each trajectory point.}
	\label{fig:te_smc_gt}
\end{figure}

In order to inspect the behavior of both approaches, the training data consists of 2D trajectories consisting of $5$ points starting at the origin moving upwards at different angles resembling a fan.
This can be seen in the left image in figure \ref{fig:te_smc_gt}.
Following this, i.i.d. noise $\epsilon \sim \mathcal{N}(0, \Sigma)$ is added to each trajectory point. 
Some samples are depicted in the center image.
Given these samples, the goal is to learn a representation of the dataset in the form of a stochastic process $\mathcal{G} = \{X_t\}_{t \in [1, ..., 5]}$ with $X_t$ being Gaussian random variables.
The expected Gaussian distribution for each of the $5$ trajectory points is depicted in the right image.

\begin{figure}[htb]
	\begin{center}
		\includegraphics[width=0.4\textwidth]{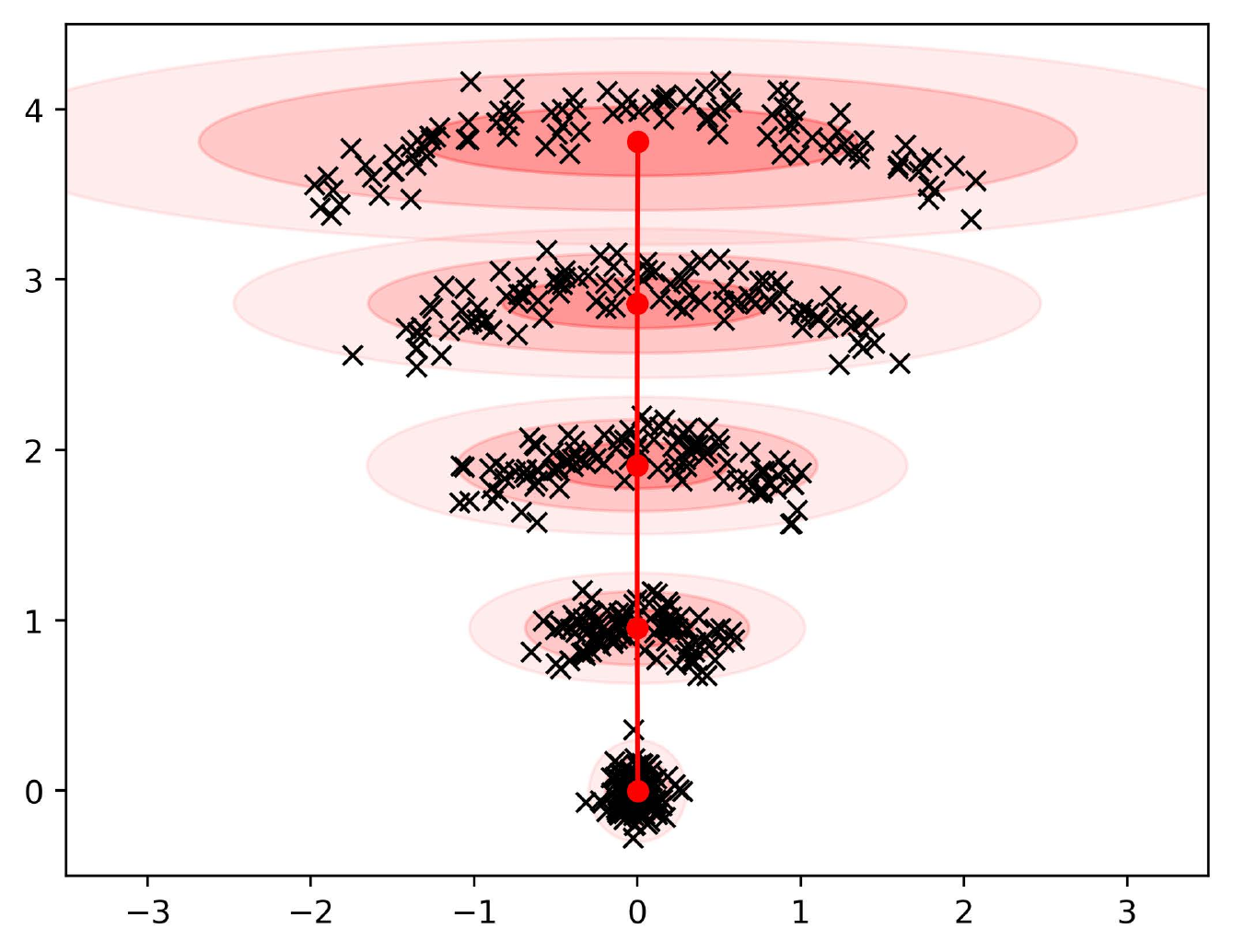}
		\includegraphics[width=0.4\textwidth]{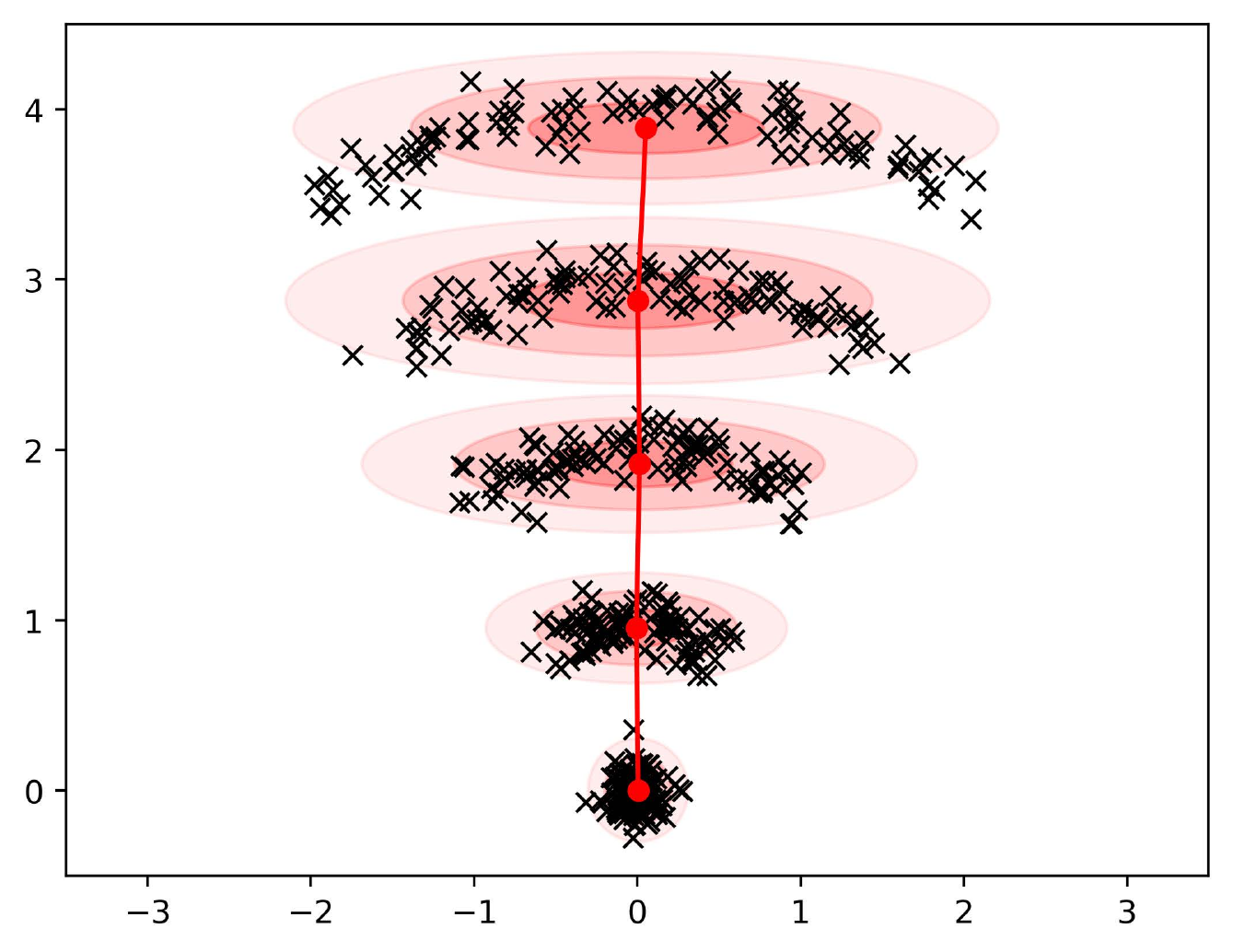}
	\end{center}
	\caption{Gaussian distributions for each trajectory point generated by the $\mathcal{N}$-Curve model (left) and an LSTM-MDN model (right).}
	\label{fig:te_smc_pred}
\end{figure}
Figure \ref{fig:te_smc_pred} shows the Gaussian distributions generated by the $\mathcal{N}$-Curve model (left) and the LSTM-MDN model (center).
While the distributions are generated using one forward-pass using the $\mathcal{N}$-Curve model, $100$ particles and $10$ iterations were used to generate the distributions using the LSTM-MDN model.
It can be seen that both models approximate the mean vectors with only small errors and the increasing variance is captured almost correctly.
Starting with the results of the $\mathcal{N}$-Curve model, it seems to over-estimate the variance a little bit, but overall it's result is similar to the expected sequence of distributions.
Looking at the LSTM-MDN model, it captures the increase in the variance in all steps except the last one.
This may be due to sampling errors or mode collapse, which is a common problem with mixture density networks \cite{makansi2019overcoming}.
Further, it looks like the model is under-estimating the variance, which is due to the fact that the model propagates Gaussian distributed samples through time, while the points in the training data follow a uniform distribution causing higher variance when approximated using a Gaussian distribution.
This is not a flaw by the model, but to be expected.

\subsection{Extending $\mathcal{N}$-Curves for Gaussian mixture probability distributions}
\label{subsec:extend}
While Gaussian probability distributions are a sufficient representation for unimodal sequence data, many real world problems require multi-modal representations.
For this case, a common approach is to use a Gaussian mixture probability distribution $\Upsilon(\{\pi_k\}_{k \in \{1, ..., K\}},\allowbreak \{(\mu_k, \Sigma_k)\}_{k \in \{1, ..., K\}})$ defined by $K$ weighted Gaussian components and probability density function
\begin{align}
	p(x) = \sum_{k=1}^{K} \pi_k \mathcal{N}(x|\mu_k, \Sigma_k), \text{with}~\sum_k \pi_k = 1.
\end{align}
In the same way the concept of $\mathcal{N}$-Curves can be extended to a mixture $\Psi$ of $K$ weighted $\mathcal{N}$-Curves $\{\psi_1, ..., \psi_K\}$ with normalized weights $\pi = \{\pi_1, ..., \pi_K\}$.
The random variables $X_t$ at index $t \in T$ then follow the Gaussian mixture distribution
\begin{align}
	X_t \sim \Upsilon(\pi, \{B_\mathcal{N}(t, \psi_k)\}_{k \in \{1, ..., K\}}).
\end{align}
Accordingly, the probability density at $t \in T$ induced by $\Psi$ is given by
\begin{align}
	p^\Psi_t(x) = \sum_{k=1}^{K} \pi_k \mathcal{N}(x|\mu^{\psi_k}(t), \Sigma^{\psi_k}(t)),
\label{eq:mix}
\end{align}
with $\mu^{\psi_k}(t)$ and $\Sigma^{\psi_k}(t)$ induced by the Gaussian distribution at $t \in T$ according to the $k$'th $\mathcal{N}$-Curve, i.e. $(\mu^{\psi_k}(t), \Sigma^{\psi_k}(t)) = B_\mathcal{N}(t, \psi_k)$.

\subsubsection{Toy example 3: Approximating "structured" multi-modal processes with varying noise.}
\label{sss:multi_te}
\begin{figure}[htb]
	\begin{center}
		\includegraphics[width=0.325\textwidth]{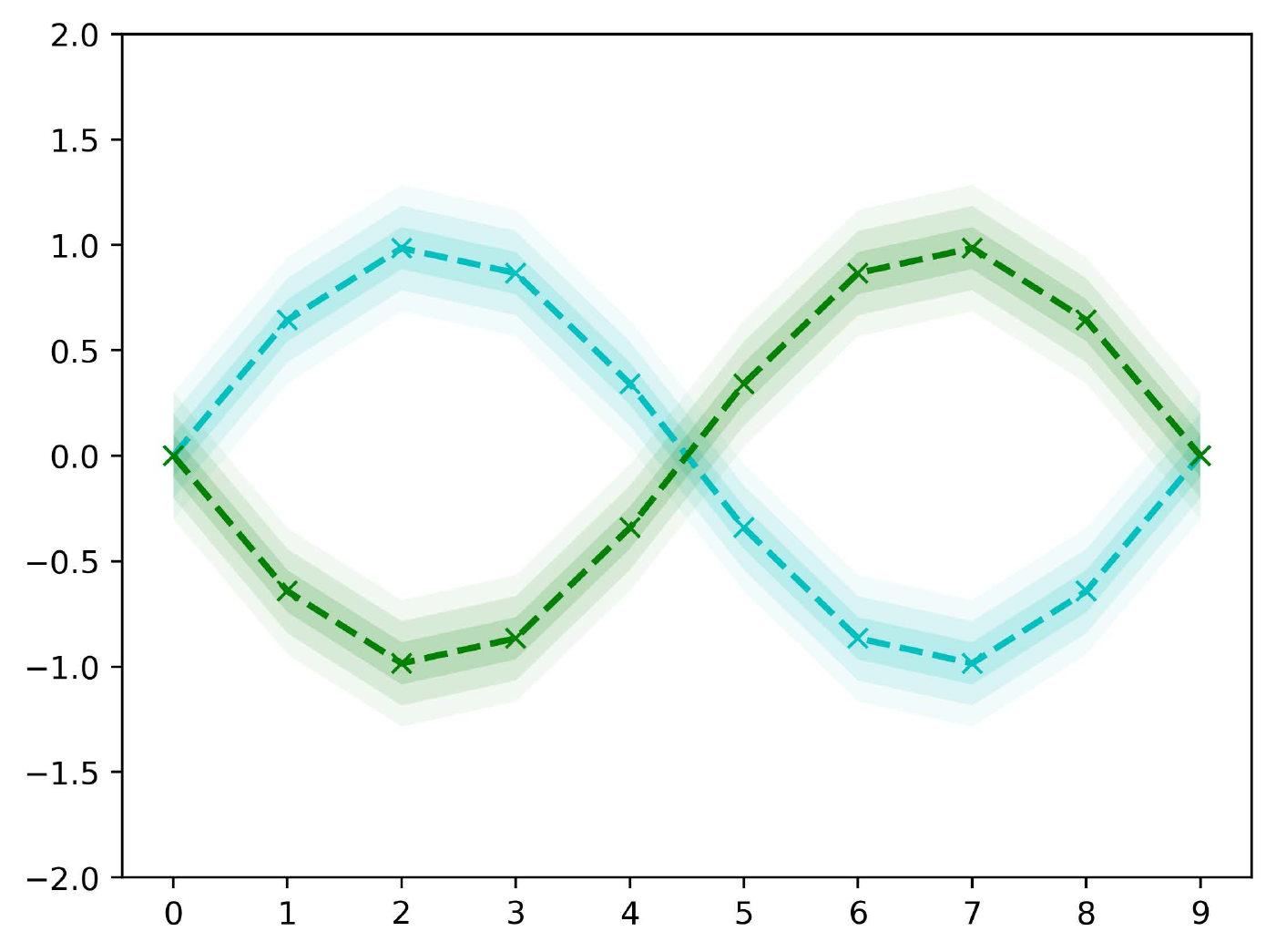}
		\includegraphics[width=0.325\textwidth]{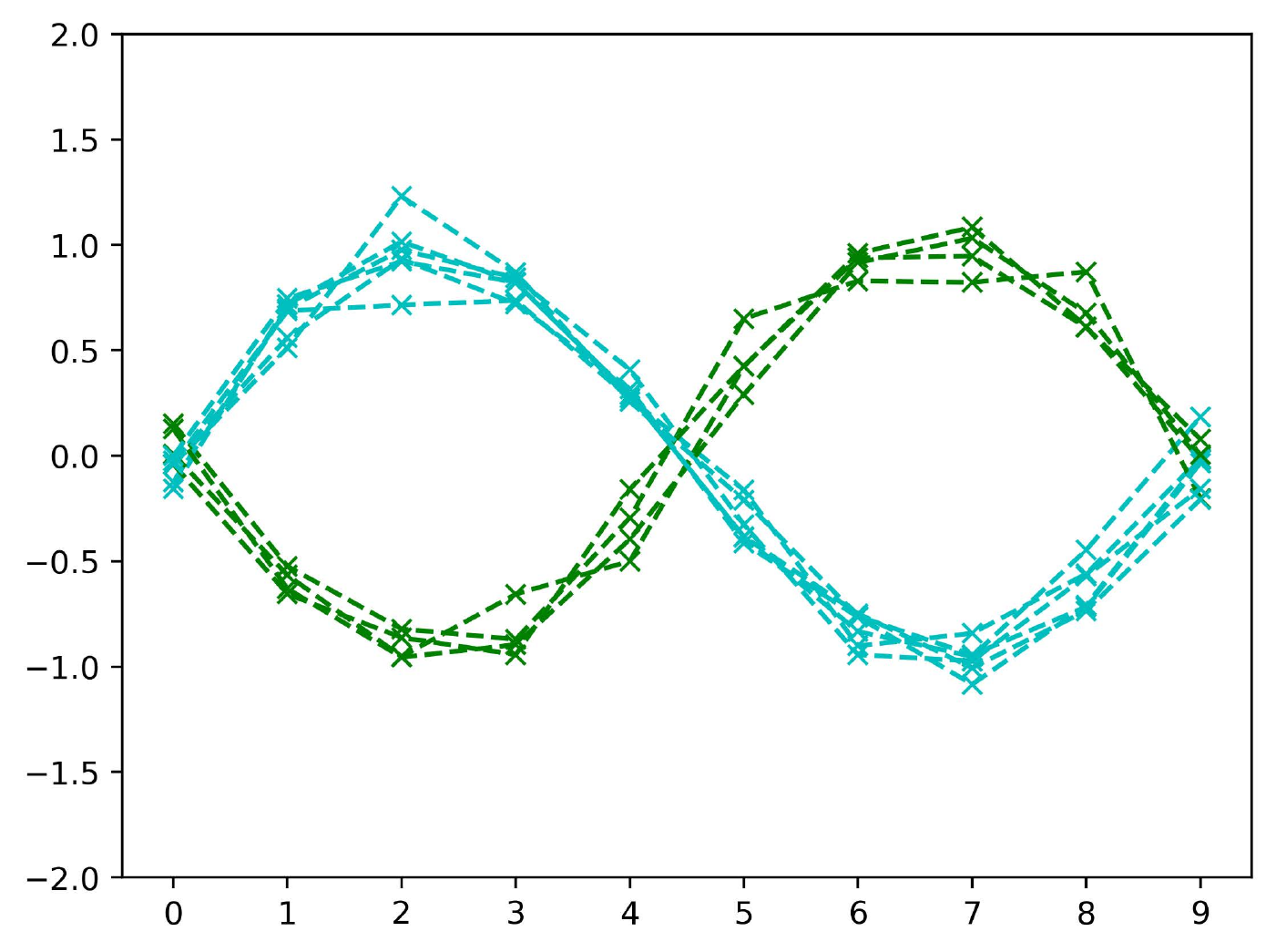}
		\includegraphics[width=0.325\textwidth]{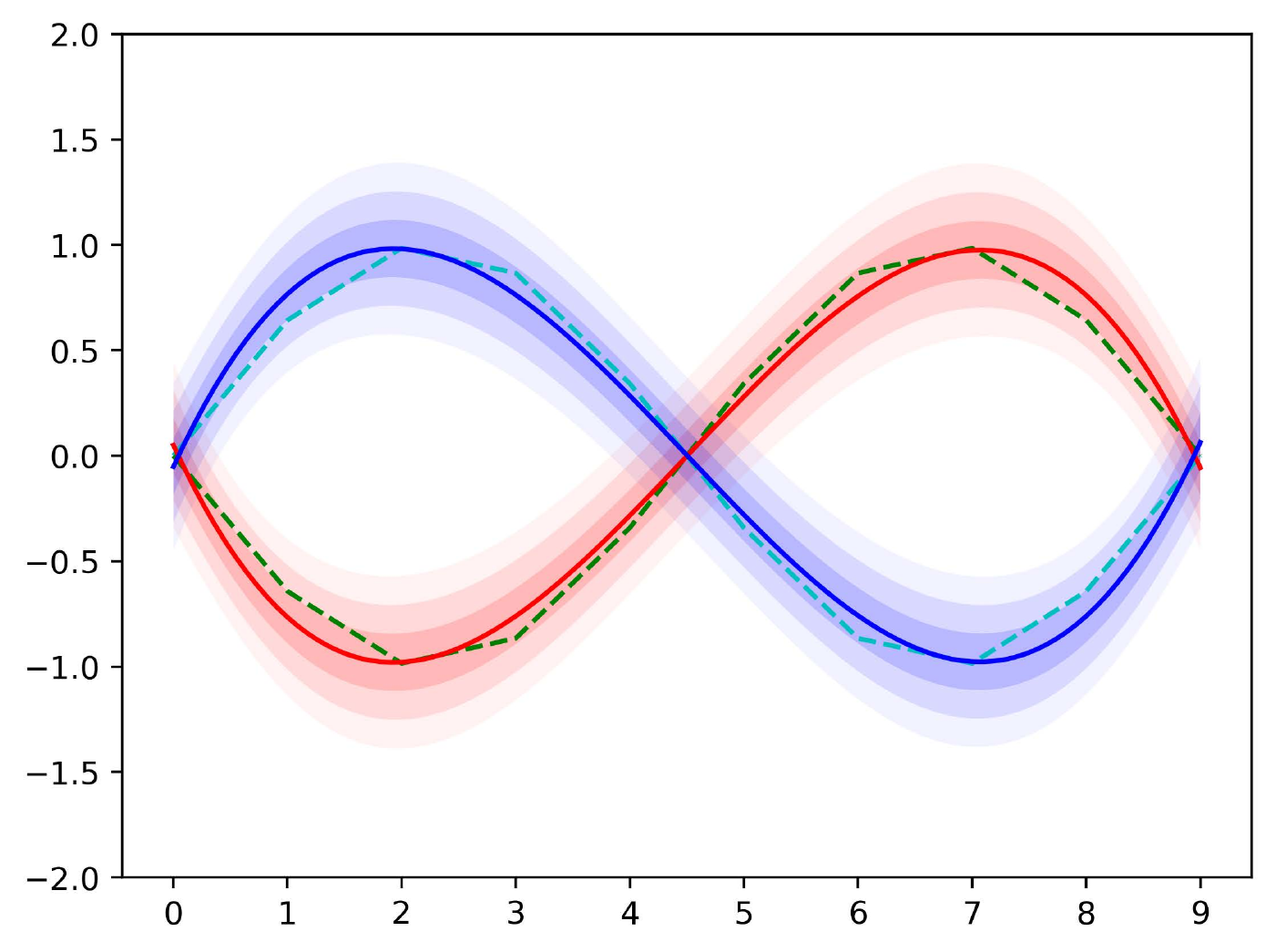}
	\end{center}
	\caption{Stochastic process modeling two curves (left), sample realizations (center) and an approximation given by an $\mathcal{N}$-Curve mixture model using $k=2$ components (right).}
	\label{fig:te_multi}
\end{figure}

Similar to the unimodal example, the capabilities of the $\mathcal{N}$-Curve mixture model in representing stochastic processes is illustrated in this experiment.
Here, each random variable of the process follows a bimodal Gaussian mixture distribution and realizations of the process follow one of two possible paths, as shown in the center image of figure \ref{fig:te_multi}.
This constraint on the realizations introduces a specific structure (two distinct curves) into the training dataset that can be captured by the $\mathcal{N}$-Curve mixture model.
The variance is constant for all time steps.
The left image in figure \ref{fig:te_multi} shows the ground truth mean and standard deviations along both curves and the right image the approximation given an $\mathcal{N}$-Curve mixture model using $k=2$ components.
It can be seen that due to the structure in the training data, the $\mathcal{N}$-Curve mixture model is capable of approximating the ground truth distributions.

\begin{figure}[htb]
	\begin{center}
		\includegraphics[width=0.4\textwidth]{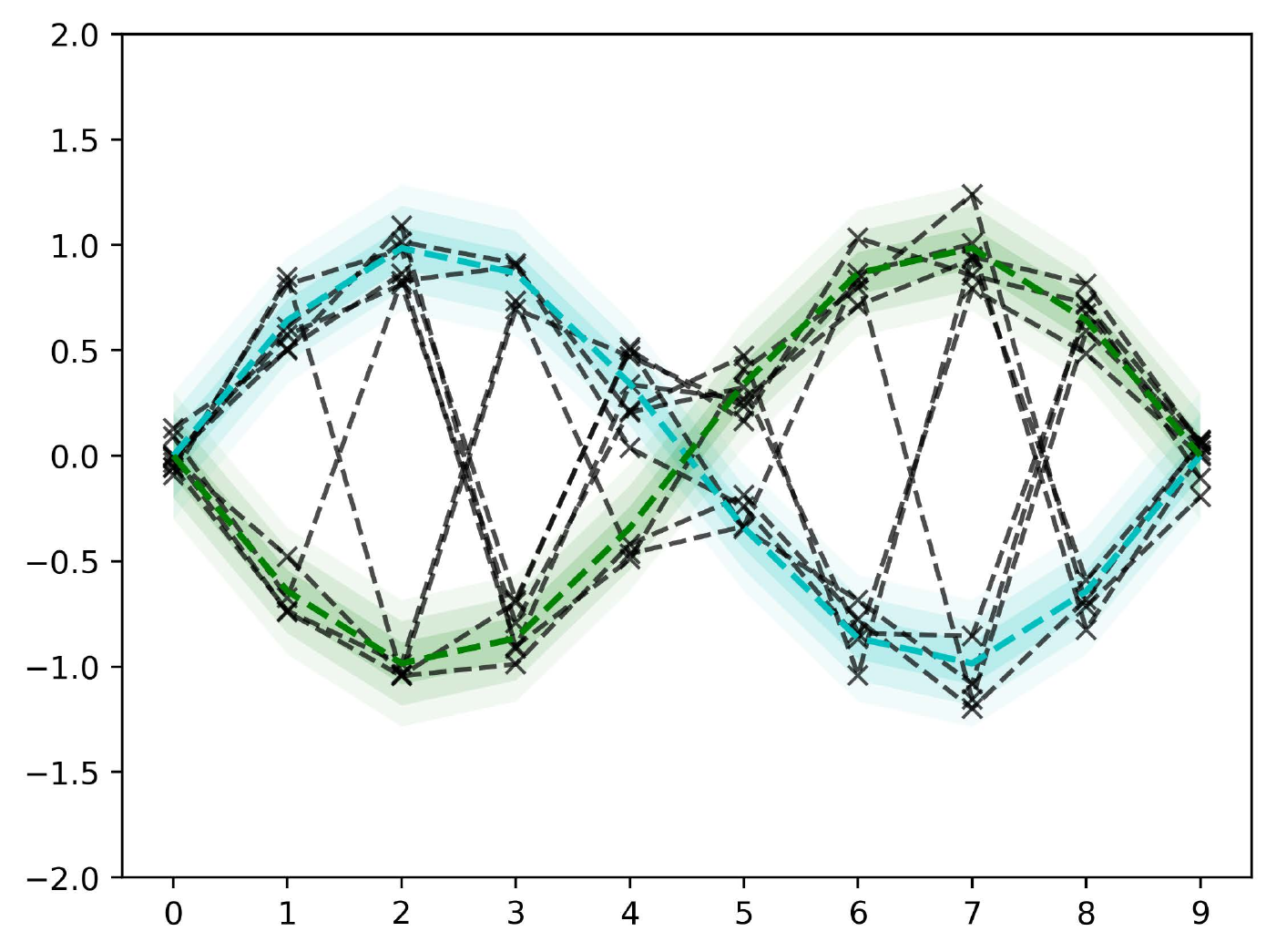}
		\includegraphics[width=0.4\textwidth]{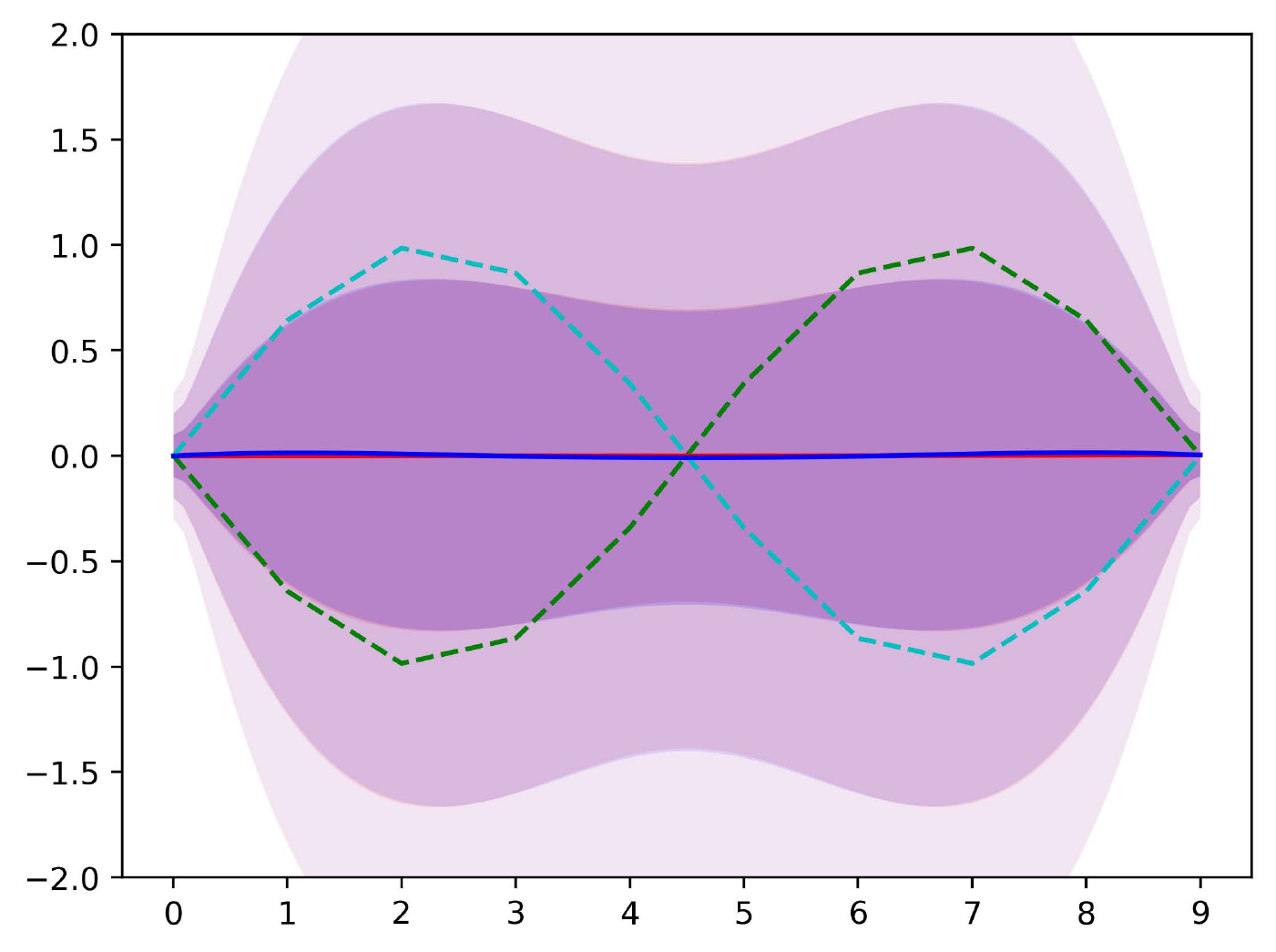}
	\end{center}
	\caption{Unstructured sample realizations (left) and an unimodal approximation given by an $\mathcal{N}$-Curve mixture model using $k=2$ components (right).}
	\label{fig:te_multi_unstructured}
\end{figure}

When removing this structure from the data, such that subsequent samples in a sequence are not constrained to the same curve, the $\mathcal{N}$-Curve mixture model collapses onto one mode (the mean between both curves).
This is illustrated in figure \ref{fig:te_multi_unstructured}.
In the right image, both $\mathcal{N}$-Curves in the mixture are overlayed.
While both curves have nearly identical parameters, the weight of one of the curves has been driven towards zero during training.
The mode collapse in this case is to be expected, as it is not possible to interpret the previous structure given unstructured data.
With respect to this, choosing mean values between both curves and compensating the curves deviation from the mean using its variance yields a correct result.

\section{$\mathcal{N}$-Curve Mixture Density Networks}  
\label{sec:ncdn}
For learning the parameters of an $\mathcal{N}$-Curve mixture from discrete sequence data, a Mixture Density Network (MDN) is proposed.
A mixture density network, as introduced in \cite{bishop2006pattern}, is a simple, most commonly single layer neural network $\Phi(\mathcal{V}) = (\pi, \mu, \Sigma|\mathcal{V})$, that takes a vector $\mathcal{V}$ as input and maps it onto the parameters of a mixture of Gaussians, i.e. the mixing weights $\pi$, the mean vectors $\mu$, standard deviations $\sigma$ and correlations $\rho$.
By generating the mapping from a given $\mathcal{V}$, the mixture can be conditioned on arbitrary inputs. 
Usually, the network is used to learn a Mixture of Gaussians and can be thought of as an alternative to using the EM algorithm that easily supports learning conditional probability distributions.
According to equation (\ref{eq:mix}), an MDN $\Phi$ outputs the parameters of an $\mathcal{N}$-Curve mixture, i.e. the mixing weights and Gaussian distribution parameters for each Gaussian control point $\Phi(\mathcal{V}) = \{(\pi_k, \psi_k)\}_{k \in \{1, ..., K\}} = \{(\pi_k, (\mu_{\mathcal{P},k}, \Sigma_{\mathcal{P},k}))\}_{k \in \{1, ..., K\}}$.
Advantages of using an MDN for learning the $\mathcal{N}$-Curve mixture parameters, rather than other algorithms, like EM, are two-fold.
First, the MDN provides an easy approach to learn and process conditional $\mathcal{N}$-Curve mixtures (and thus conditional stochastic processes), allowing the model to be used in a conditional inference framework.
Second, the MDN can be incorporated easily into (almost) any neural network architecture without the need to control the gradient flow.  

In the following, a method for learning $\mathcal{N}$-Curve control points and mixture weights from training data is described.
Let $\hat{\mathcal{S}} = \{S_1, ..., S_M\}$ be a set of $M$ realizations of a stochastic process with $\mathcal{S}_j = \{x^{\mathcal{S}_j}_1, ..., x^{\mathcal{S}_j}_n\}$ where each $x^{\mathcal{S}_j}_i$ with $i \in \{1, ..., n\}$ is a sample value for the respective random variable $X_{t_i}$ at time $i$ for $t_i \in T_*$ (see \ref{subsec:modeling}).
In order to simplify the training procedure, independence of Gaussian distributions along the $\mathcal{N}$-Curve is assumed.
This assumption yields, that the joint probability of the samples in a sequence $\mathcal{S}_j$ along an $\mathcal{N}$-Curve $\psi$ can be calculated as  
\begin{align}
	p^\psi(\mathcal{S}_j) = p^\psi(x^{\mathcal{S}_j}_1, ..., x^{\mathcal{S}_j}_{n}) = \prod_{i=1}^{n} p^\psi_{t_i}(x^{\mathcal{S}_j}_{i}).
\label{eq:seq_likelihood}
\end{align}
This property is exploited when defining the loss function.
It has to be noted, that $\prod_{i=1}^{n} p^\psi_{t_i}(x^{\mathcal{S}_j}_{i})$ is an unnormalized Gaussian density, with all $p^\psi_{t_i}(x^{\mathcal{S}_j}_{i})$, for $i \in \{1, ..., n\}$ and $t_i \in T_*$, being Gaussian densities.

For a single sequence $\mathcal{S}_j$ and a single $\mathcal{N}$-Curve $\psi = \Phi(\mathcal{V})$, the loss function is defined by the negative log-likelihood (unnormalized likelihood) 
\begin{align}
\begin{split}
	\mathcal{L} &= -log~p^\psi(x^{\mathcal{S}_j}_1, ..., x^{\mathcal{S}_j}_{n}) \\
				&= -log\left( \prod_{i=1}^{n} p^\psi_{t_i}(x^{\mathcal{S}_j}_i) \right) \\
				&= -\sum_{i=1}^{n} log~p^\psi_{t_i}(x^{\mathcal{S}_j}_i) \\
				&= -\sum_{i=1}^{n} log~p(x^{\mathcal{S}_j}_i|\mu^\psi(t_i), \Sigma^\psi(t_i))
\end{split}
\end{align}
of the sequence given an input vector $\mathcal{V}$, using the equations introduced in section \ref{subsec:modeling}.
Therefore, the loss for a set of $M$ sequences $\hat{\mathcal{S}} = \{\mathcal{S}_1, ..., \mathcal{S}_M\}$ is simply defined as the sum over the negative log-likelihoods for each sequence
\begin{align}
	\mathcal{L} = \sum_{j=1}^{M} \left( -\sum_{i=1}^{n} log~p(x^{\mathcal{S}_j}_i|\mu^\psi(t_i), \Sigma^\psi(t_i)) \right). 
\label{eq:loss_single}
\end{align}

Equation (\ref{eq:loss_single}) can easily be extended for $\mathcal{N}$-Curve mixtures.
Given an $\mathcal{N}$-Curve mixture $\Psi = \Phi(\mathcal{V})$, the likelihood of a single training sequence $\mathcal{S}_j$ is now calculated as the weighted linear combination of the likelihood of $\mathcal{S}_j$ for each $\psi_k$ (see equation (\ref{eq:seq_likelihood})):
\begin{align}
\begin{split}
	p^{\Psi}(\mathcal{S}_j) &= \sum_{k=1}^{K} \pi_k p^{\psi_k}(\mathcal{S}_j).
\end{split}
\end{align}
Thus, the loss for a set of $M$ sequences $\hat{\mathcal{S}}$ can then be defined as 
\begin{align}
\begin{split}
	\mathcal{L} &= \frac{1}{M} \sum_{j=1}^{M} -log~\sum_{k=1}^{K} \pi_k p^{\psi_k}(\mathcal{S}_j) \\
				&= \frac{1}{M} \sum_{j=1}^{M} -log~\sum_{k=1}^{K} \pi_k \prod_{i=1}^{n} p^\psi_{t_i}(x^{\mathcal{S}_j}_i) \\
				&= \frac{1}{M} \sum_{j=1}^{M} -log~\sum_{k=1}^{K} exp \left( log \pi_k + \sum_{i=1}^{n} log \left( p^\psi_{t_i}(x^{\mathcal{S}_j}_i) \right) \right).			
\end{split}
\label{eq:logloss}
\end{align}
Given the loss function and training sequences, the $\mathcal{N}$-Curve mixture density network model can be trained with gradient descent. 
Since the sum of negative log likelihoods may result in large loss values and a less stable optimization it is better to use the mean of the likelihoods, when long sequences or many samples should be processed.

\subsection{Toy example 4: Learning $\mathcal{N}$-Curve control points from noisy data}
For testing the $\mathcal{N}$-Curve mixture density network's capability of learning the parameters of an $\mathcal{N}$-Curve mixture from sequence data, a simple experiment is conducted.
In order to remove as much complexity as possible, the input vector $\mathcal{V}$ is set to be constant.
In consequence, the parameters of an unconditioned $\mathcal{N}$-Curve mixture are learned.
To enable proper visualization of results, this examples uses $2$-dimensional data.
Following this, an experiment is set up like this:
\begin{enumerate} 
	\item Define an arbitrary $2$-component $\mathcal{N}$-Curve mixture $\Psi_{GT}$ with $4$ Gaussian control points per component by setting the mixing weights $\pi$, and $\mu^k_\mathcal{P}$ and $\Sigma^k_\mathcal{P}$ for the Gaussian control points $\mathcal{P}^k_\mathcal{N}$ of each $\mathcal{N}$-Curve $\psi_k$
	\item Draw a set of $M = 1000$ Bézier curves from $\Psi_{GT}$
	\item Determine $I = 25$ (arbitrary, but fixed time horizon) evenly distributed values for $t$ in the range $[0, 1]$, thus $t \in \{\frac{i}{I-1}|i \in [0, ..., I-1]\}$ for discretizing each of the $M$ curves into a set $\hat{\mathcal{S}}$ of sequences of samples
	\item Train the network using $\hat{\mathcal{S}}$ and check if the network is able to reconstruct $\Psi_{GT}$, i.e. the mixing weights $\pi$ and the parameters of the $\mathcal{N}$-Curves in the mixture
\end{enumerate}

\begin{figure}[htb]
	\begin{center}
		\includegraphics[width=0.4\textwidth]{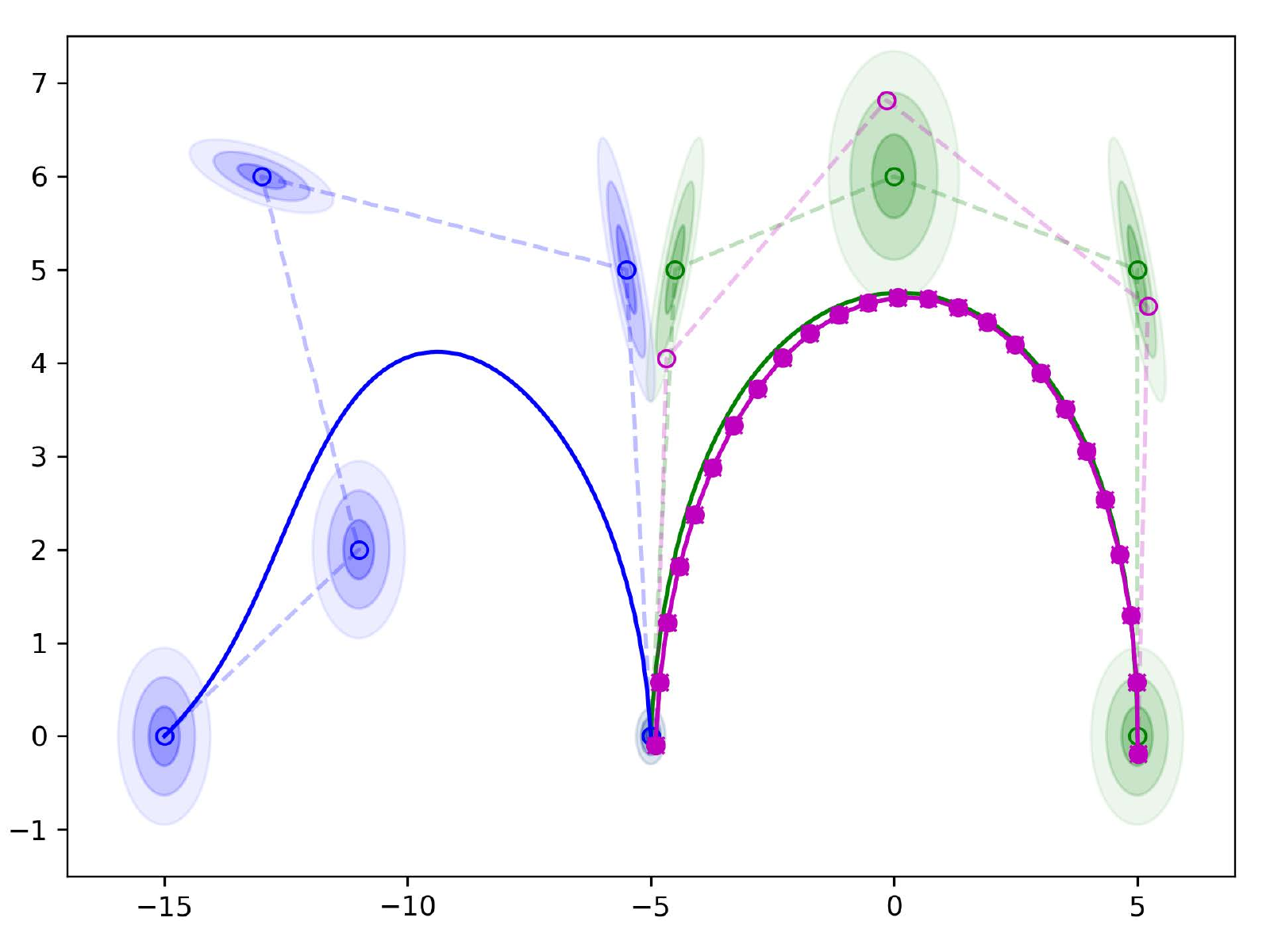}
		\includegraphics[width=0.4\textwidth]{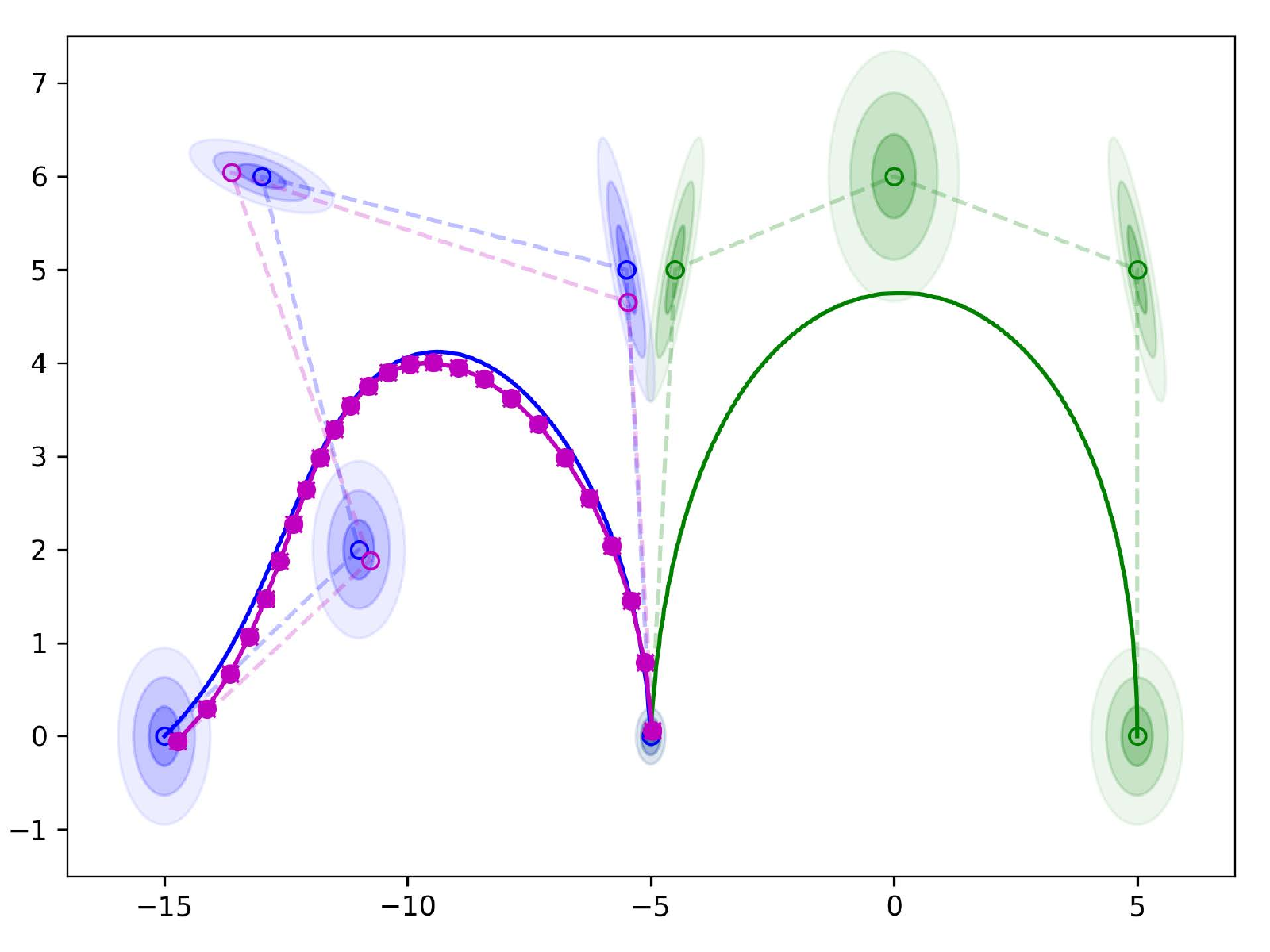}
	\end{center}
	\caption{Ground truth $\mathcal{N}$-Curves (green and blue) alongside sample sequences (purple).}
	\label{fig:te_learn_cpts_samples}
\end{figure}

The ground truth $\mathcal{N}$-Curve mixture and samples are depicted in figure \ref{fig:te_learn_cpts_samples}.
The $\mathcal{N}$-Curves start at $(-5, 0)$ and end at $(6, 0)$ and $(-15, 0)$, respectively.
The mixing weights are $\pi_{blue} = 0.25$ and $\pi_{green} = 0.75$.

\begin{figure}[htb]
	\begin{center}
		\begin{tabular}{ccc}
			\includegraphics[width=0.325\textwidth]{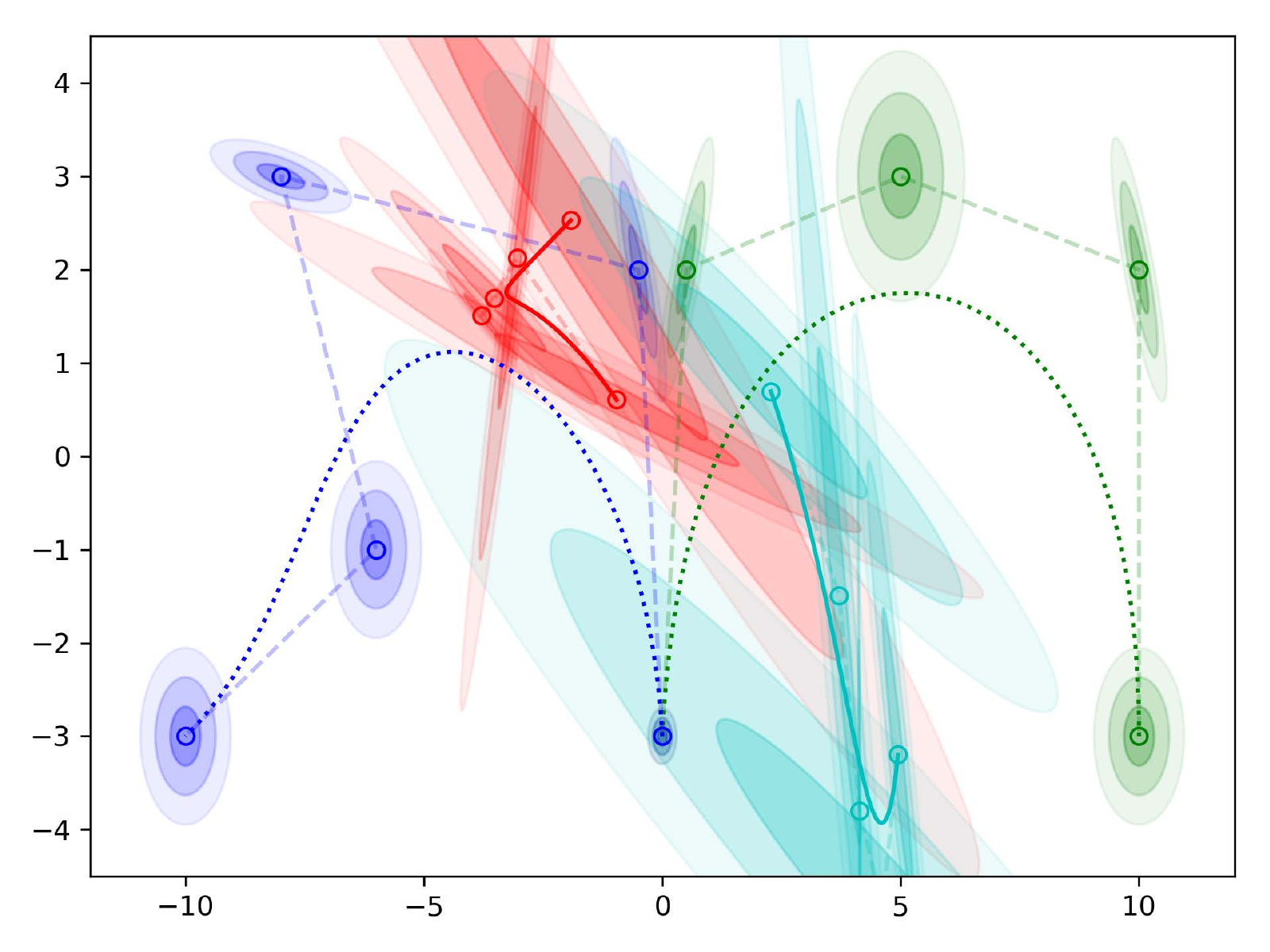} 
			& \includegraphics[width=0.325\textwidth]{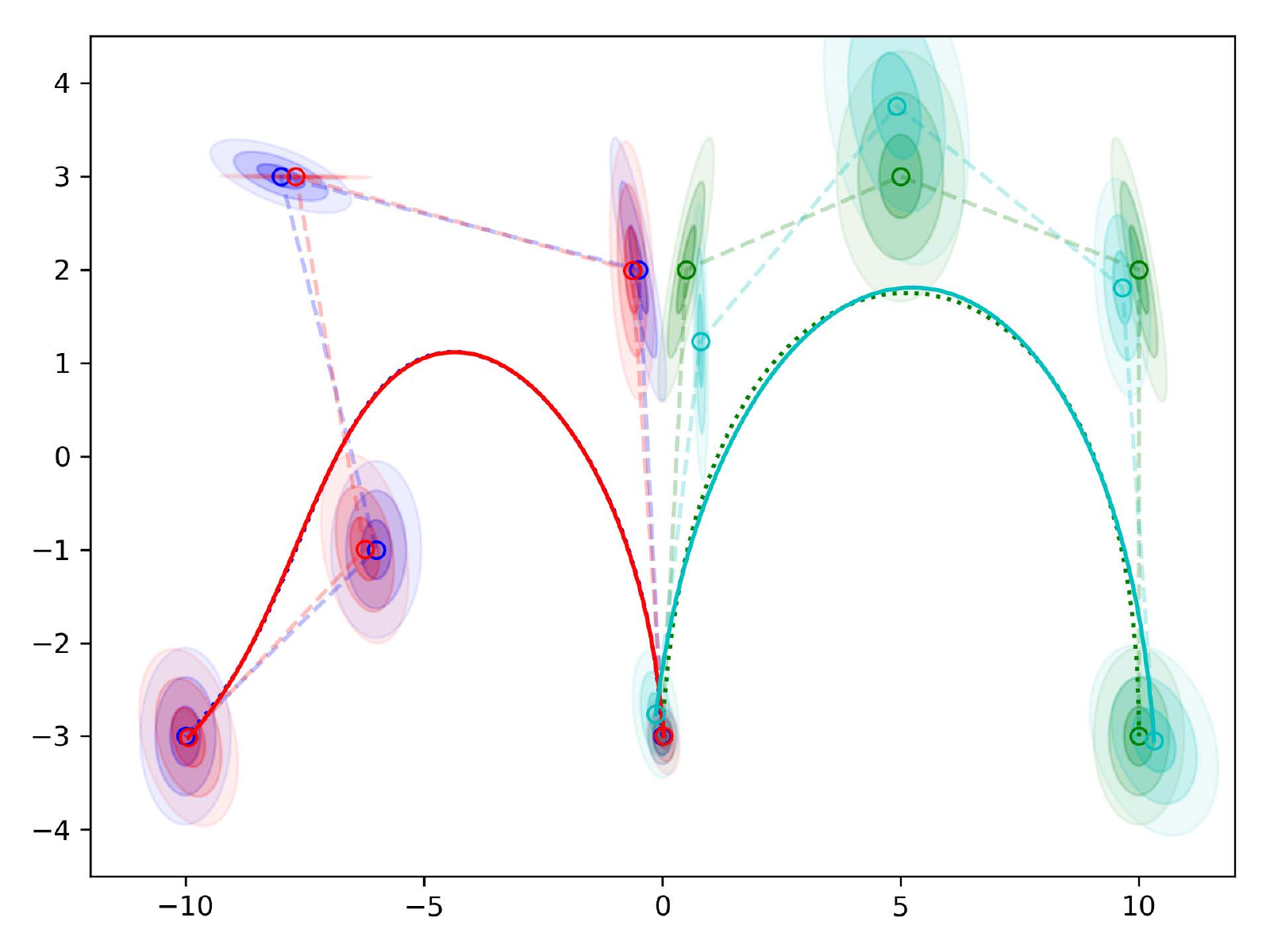}
			& \includegraphics[width=0.325\textwidth]{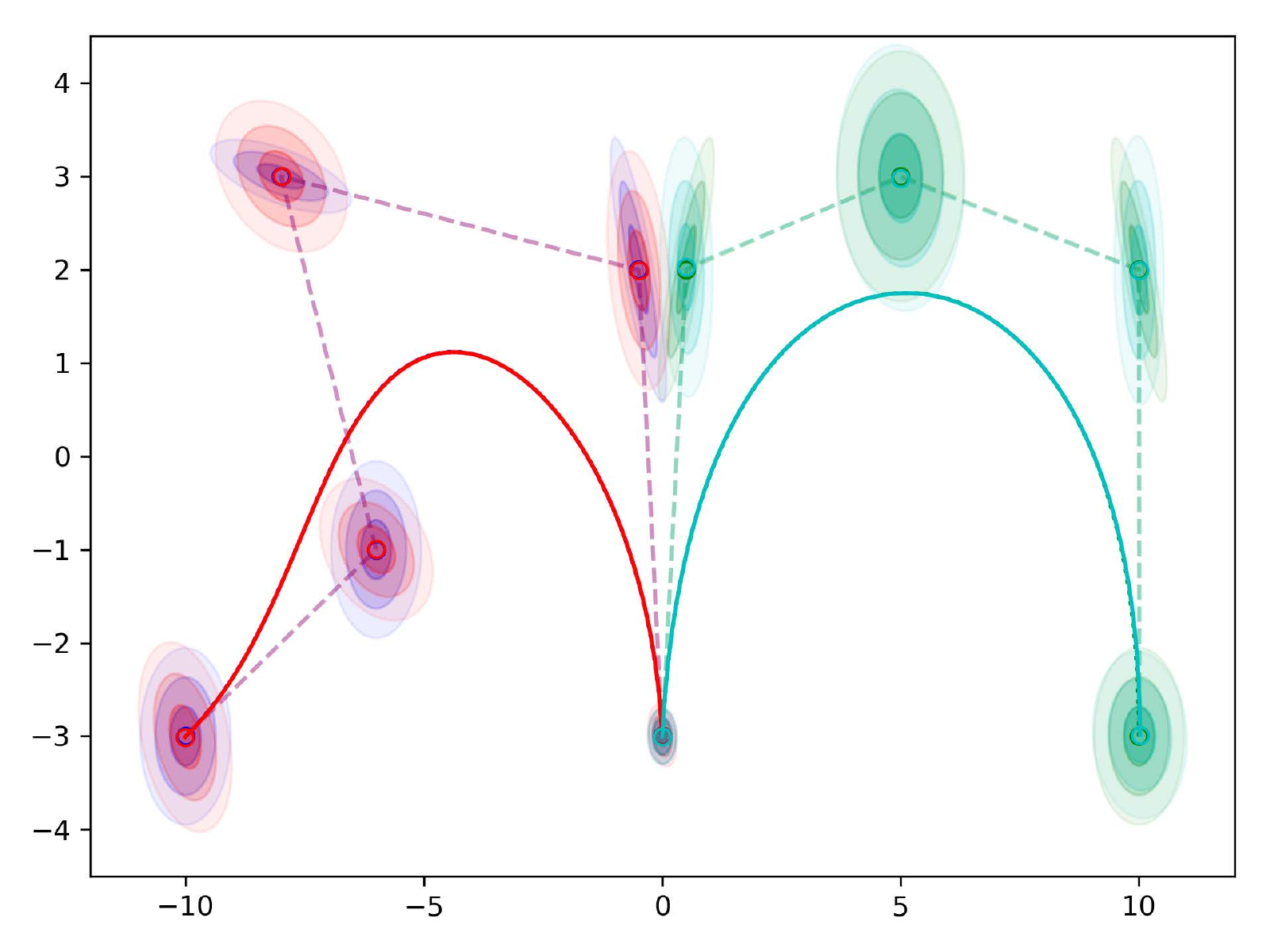} \\
			\small$\pi_{red} = 0.00683$ & \small$\pi_{red} = 0.25$ & \small$\pi_{red} = 0.25$ \\
			\small$\pi_{cyan} = 0.932$ & \small$\pi_{cyan} = 0.75$ & \small$\pi_{cyan} = 0.75$ \\
		\end{tabular}		
	\end{center}
	\caption{Ground truth $\mathcal{N}$-Curves (green and blue) and learned $\mathcal{N}$-Curves (red and cyan) after $0$, $4500$ and $9000$ iterations (left to right) of training.}
	\label{fig:te_learn_cpts_training}
\end{figure}

Figure \ref{fig:te_learn_cpts_training} depicts the initial approximation of $\Psi_{GT}$ and the learned approximations after $4500$ and $9000$ iterations.
After $4500$ iterations, the estimate of the mixing weights is correct, but the mean vectors and covariance matrices are still off.
Finally, after $9000$ iterations, the mean vectors are estimated correctly.
For some control points, the covariance is still not quite correct, which is mostly likely due to the existence of multiple similar solutions, with respect to the resulting curve points, using slightly different variance.

\subsection{Toy example 5: Presence of superfluous mixture components}
\label{ss:te_high_k}
\begin{figure}[htb]
	\begin{center}
		\begin{tabular}{cccc}			
			\includegraphics[width=0.23\textwidth]{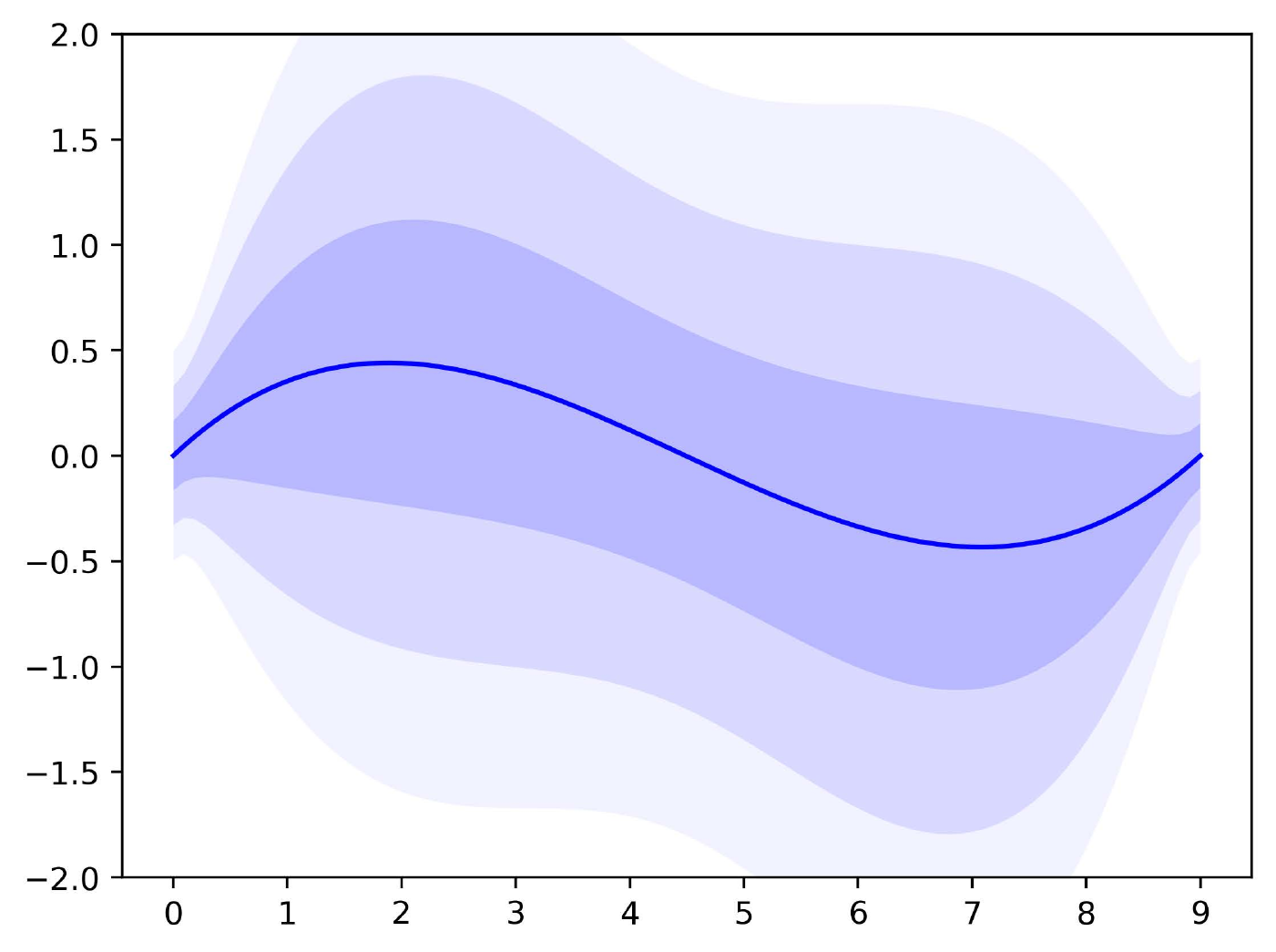}  
			& \includegraphics[width=0.23\textwidth]{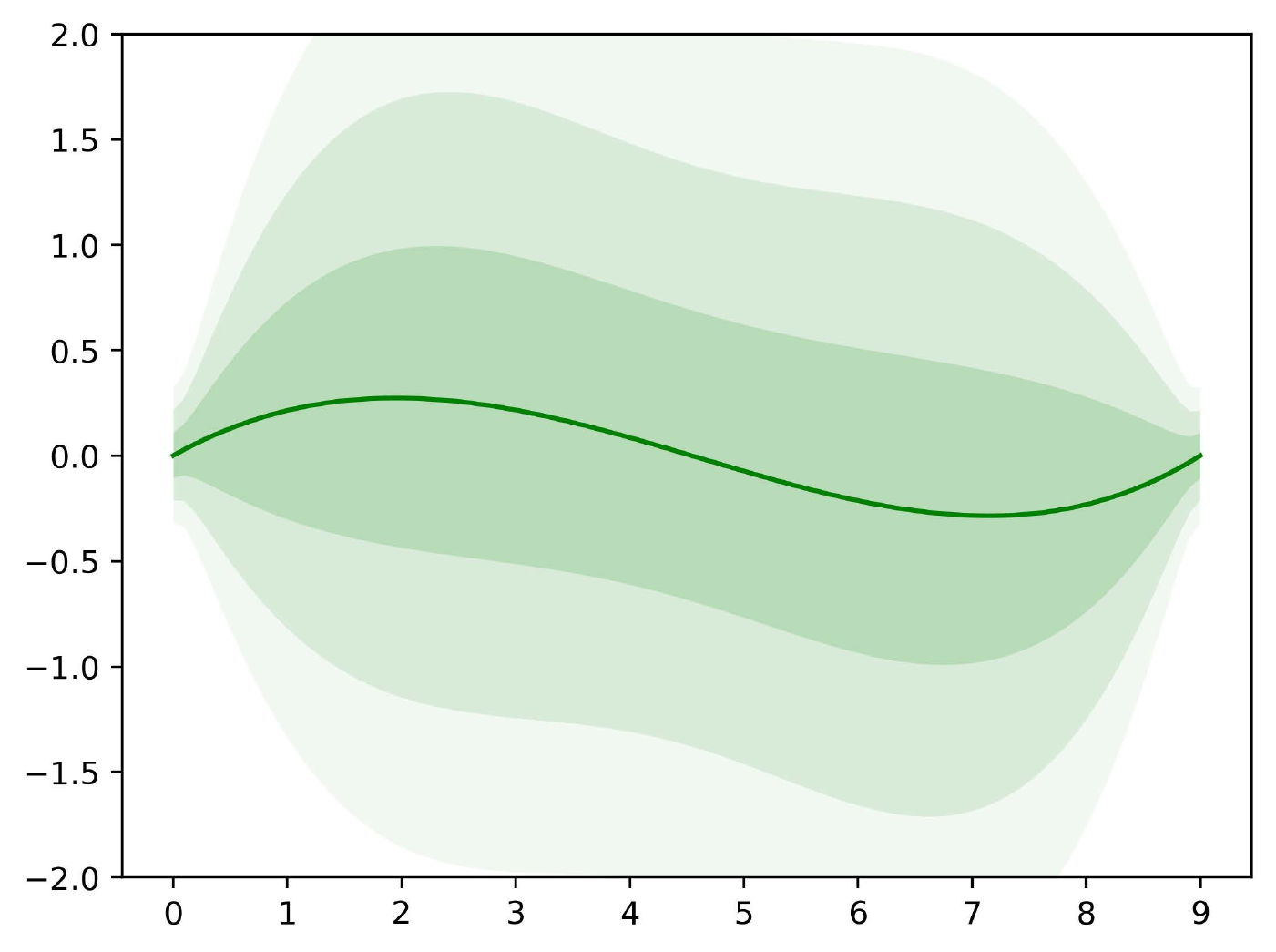} 
			& \includegraphics[width=0.23\textwidth]{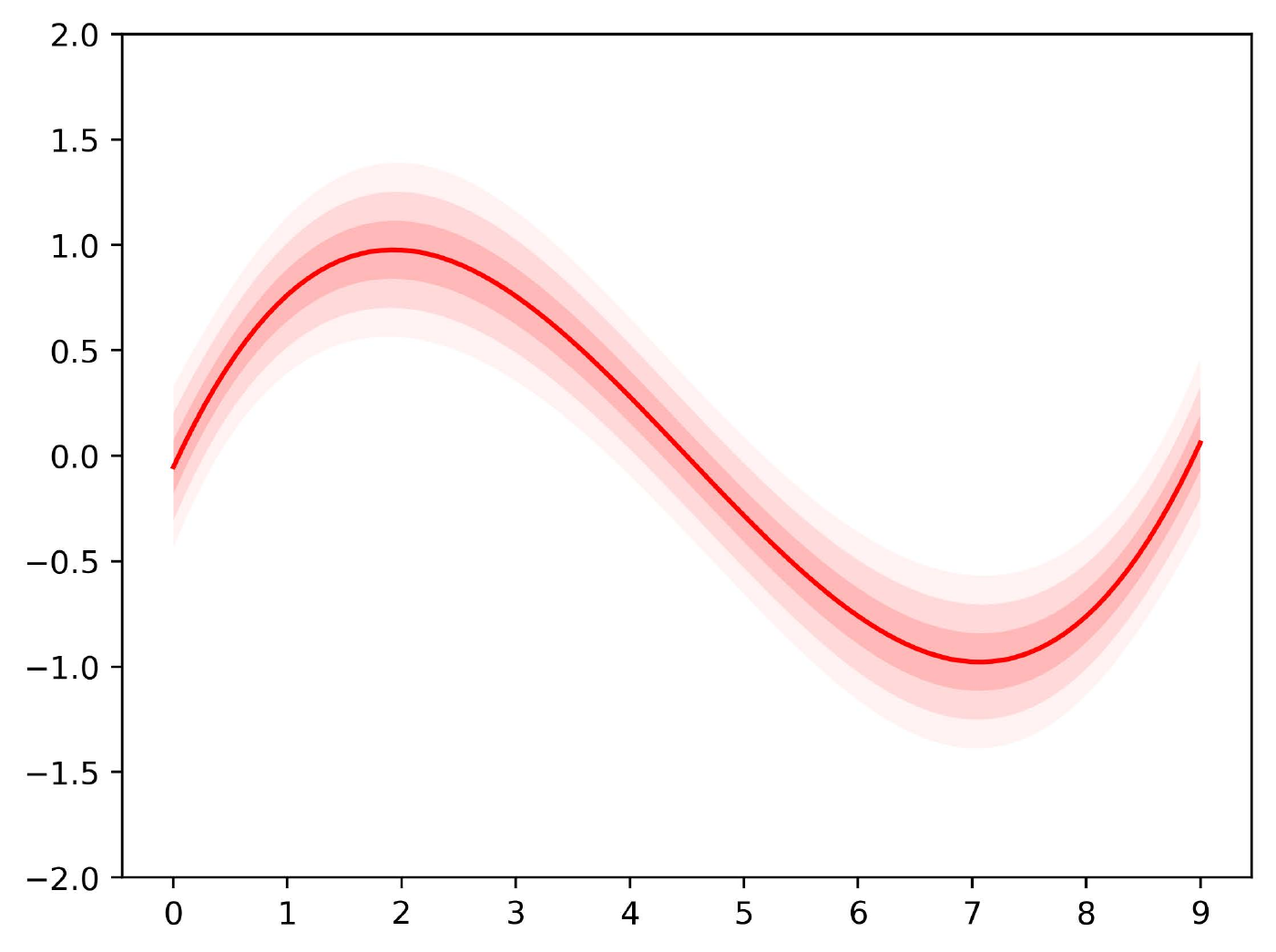} 
			& \includegraphics[width=0.23\textwidth]{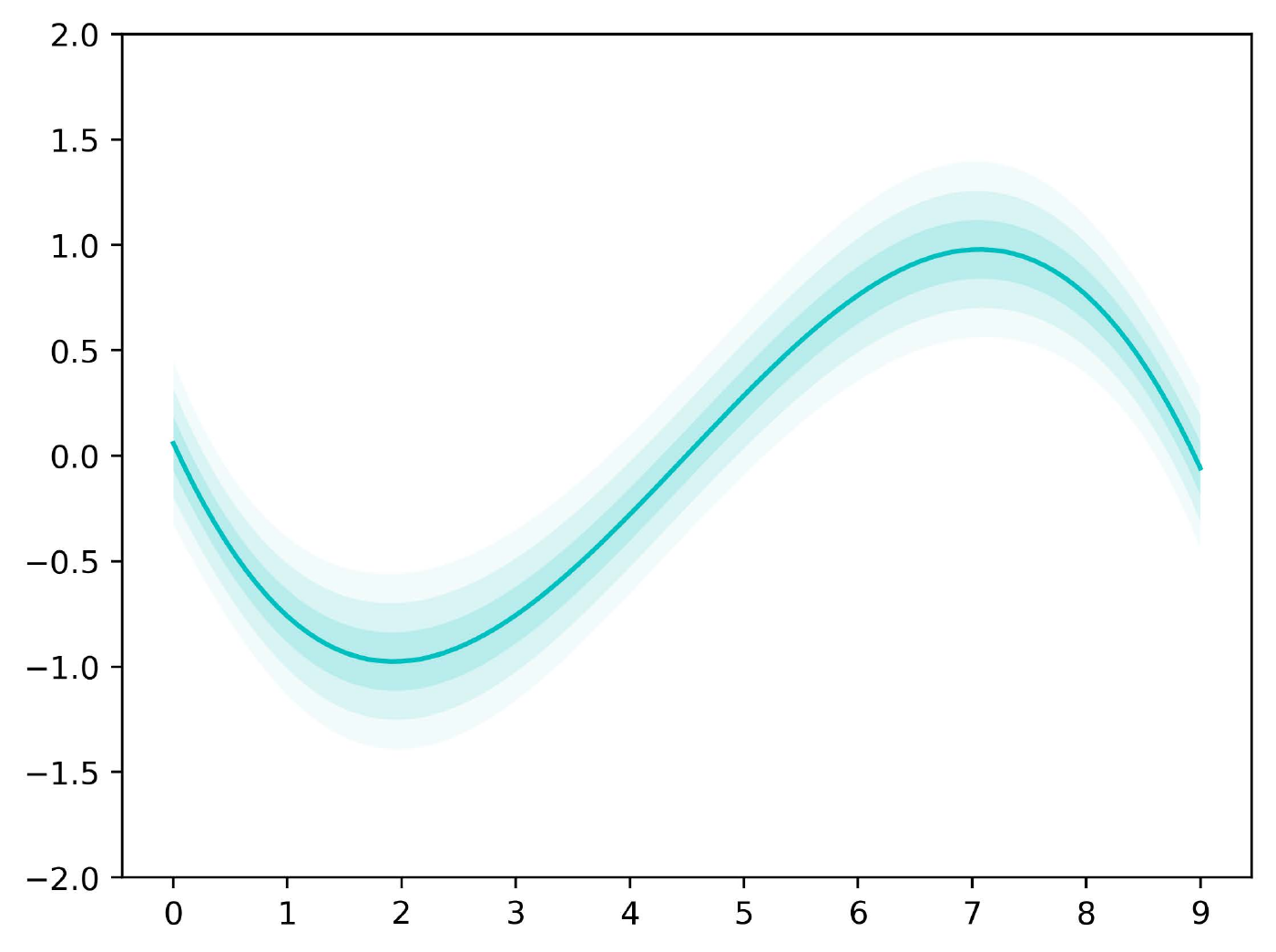}\\ 			
			\small$\pi = 0$ & \small$\pi = 0$ & \small$\pi = 0.128$ & \small$\pi = 0.355$ \\
			
			\includegraphics[width=0.23\textwidth]{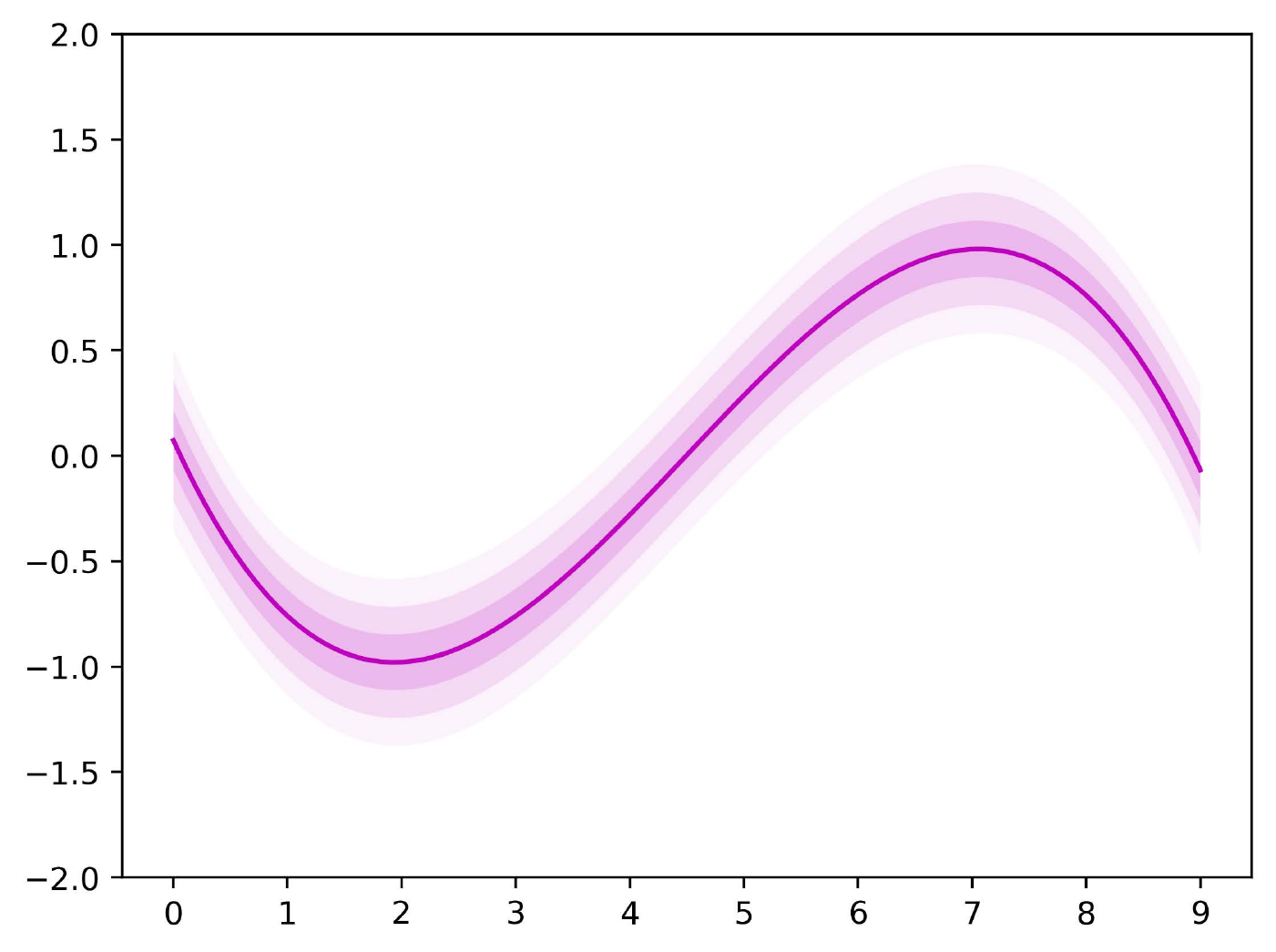}  
			& \includegraphics[width=0.23\textwidth]{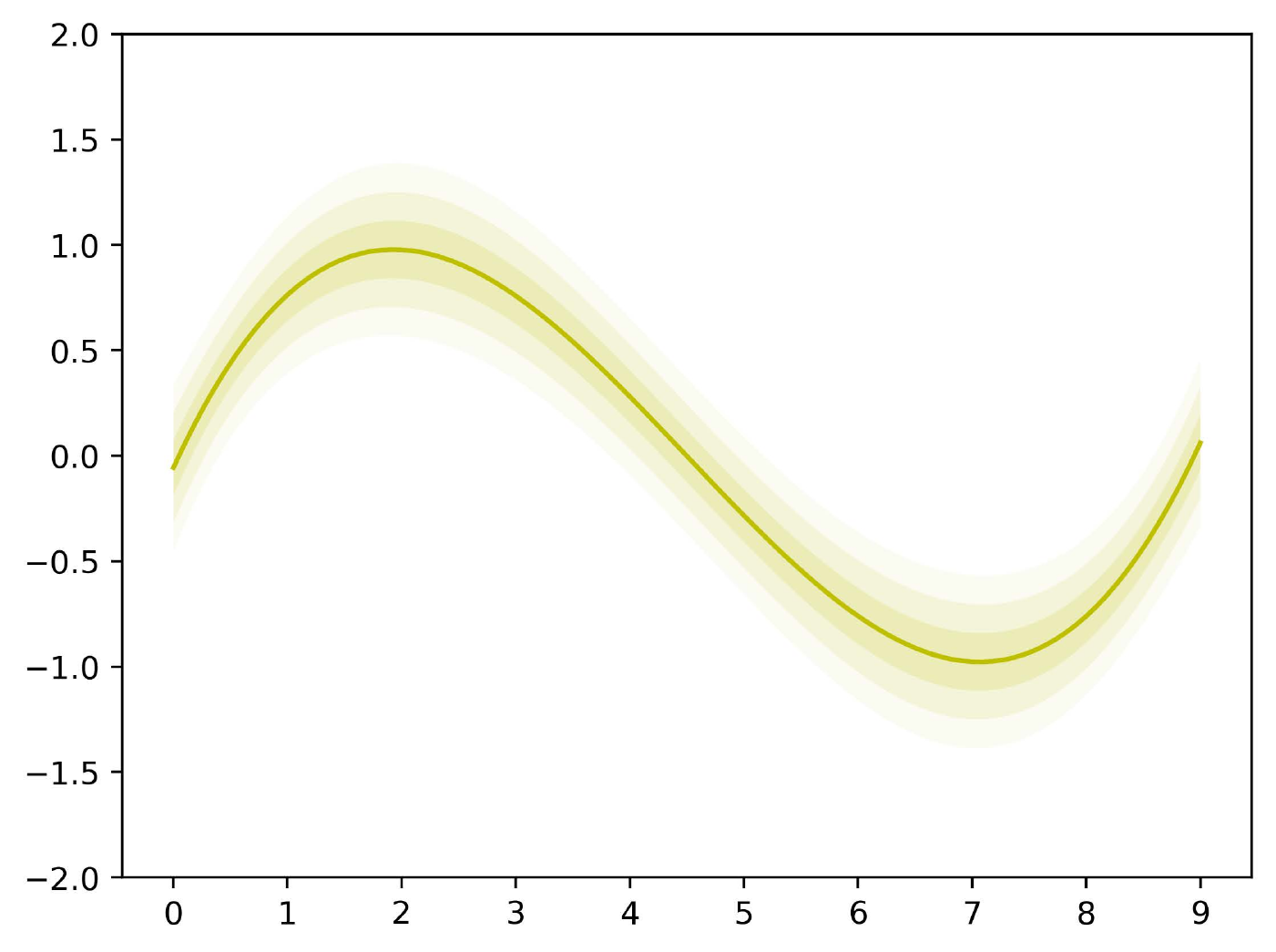} 
			& \includegraphics[width=0.23\textwidth]{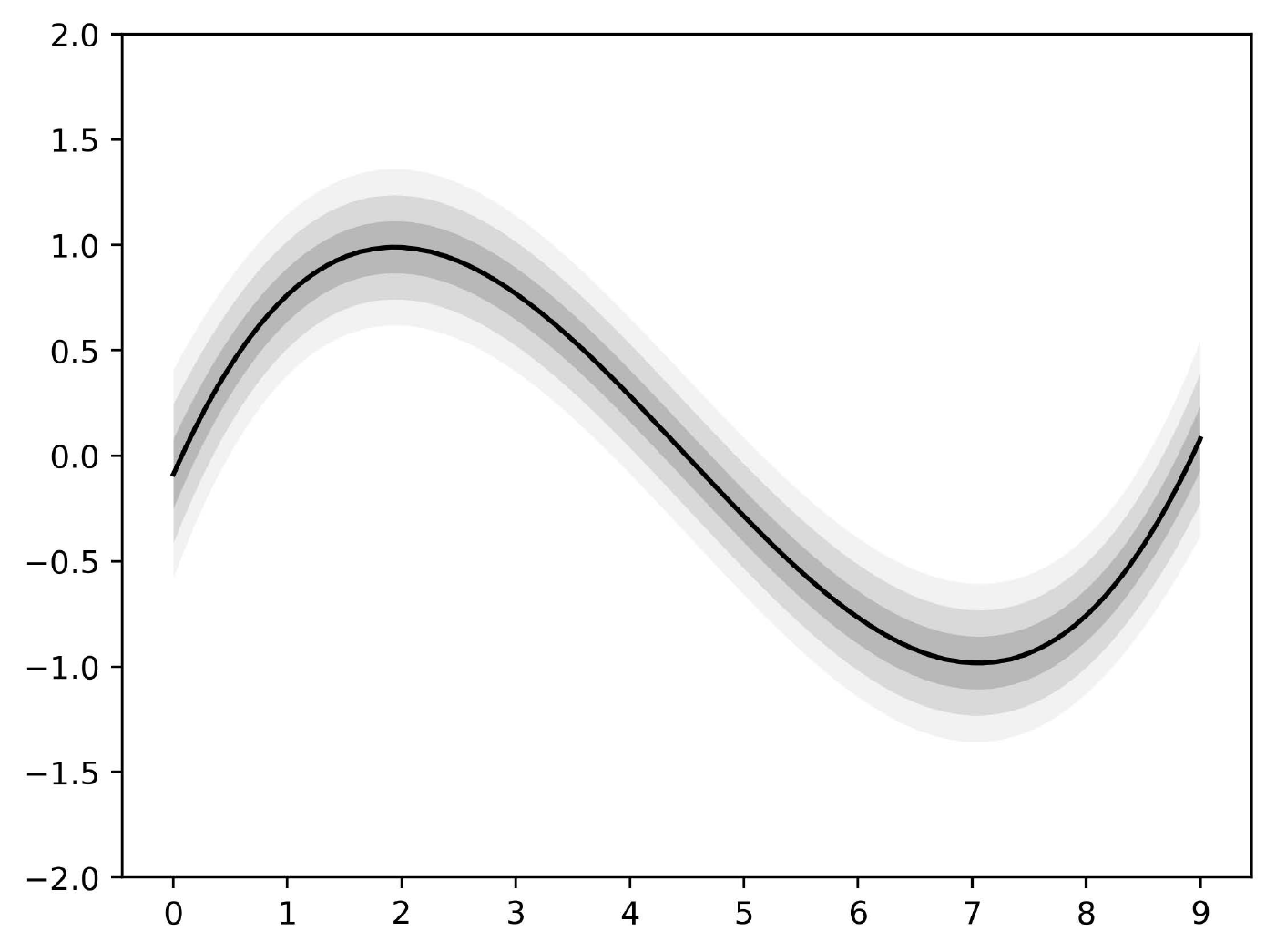} 
			& \\ 			
			\small$\pi = 0.142$ & \small$\pi = 0.357$ & \small$\pi = 0.018$ & \\			
		\end{tabular}
	\end{center}
	\caption{Learned $\mathcal{N}$-Curves for all 7 components in the mixture. Some superfluous components have been suppressed in training (blue and green), while multiple similar $\mathcal{N}$-Curves emerge for each of the two ground truth curves (red/yellow/black and cyan/magenta).}
	\label{fig:te_high_k_single}
\end{figure}

As a follow-up of training related experiments, the behavior of the model in the presence of superfluous mixture components is examined.
Here, the structured example data with $2$ curves from the multi-mode toy example (section \ref{sss:multi_te}) is used to train an $\mathcal{N}$-Curve mixture model with $k=7$ components, i.e. $5$ superfluous components.
Preferably, in the resulting model, the mixing weights of all $5$ unnecessary components are set to $0$ and the remaining $\mathcal{N}$-Curves model the two alternatives in the data and are weighted equally ($\pi = 0.5$).
Figure \ref{fig:te_high_k_single} depicts all $7$ $\mathcal{N}$-Curves of the mixture and corresponding mixing weights after training.
It can be seen, that only the weights of $2$ components have been set to zero and $5$ components are used to model the data.
The components shown in \emph{red}, \emph{yellow} and \emph{black} describe one curve and the \emph{cyan} and \emph{magenta} colored components the other one.
Still, the sum of weights for each curve in the data is approximately equal to $0.5$ ($0.503$ and $0.497$) and the variances of these components is consistent.
As a consequence, this behavior could be tackled by for example incorporating a post-processing step that collapses (nearly) identical components/modes into a single one.

\begin{figure}[htb]
	\begin{center}
		\includegraphics[width=0.4\textwidth]{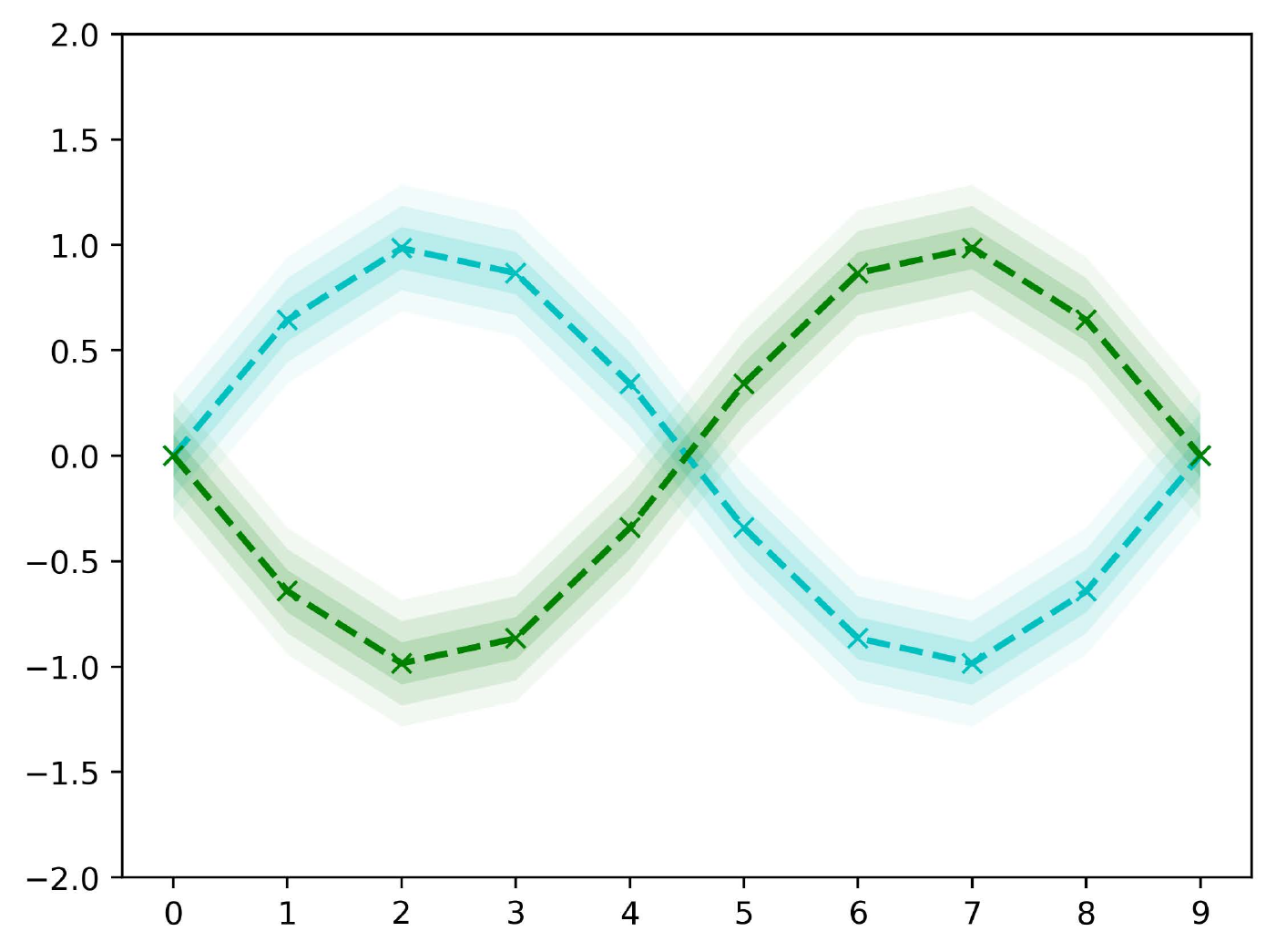}
		\includegraphics[width=0.4\textwidth]{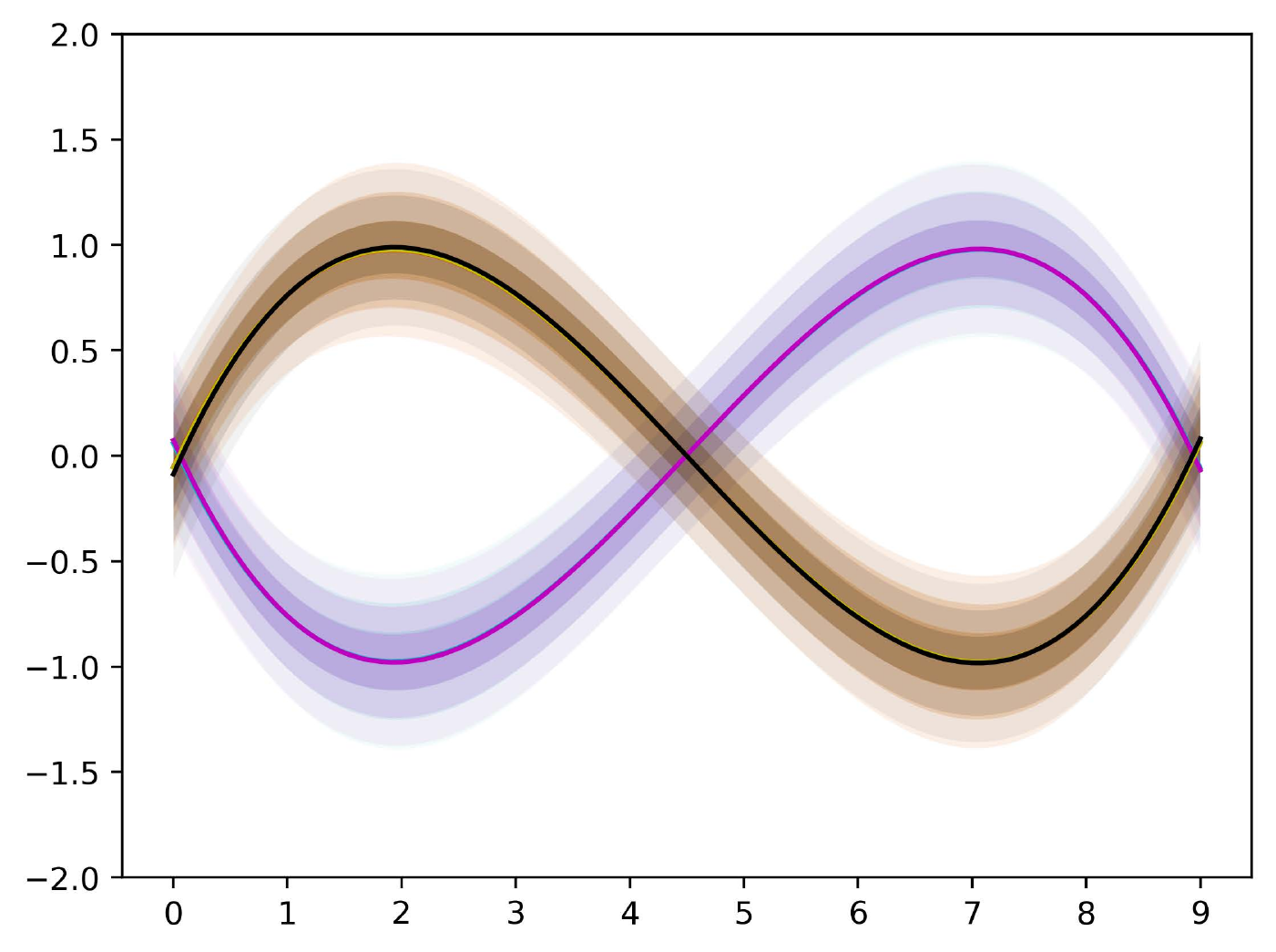}
	\end{center}
	\caption{Stochastic process modeling two curves (left) alongside overlayed non-zero components of the learned $\mathcal{N}$-Curve mixture model (right).}
	\label{fig:te_high_k_res}
\end{figure}

The left image in figure \ref{fig:te_high_k_res} shows the ground truth mean and standard deviations along both curves and the right image the approximation given an $\mathcal{N}$-Curve mixture model using $k=7$ components.
Here, components with weights driven towards zero are not shown.
Again, it can be seen that due to the structure in the training data, the $\mathcal{N}$-Curve mixture model is capable of approximating the ground truth distributions.
Further, overlaying the components shows that these are close to identical (right image in figure \ref{fig:te_high_k_res}).

\section{Real world evaluation}
\label{sec:experiments}
In this section, the performance of the proposed $\mathcal{N}$-Curve mixture model is compared to different state-of-the-art models in the task of sequence prediction.
Following this, a qualitative evaluation is performed, inspecting results of the prediction evaluation and re-viewing different aspects of the model shown in previous sections through toy examples.

\subsection{Quantitative evaluation}
This quantitative evaluation tests the performance of the proposed model and state-of-the-art models in two $n$-step sequence prediction tasks: trajectory prediction and human motion sequences.
These tasks have been chosen to allow comparison with the respective models considered for each task.
In both tasks, evaluated models need to represent a stochastic process describing $m + n$ time steps ($\|T_*\| = m + n$), such that given $m$ observations of a realization of the process, the remaining $n$ steps can be inferred.
In the case of trajectory prediction, observation and prediction sequences consist of subsequent 2D coordinates.
For human motion modeling, these sequences consist of 59-dimensional skeleton description vectors.
According to \cite{mattos2015recurrent}, the observation sequence additionally incorporates the y coordinate of the left toes for all $m + n$ time steps as a control input.

\subsubsection{Trajectory prediction.}
\label{subsec:traj_pred}
\begin{figure}[b]  
	\begin{center}
		\includegraphics[width=0.2675\textwidth]{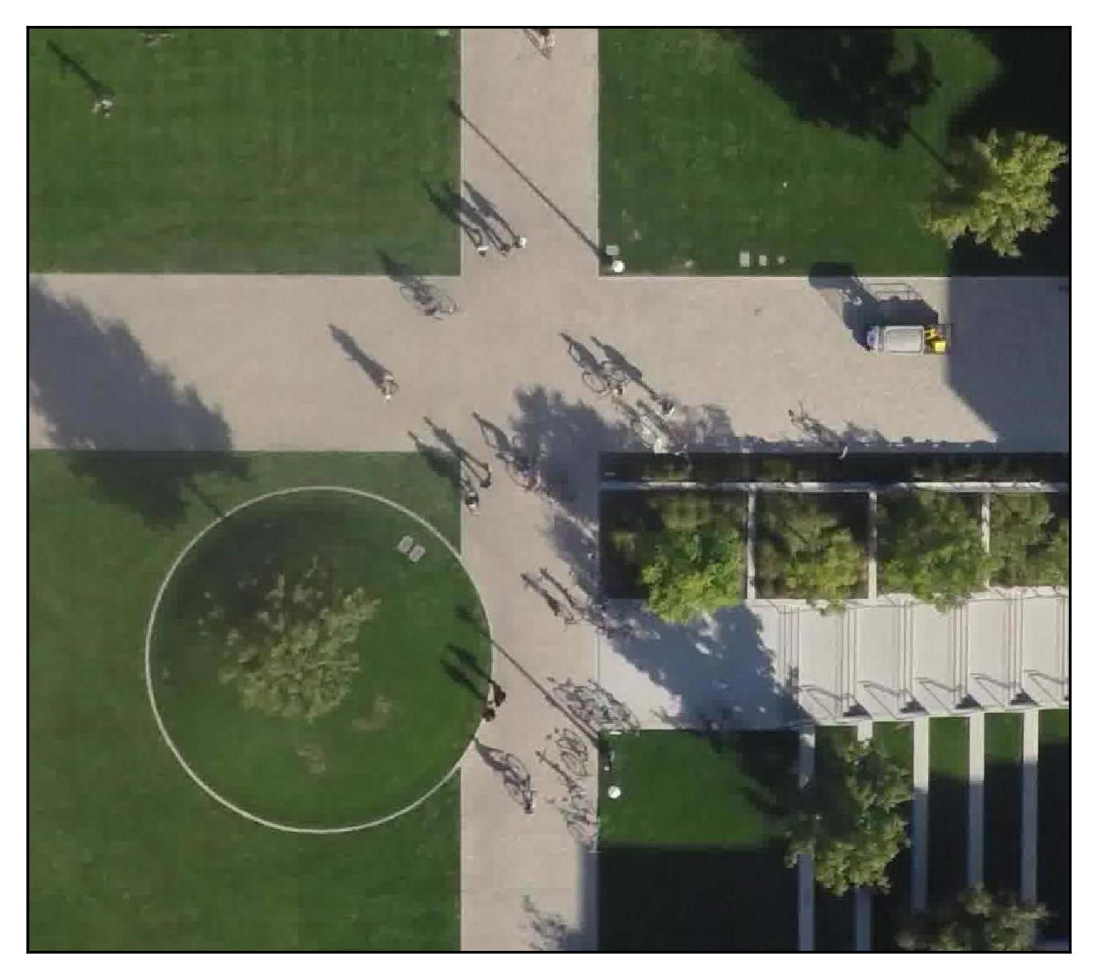}
		\includegraphics[width=0.2675\textwidth]{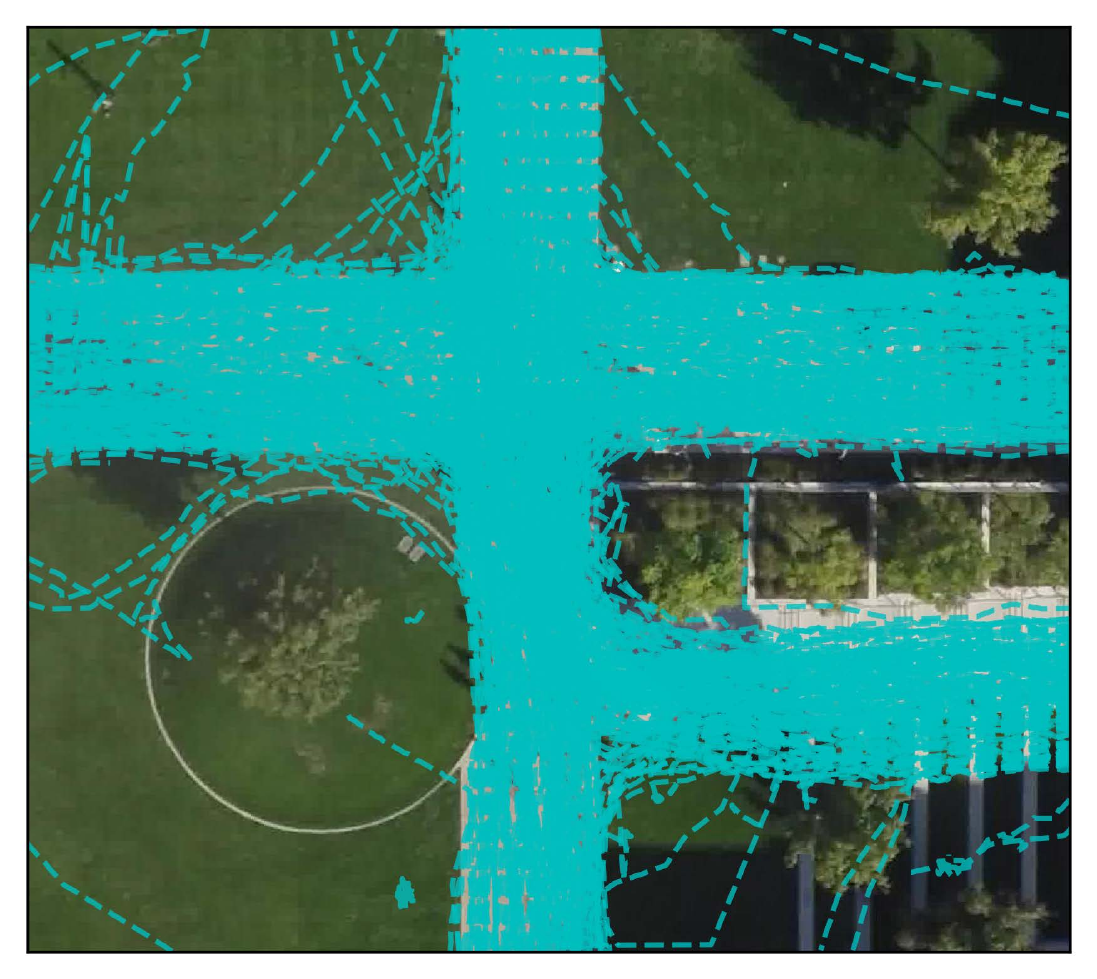}
		\includegraphics[width=0.2\textwidth]{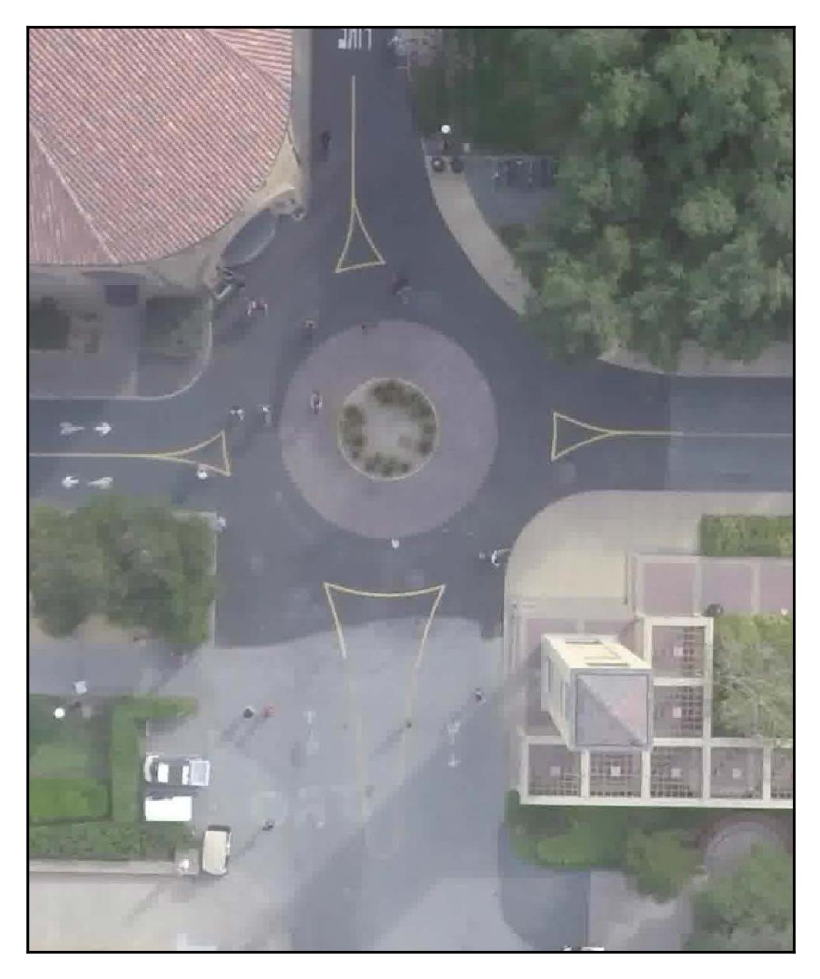}
		\includegraphics[width=0.2\textwidth]{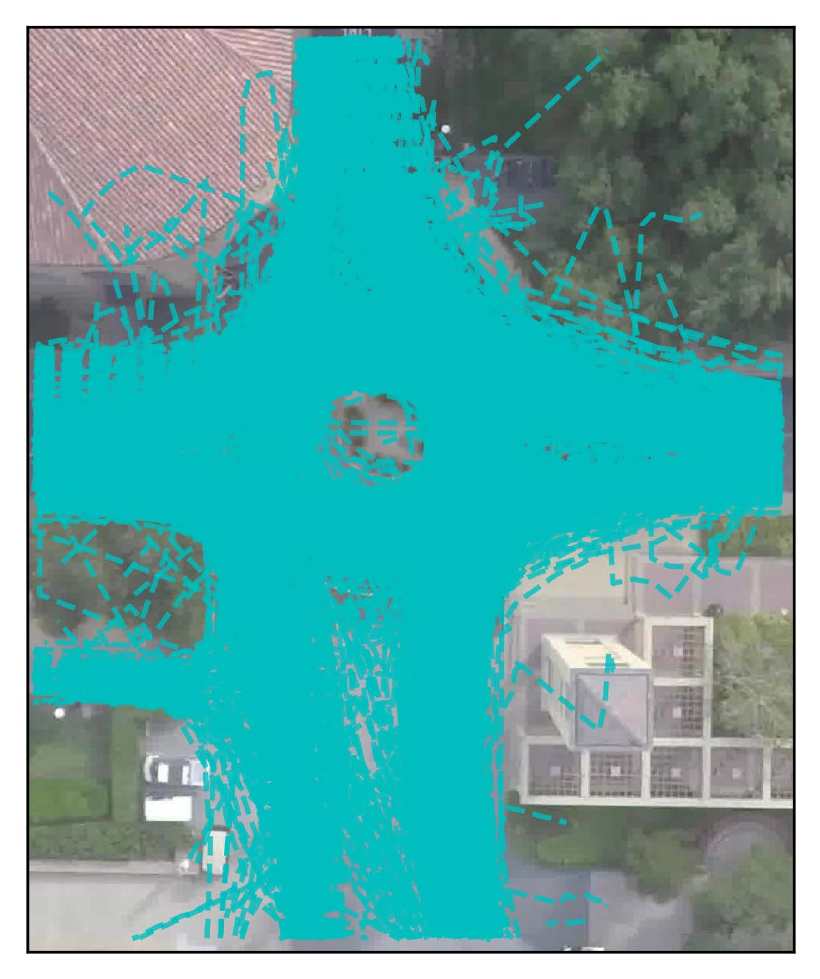}
	\end{center}
	\caption{Images of the Stanford Drone Dataset\cite{robicquet2016learning} scenes \emph{hyang} (left) and \emph{deathcircle} (right) with and without trajectories.}
	\label{fig:exp_tr_pred_data}
\end{figure}

An exemplary use case for multi-modal path prediction are smart infrastructure or related video surveillance applications.
Path prediction in these applications is mainly done in image coordinates, where tracklets recorded from a static camera are used as observations. 
To ensure that multi-modal sequence modeling is necessary (due to the scene geometry including for example junctions), trajectories from the Stanford Drone Dataset \cite{robicquet2016learning} are used.
According to \cite{hug2018particle} sequences from the scenes \emph{hyang} and \emph{deathcircle} from multiple recordings of the same scene are used.
Here, annotations are combined into a single coordinate system in order to increase the overall amount of trajectories.
Following this, the preprocessed \emph{hyang} scene consists of 613 pedestrian trajectories and the \emph{deathcircle} scene consists of 1447 biker trajectories.
The datasets are illustrated in figure \ref{fig:exp_tr_pred_data}.
For \emph{deathcircle}, biker trajectories are used, as there are more bikers than pedestrians in the scene and biker trajectories give more inspectable results as these are more likely to follow the roundabout in the scene.
Further, image information is omitted and both datasets are scaled (individually) such that each trajectory point lies in $[-1, 1]$.
The sampling rate has been reduced to 6 Hz.
Observation and prediction lengths are fixed to $m = 20$ (3.2 seconds) and $n = 40$ (6.6 seconds) respectively.
In this experiment, the proposed $\mathcal{N}$-Curve mixture model is compared to two recent approaches for multi-modal trajectory prediction: The Particle LSTM-MDN model by \cite{hug2018particle} and the "Best of Many Samples" LSTM model (abbrev.: LSTM-BMS) by \cite{bhattacharyya2018accurate}.
Being two of the few approaches actually focusing on multi-modal trajectory prediction, these approaches are chosen for comparison (see also section \ref{ss:rw_multi}).
The former embeds an LSTM network combined with an MDN into a particle filter cycle in order to produce $n$-step multi-modal predictions, while the latter utilizes an LSTM network to produce multiple sample predictions and is learned using a "Best of Many Samples" objective function, yielding diverse samples.

For evaluation, a subset of 200 randomly selected tracklets (of length 60) have been used.
The results are reported in terms of the \emph{Final Displacement Error} (\emph{FDE}), being the RMSE between the endpoints of the ground truth trajectories and the endpoints of the most likely prediction (maximum likelihood estimate). 
The \emph{Negative Log-Likelihood} (\emph{NLL}, see equation \ref{eq:logloss}) provides a measure which incorporates the multi-modal probabilistic prediction generated by each model.
In case of the $\mathcal{N}$-Curve mixture model, $k=3$ mixture components with $6$ Gaussian control points per component are used.
The observation sequence is encoded using an LSTM network in order to produce the $\mathcal{N}$-Curve MDN input vector $\mathcal{V}$.
For calculating the FDE, the endpoint of the $\mathcal{N}$-Curve with the highest prior probability $arg max_k~\pi_k$ is used.
In case of the LSTM-BMS model, $100$ sample predictions were generated, which are then clustered using k-means, with $k=3$ centroids (code provided by the authors\footnote{\url{https://github.com/apratimbhattacharyya18/CGM_BestOfMany}}).
Clusters have been weighted according to the number of trajectories in each cluster.
For the FDE the endpoint of the cluster centroid with the highest weight is used.
For the NLL, the centroid trajectory points represent the mean values for each time step and the covariance is calculated using the trajectories in each respective cluster.
The same procedure is applied for the Particle LSTM-MDN model using $100$ particles.

\begin{table}
	\centering
	\begin{tabular}{l@{\hskip 12pt}c@{\hskip 12pt}c@{\hskip 12pt}c}
		\hline
		 & Particle LSTM-MDN\cite{hug2018particle} & LSTM-BMS\cite{bhattacharyya2018accurate} & $\mathcal{N}$-Curve MDN \\
		\hline
		Hyang & 0.129 (18.552) & 0.147 (-2.574) & \textbf{0.088} (\textbf{-4.760}) \\
		Deathcircle & 0.366 (15.028) & 0.483 (0.389) & \textbf{0.314} (\textbf{-3.617}) \\
		\hline
	\end{tabular}
	\caption{FDE (and NLL) values for evaluated models on \emph{hyang} and \emph{deathcircle} scenes taken from the Stanford Drone Dataset\cite{robicquet2016learning}.}
	\label{tab:traj_pred}
\end{table}
The results are reported in table \ref{tab:traj_pred}.
It can be seen that the $\mathcal{N}$-Curve mixture model performs best in terms of FDE and NLL among the provided models.
Larger NLL values for the Particle LSTM-MDN model are most likely due to the particle filter collapsing onto few particles in regions with small variation, leading to small variances and thus higher NLL values.

\subsubsection{Human motion modeling.} 
\label{subsec:motion_modeling}
For providing a high dimensional (59 in this case) sequence prediction example, sequences from the CMU motion capture database\footnote{\url{http://mocap.cs.cmu.edu/}} are used.
According to \cite{mattos2015recurrent}, training is performed on sequences 1 to 4 from subject 35, and testing is performed on sequences 5 to 8 from the same subject.
In order to make results comparable to \cite{mattos2015recurrent} with the code provided by the authors\footnote{\url{https://github.com/zhenwendai/RGP}}, only walking motion is considered.
The test set has been modified to only contain the first 70 points of each sequence, in order to conform with the fixed sequence length currently necessary for the $\mathcal{N}$-Curve mixture model.
Further, the data is standardized with zero mean and unitary standard deviation.
In contrast to the previous experiment, a control input is given in terms of the y coordinate of the left toes for each time step (during observation and prediction).
Here, the observation and prediction lengths are set to $m = 20$ and $n = 50$ time steps respectively.
In this experiment, the $\mathcal{N}$-Curve mixture model is compared to a simple multilayer perceptron (\emph{MLP}) and the recurrent Gaussian process model (\emph{RGP}) introduced in \cite{mattos2015recurrent}.

For the evaluation on the test set, the RMSE over all predicted values in the sequence is reported.
Again, for the $\mathcal{N}$-Curve mixture model, the observation is encoded using an LSTM network and the number of components is set to $k=1$ using $10$ Gaussian control points.
For the MLP, a single hidden layer with 1000 units and $tanh$ activation is used.
This MLP directly maps the concatenated observation sequence combined with the control sequence onto the prediction sequence.
In case of the RGP model, a 2 hidden layer model with 200 inducing points is used according to the evaluation performed in \cite{mattos2015recurrent}.

\begin{table}
	\centering
	\begin{tabular}{c@{\hskip 12pt}c@{\hskip 12pt}c}
		\hline
		MLP & RGP\cite{mattos2015recurrent} & $\mathcal{N}$-Curve MDN \\
		\hline
		0.911 & 0.822 & \textbf{0.794} \\
		\hline
	\end{tabular}
	\caption{RMSE values for different models on sequences taken from the CMU motion capture database.}
	\label{tab:motion_modeling}
\end{table}
The results are reported in table \ref{tab:motion_modeling}.
Again, the $\mathcal{N}$-Curve mixture model performs best in terms of RMSE among the provided models. 

\subsection{Qualitative evaluation}
\begin{figure}[t]  
	\begin{center}
		\includegraphics[width=0.355\textwidth]{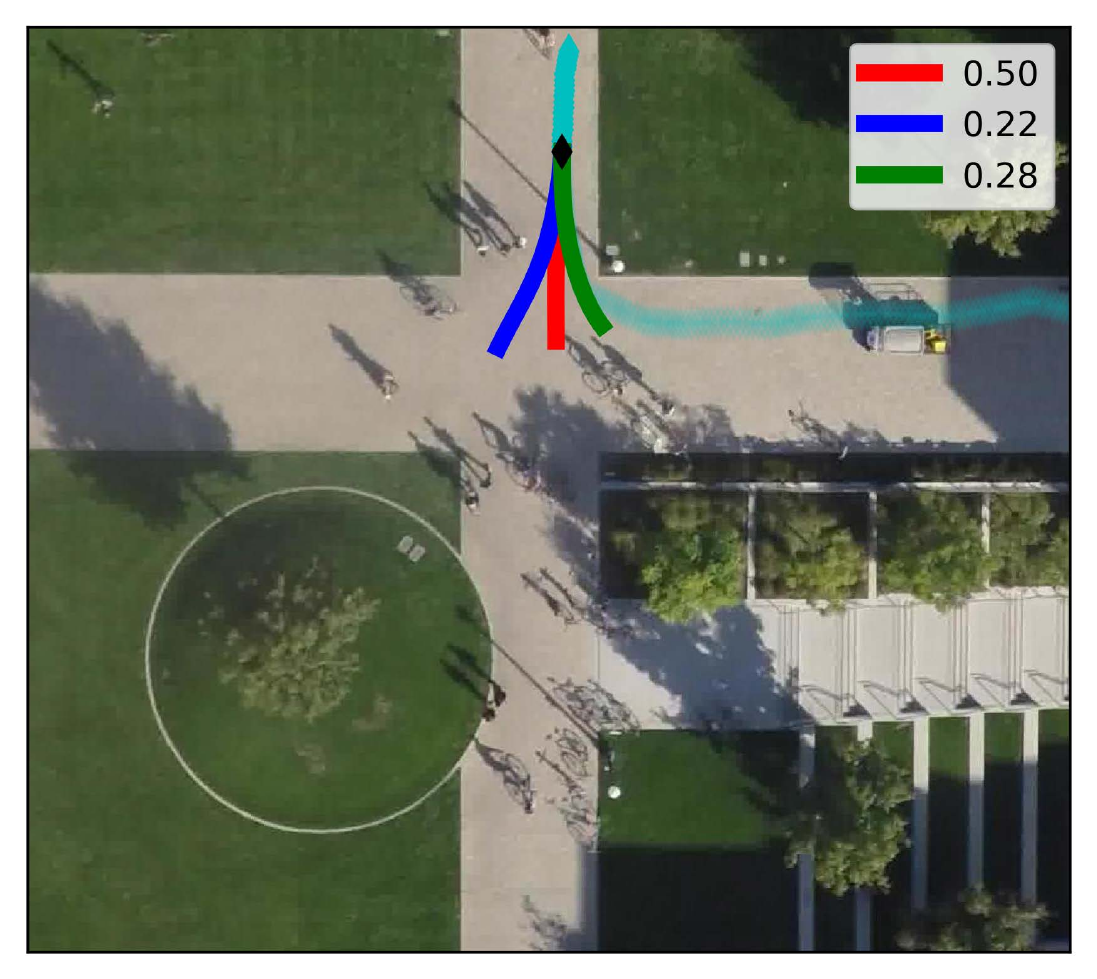}
		\includegraphics[width=0.355\textwidth]{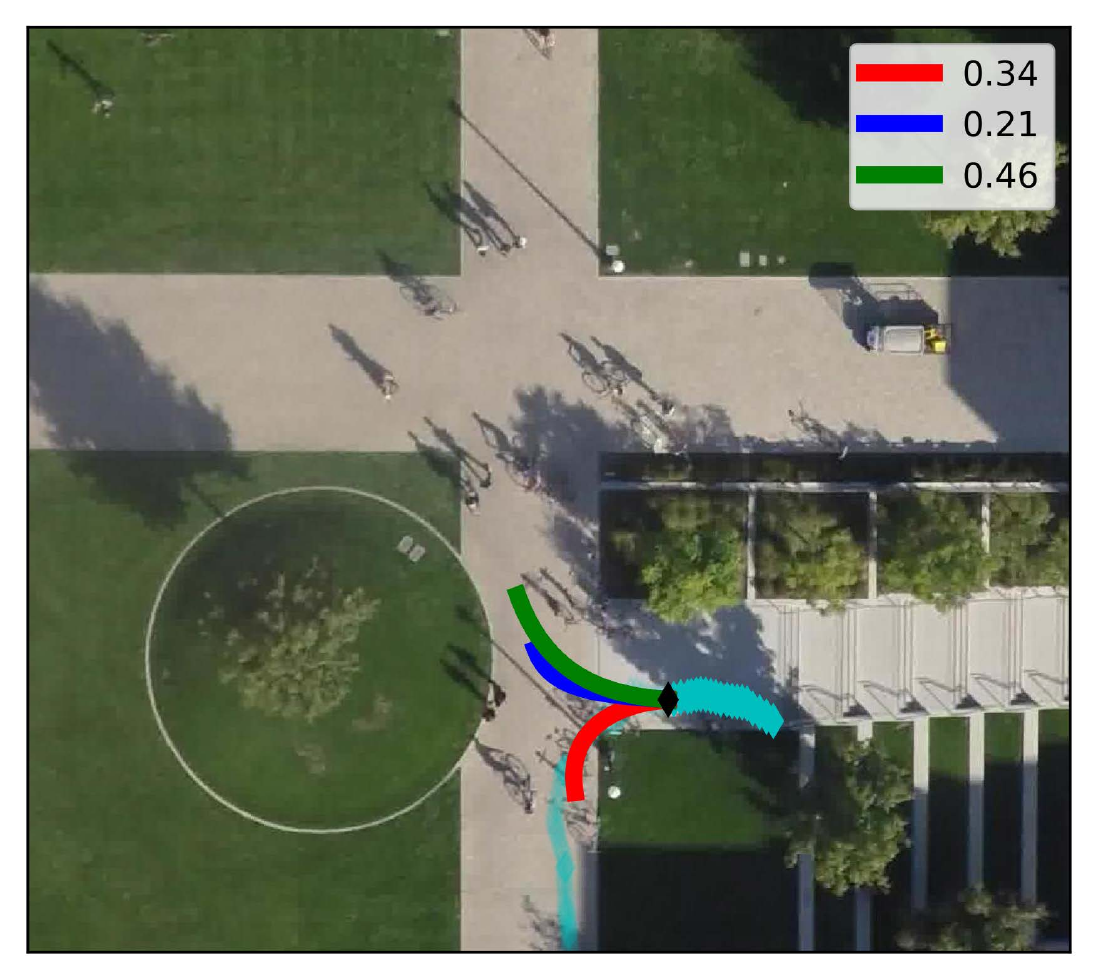}  	
		\includegraphics[width=0.26\textwidth]{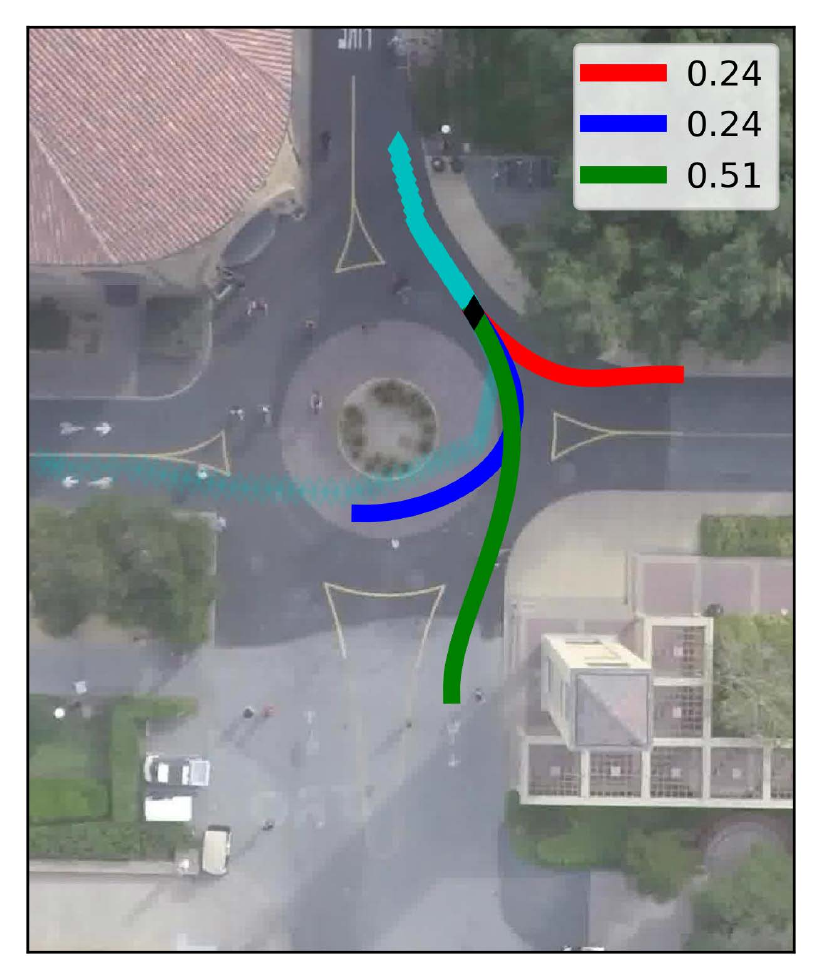}	
		
		\includegraphics[width=0.325\textwidth]{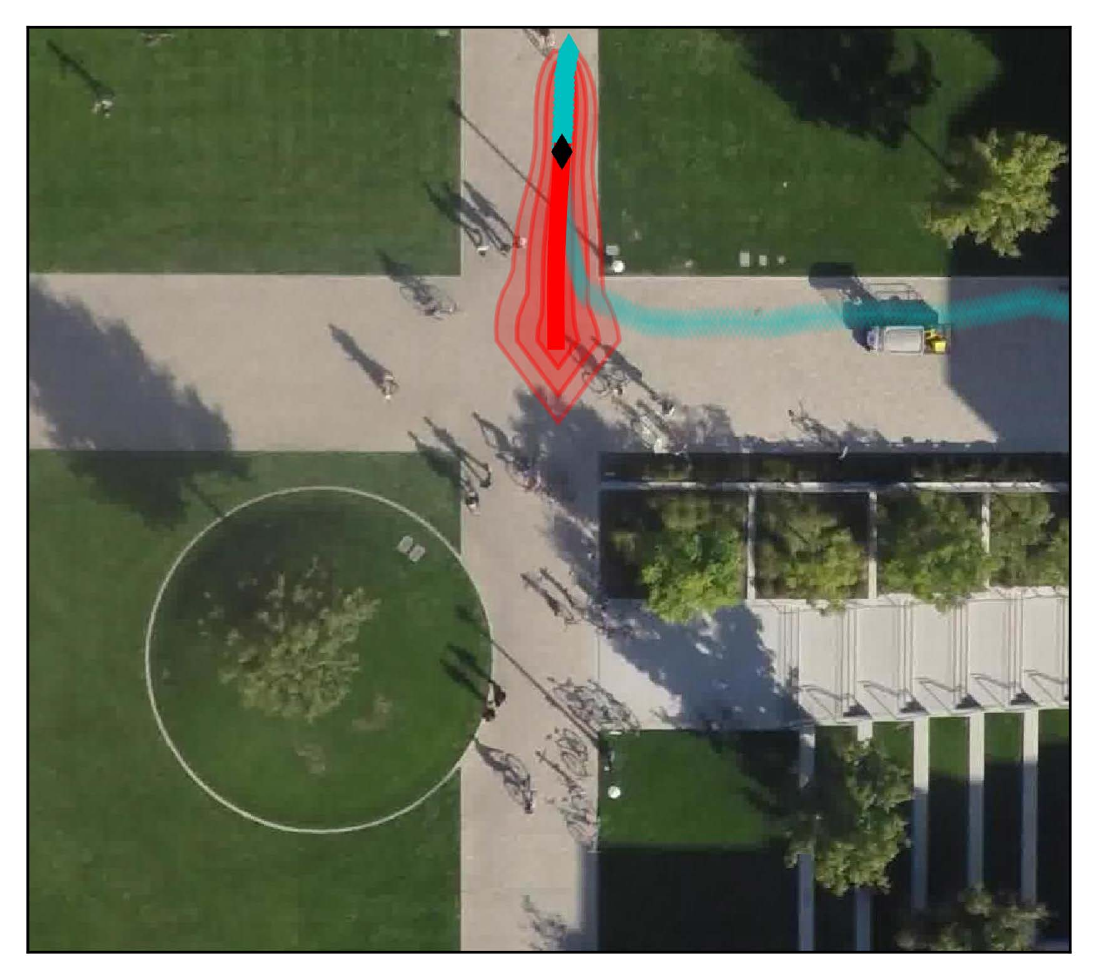}
		\includegraphics[width=0.325\textwidth]{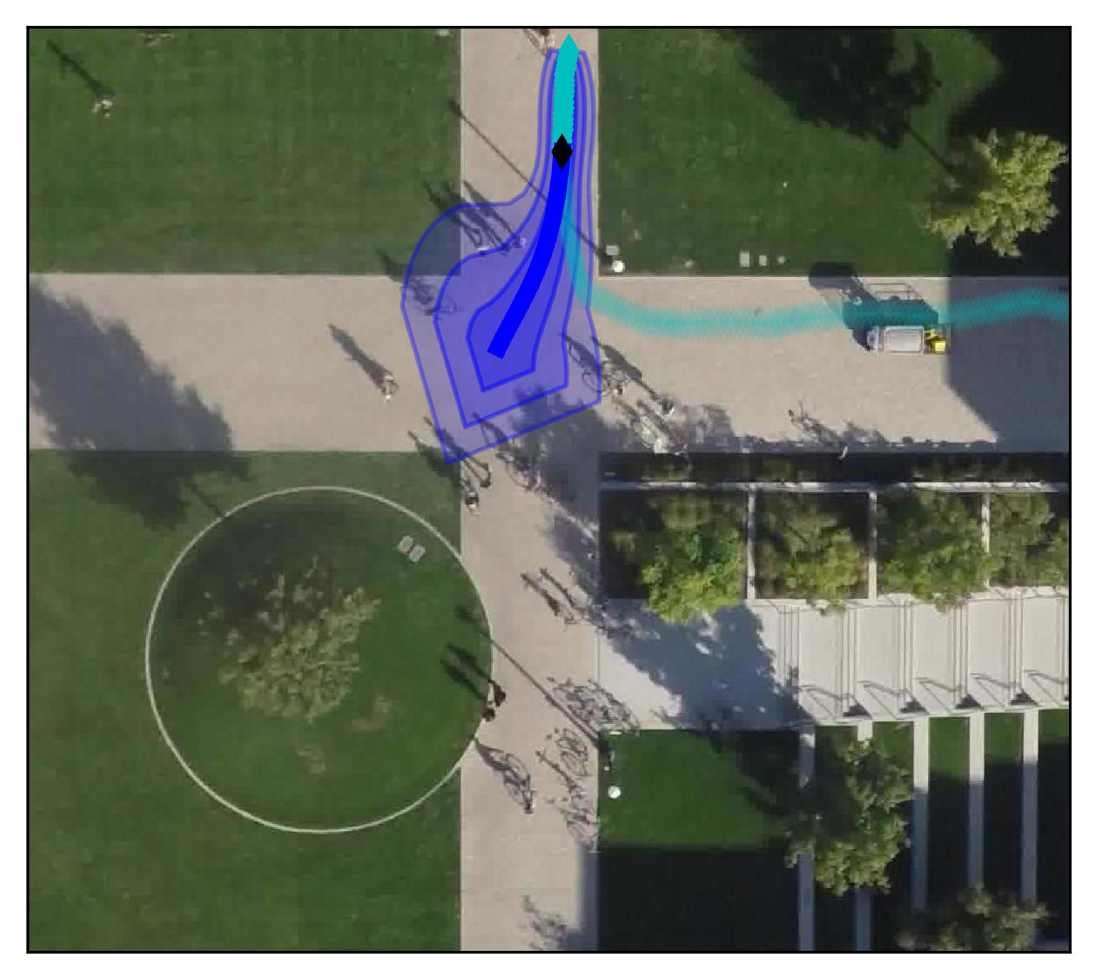}
		\includegraphics[width=0.325\textwidth]{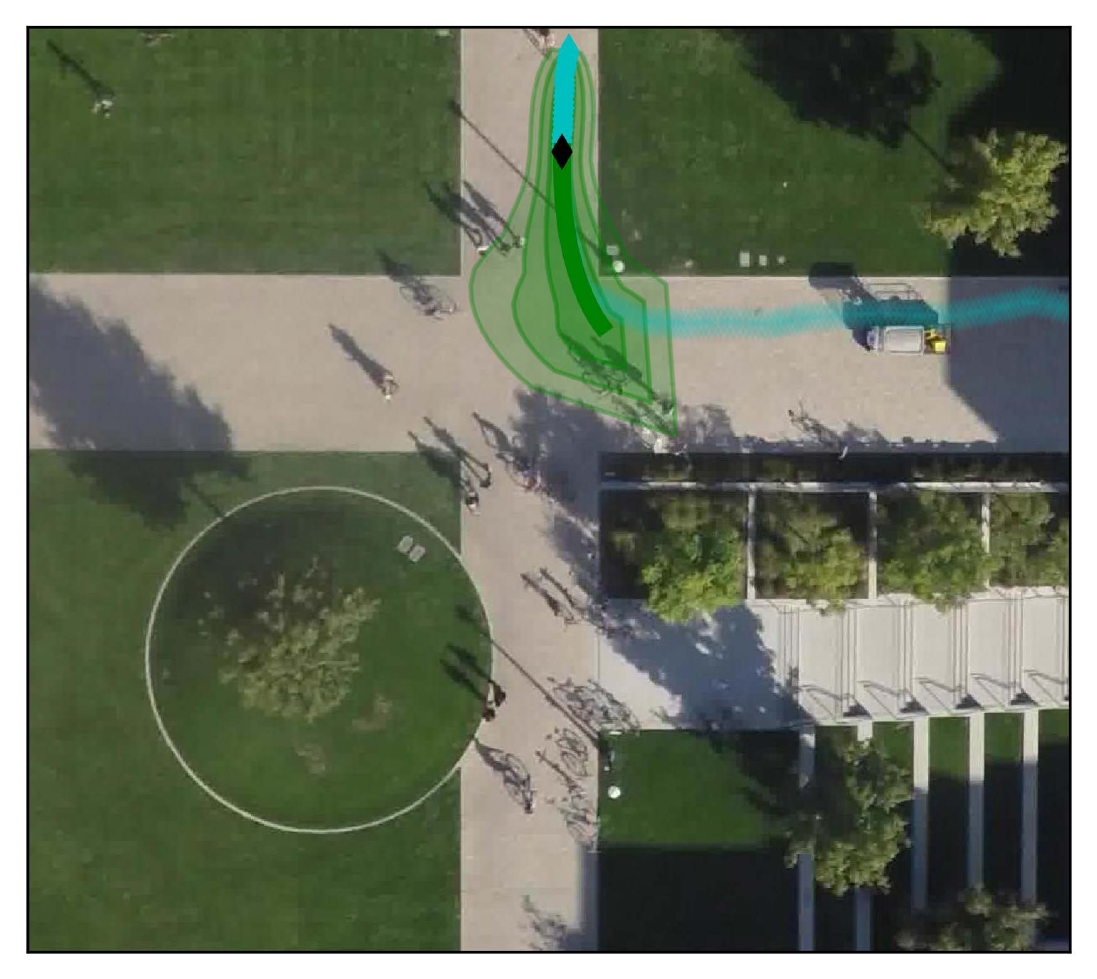}
	\end{center}
	\caption{First row: Prediction generated by a 3-component $\mathcal{N}$-Curve mixture model for \emph{hyang} (left and center) and \emph{deathcircle} (right) scenes taken from the Stanford Drone Dataset.
		Observation sequence is shown in (saturated) cyan, ground truth trajectory in transparent cyan and predictions in red, green and blue.
		Predicted path probabilities are indicated in the image legend.
		Second row: Individual components of the first prediction examples with variance.}
	\label{fig:examples_tr_pred}
\end{figure}

Before discussing different aspects of the model shown in toy examples conducted in previous sections, some qualitative results for both, trajectory prediction and human motion modeling, are demonstrated.
Starting with trajectory prediction, some examples generated by the $\mathcal{N}$-Curve mixture model are depicted in figure \ref{fig:examples_tr_pred}.
The examples show, that the model produces diverse predictions, following different possible paths through the scene with respective probabilities.
In case of the \emph{hyang} scene, different tracklets just before an intersection show possible predictions in respective directions given by the pathways.
For \emph{deathcircle}, a tracklet entering the roundabout is given, thus predictions leaving at all possible exits are generated.
The images in the second row of figure \ref{fig:examples_tr_pred} depict the individual components of the $\mathcal{N}$-Curve mixture model for the first prediction example.
It should be noted, that the trained $\mathcal{N}$-Curve mixture model represents the entire $m + n$ step sequence, thus the variance for the observation sequence can also be shown.
As changing direction in that example (blue and green paths) are less likely, the variance increases towards the end of the sequence, overlapping with the straight path, increasing its probability.
This approximately also matches the data: When calculating path probabilities using similar trajectories from the dataset, $15\%$ follow the blue path, $60\%$ the red path and $25\%$ the green path.

\begin{figure}[t]  
	\begin{center}
		\includegraphics[width=0.7\textwidth]{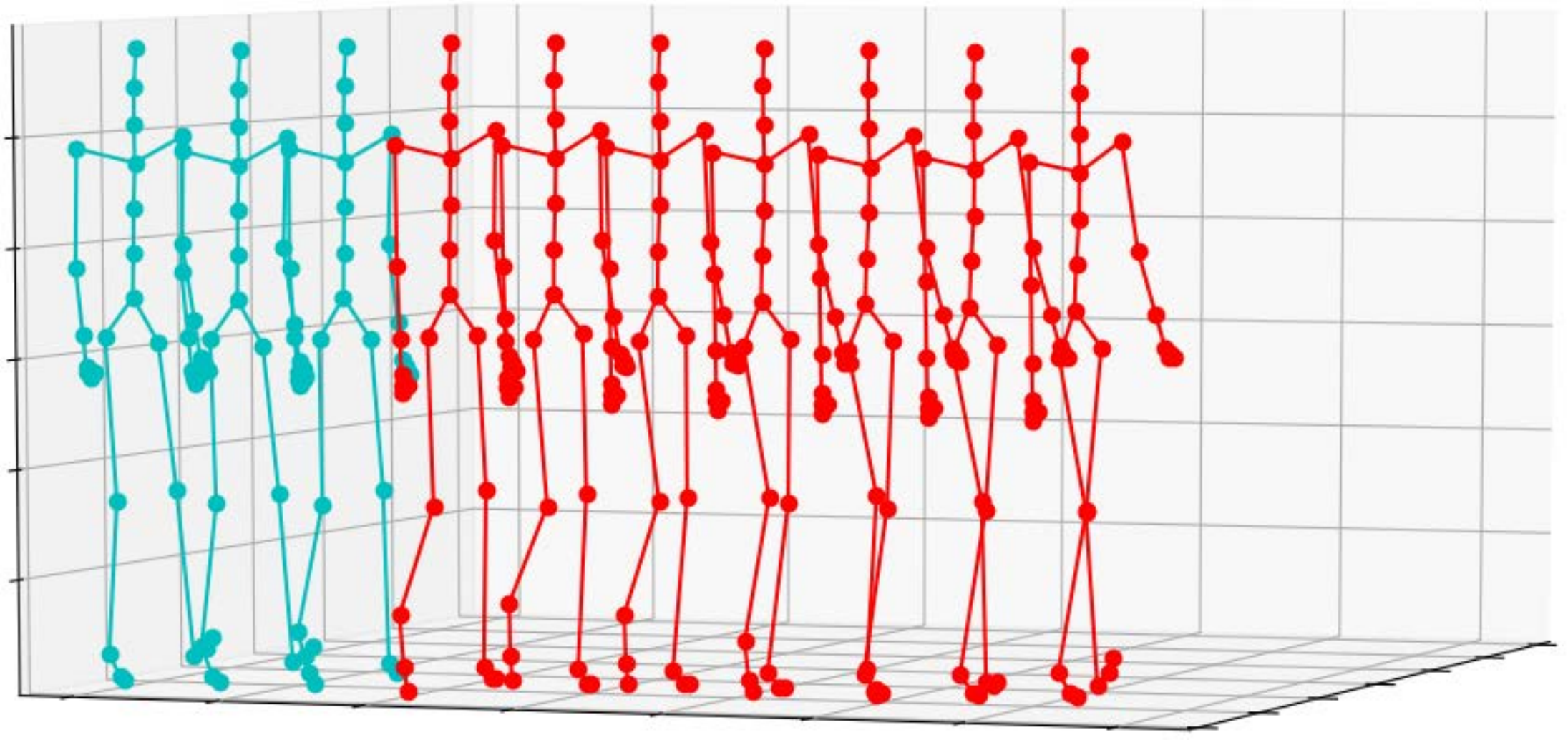}
	\end{center}
	\caption{Final $3$ steps of observed motion (cyan) and first $7$ steps of predicted motion (red).}
	\label{fig:example_mocap_pred}
\end{figure}

For the human motion modeling task, the last $3$ steps of the observation (cyan) and the mean values for the first $7$ steps of the predicted sequence (red) are illustrated as skeletal motion in figure \ref{fig:example_mocap_pred}.
As this task is concerned with walking motion, most movement appears at the legs and feet of the skeleton.
It can be seen that at the end of the observation sequence, the right foot starts rising and throughout the prediction sequence finishes one step forward, thus correctly approximating the walking motion.

In the following sections, the system properties indicated by toy examples \ref{sss:approx_uni} (smoothing in unimodal processes), \ref{ss:te_high_k} (superfluous components) and \ref{sss:smc} (comparison to SMC inference) are verified using real world data in an inference setting, rather than a representation task.

\subsubsection{Smoothing.}
\begin{figure}[htb]
	\begin{center}		
		\includegraphics[width=0.325\textwidth]{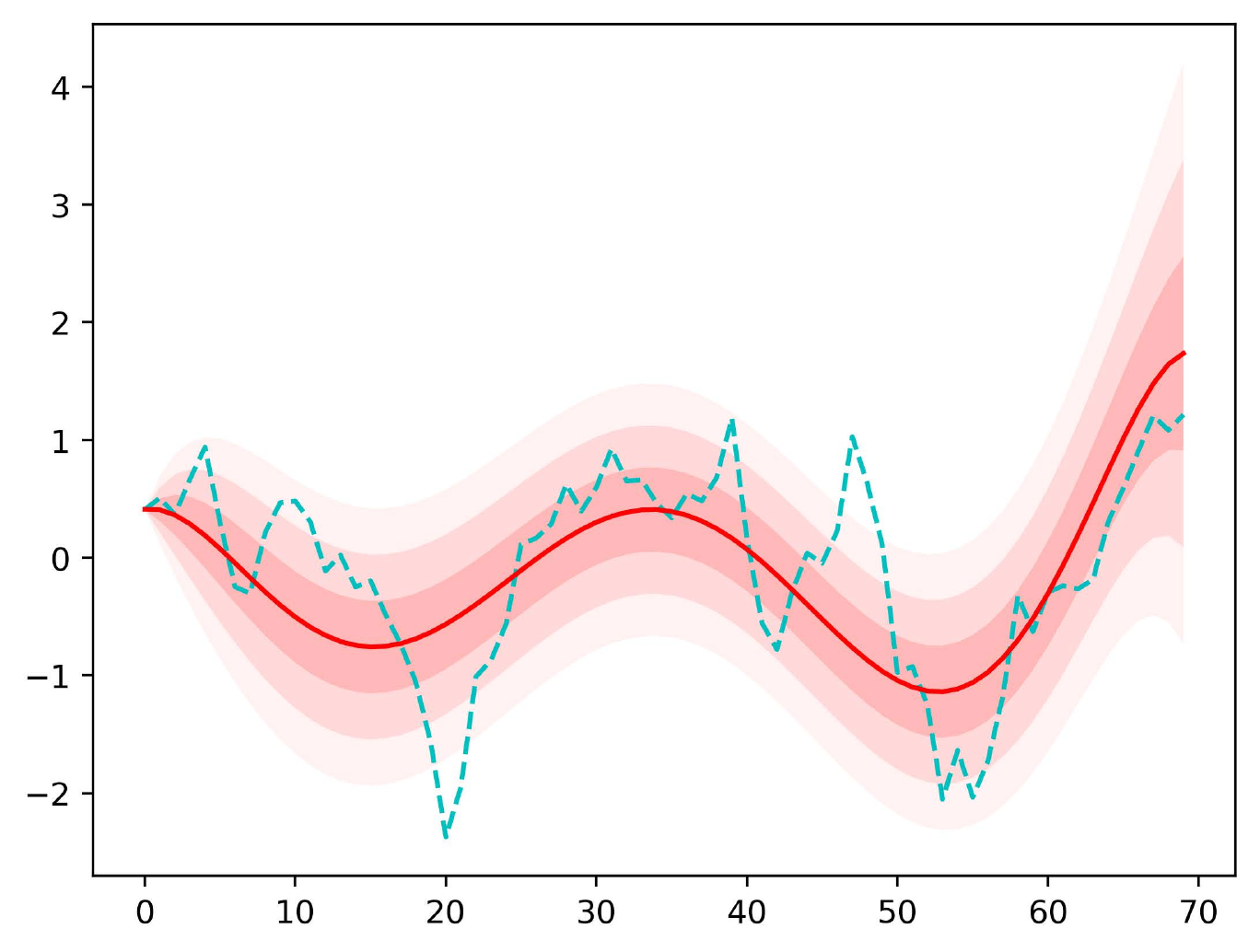}
		\includegraphics[width=0.325\textwidth]{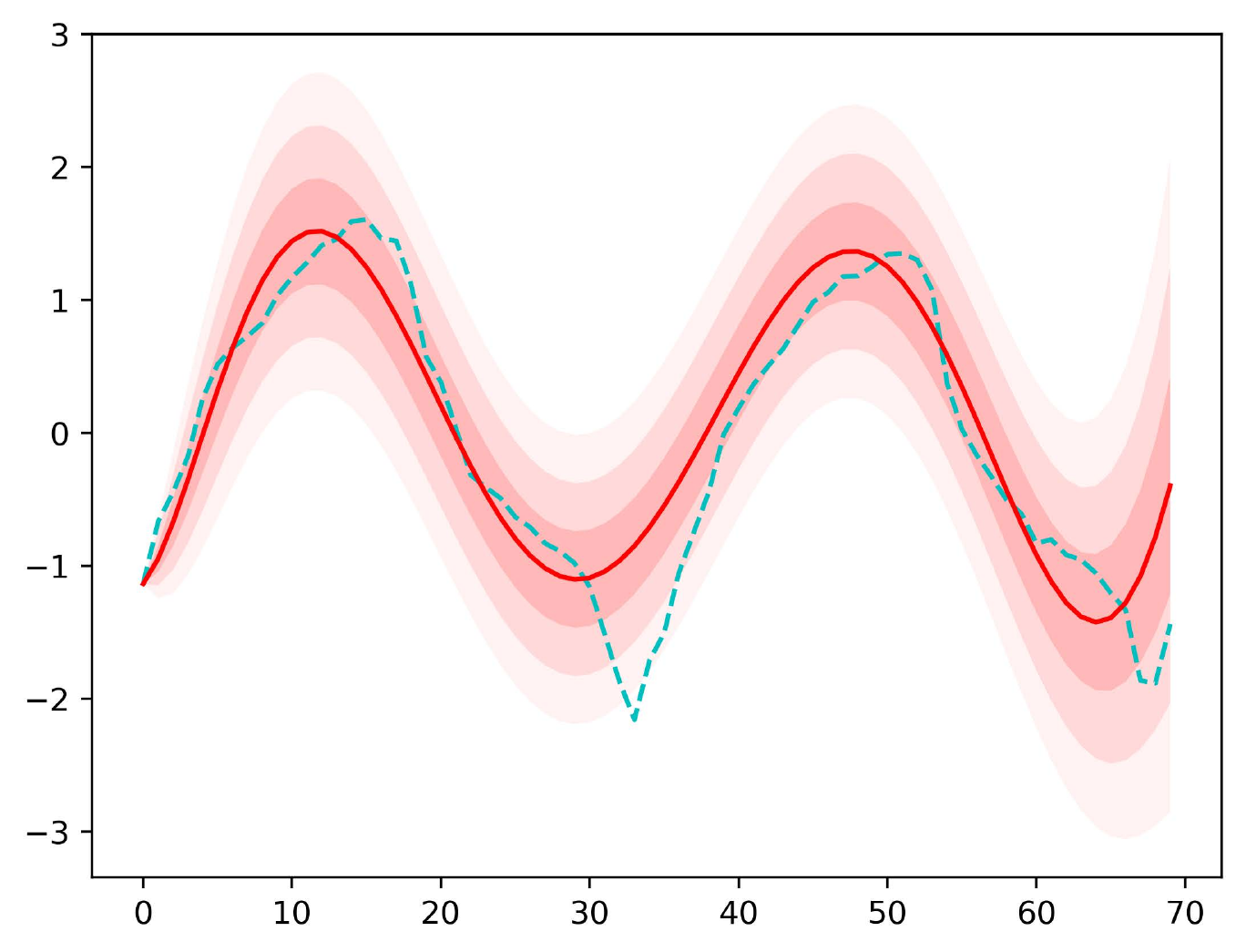}
		\includegraphics[width=0.325\textwidth]{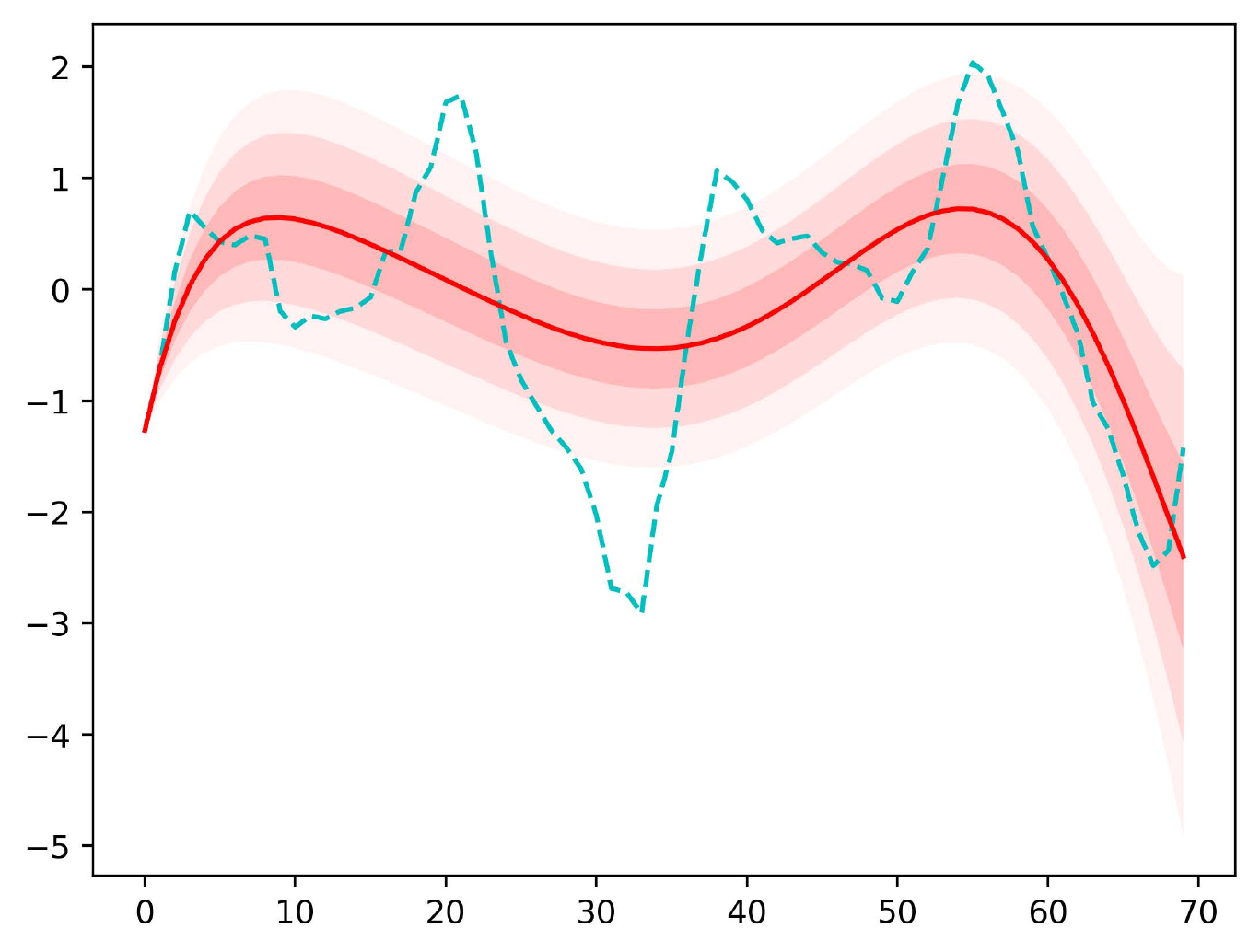}
	\end{center}
	\caption{$\mathcal{N}$-Curve approximation (mean and standard deviations) of different channels (left to right: left hand, right tibia and right toes) in human motion prediction.}
	\label{fig:eval_smoothing}
\end{figure}

The smoothing property of the $\mathcal{N}$-Curve model apparent when looking at the representation of single channels in the case of human motion modeling.
Figure \ref{fig:eval_smoothing} shows the $70$ step sequences for the \emph{left hand}, \emph{right tibia} and \emph{right toes}.
It is clearly visible, that the model, having only $10$ control points to work with, learns a smooth mean sequence to represent the entire sequence and tries to cope with the noise in the sequence by varying the variance of the control points.
For some channels, e.g. the \emph{right toes} channel, using more control points might be suited to also capture the stronger peak downwards in the middle of the sequence.

\subsubsection{Superfluous components.}
\begin{figure}[htb]
	\begin{center}		
		\includegraphics[width=0.325\textwidth]{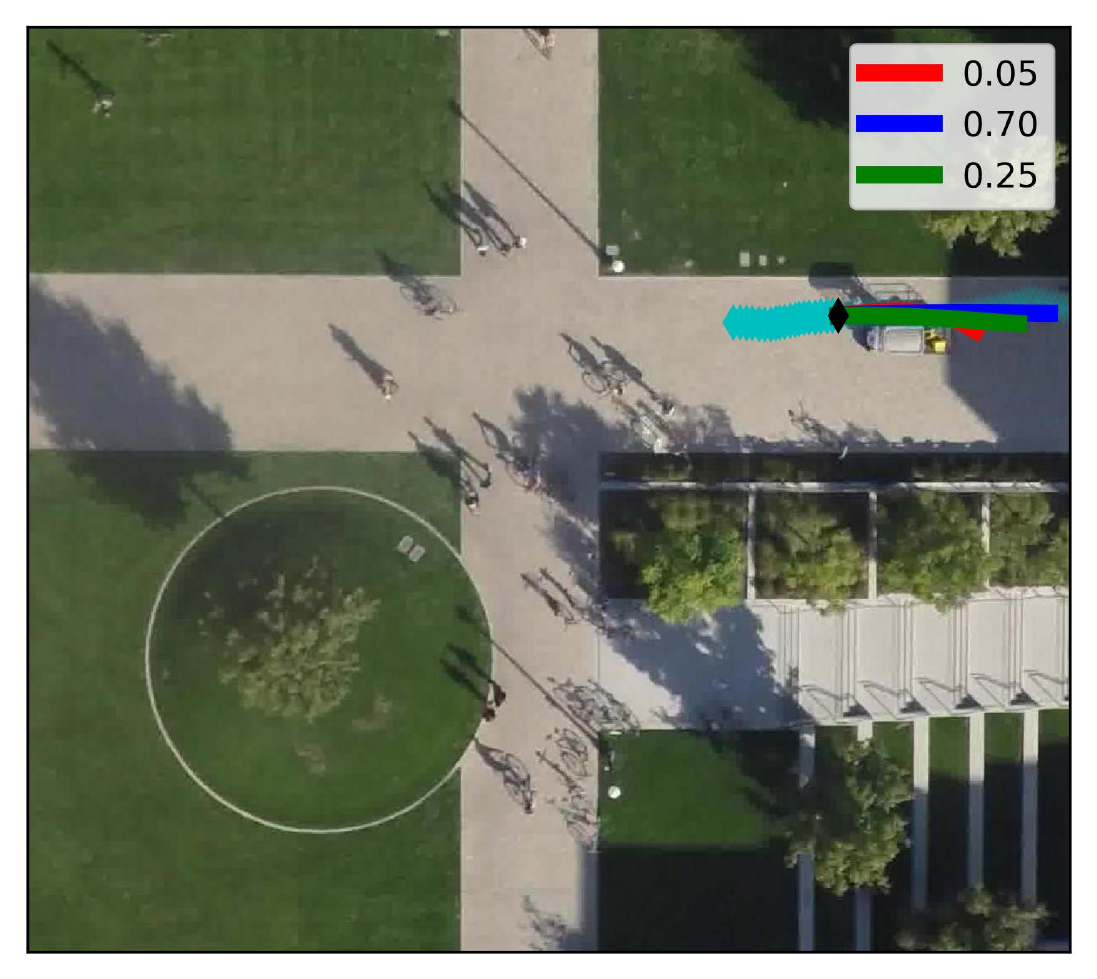}
		\includegraphics[width=0.325\textwidth]{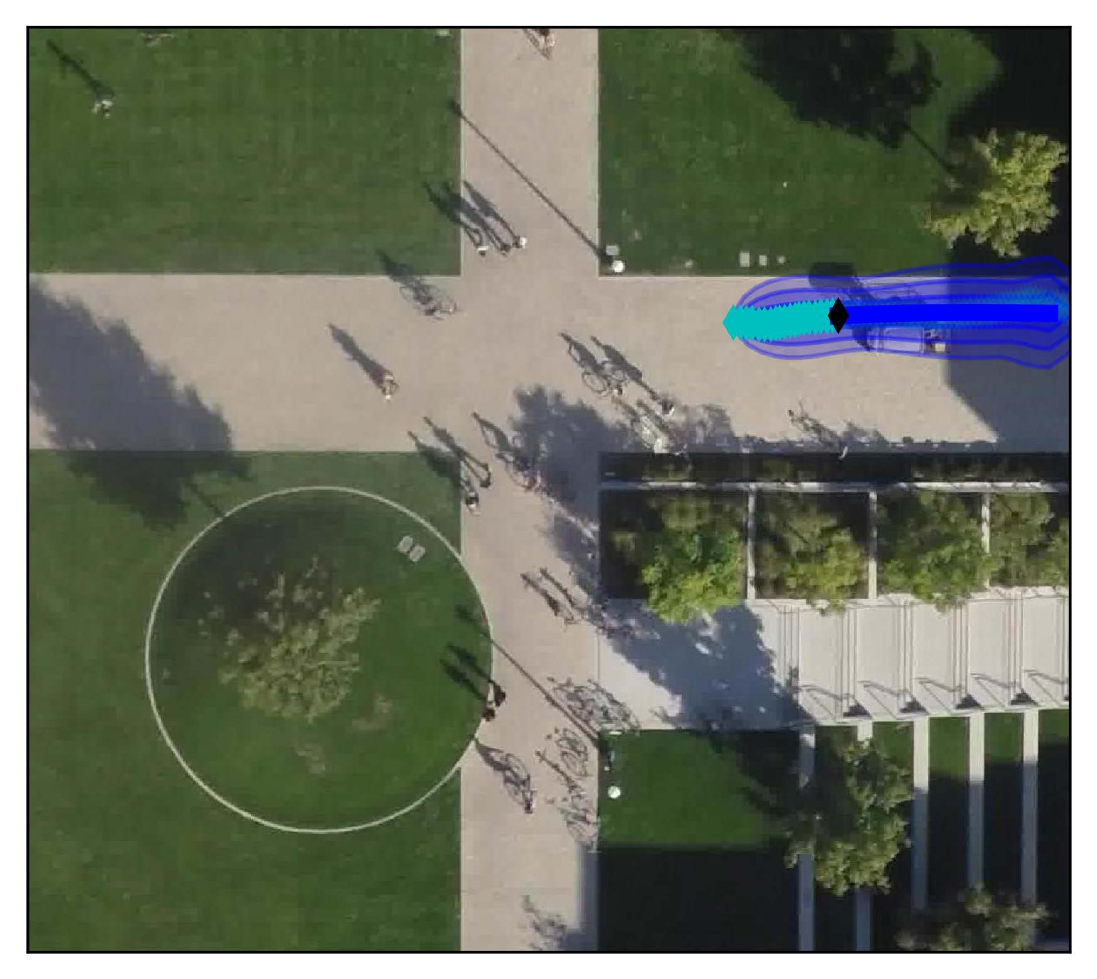}
		\includegraphics[width=0.325\textwidth]{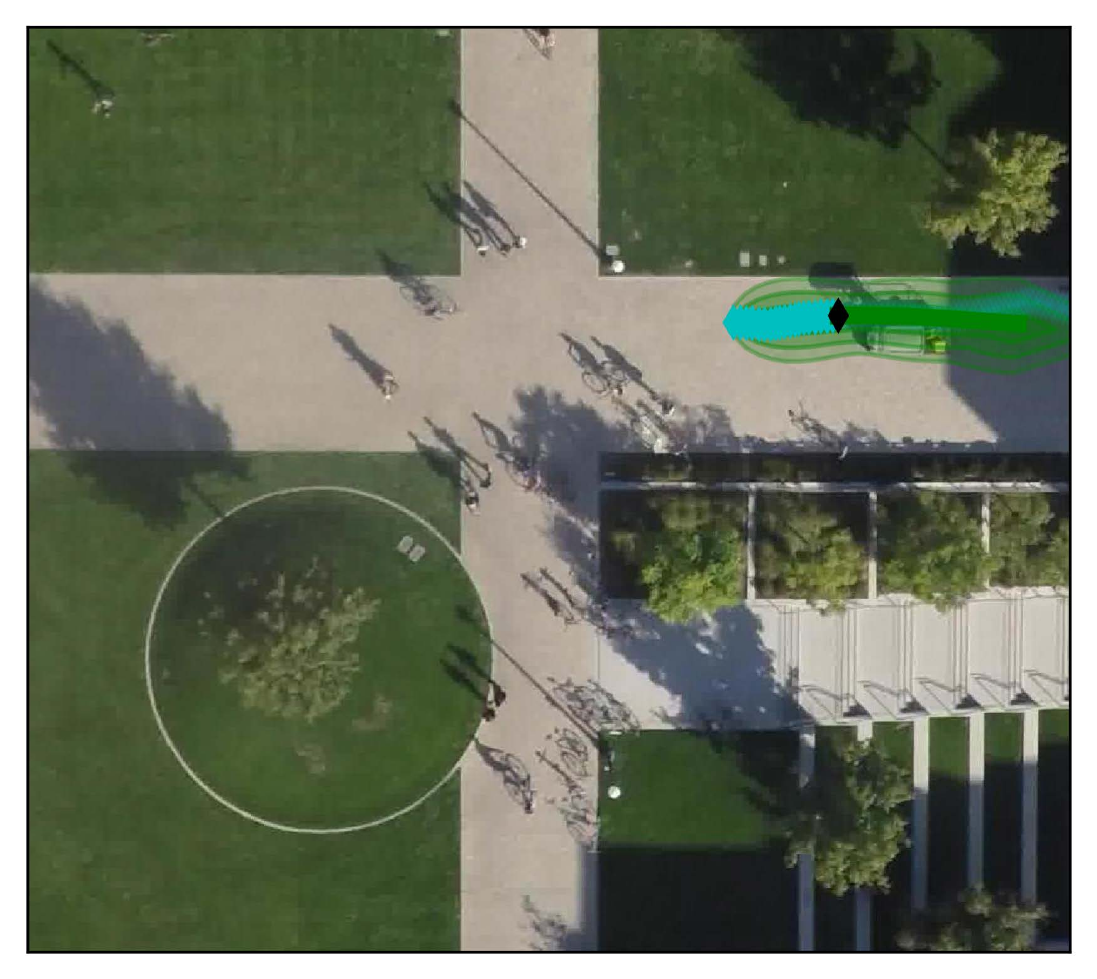}
	\end{center}
	\caption{Left: $\mathcal{N}$-Curve mixture model trajectory prediction example with superfluous components.
			Center and right: Similar components with variance.}
	\label{fig:eval_high_k}
\end{figure}

The toy example in \ref{ss:te_high_k} indicates that the model is incapable of completely suppressing superfluous components, but instead overlaps several nearly equal components in order to approximate single modes.
This behavior can be confirmed looking at the example depicted in figure \ref{fig:eval_high_k}.
While the red component is driven towards zero, the blue and green components are quite similar, only differing in the modeled movement speed slightly as indicated by the length of the predicted trajectory.
As before in the toy example, the variances of both components are similar to each other, thus both components might again be collapsed into one mode in a post-processing step in order to remove superfluous components when necessary.

\subsubsection{Predicted distribution comparison with an SMC approach.}
\begin{figure}[htb]
	\begin{center}		
		\includegraphics[width=0.45\textwidth]{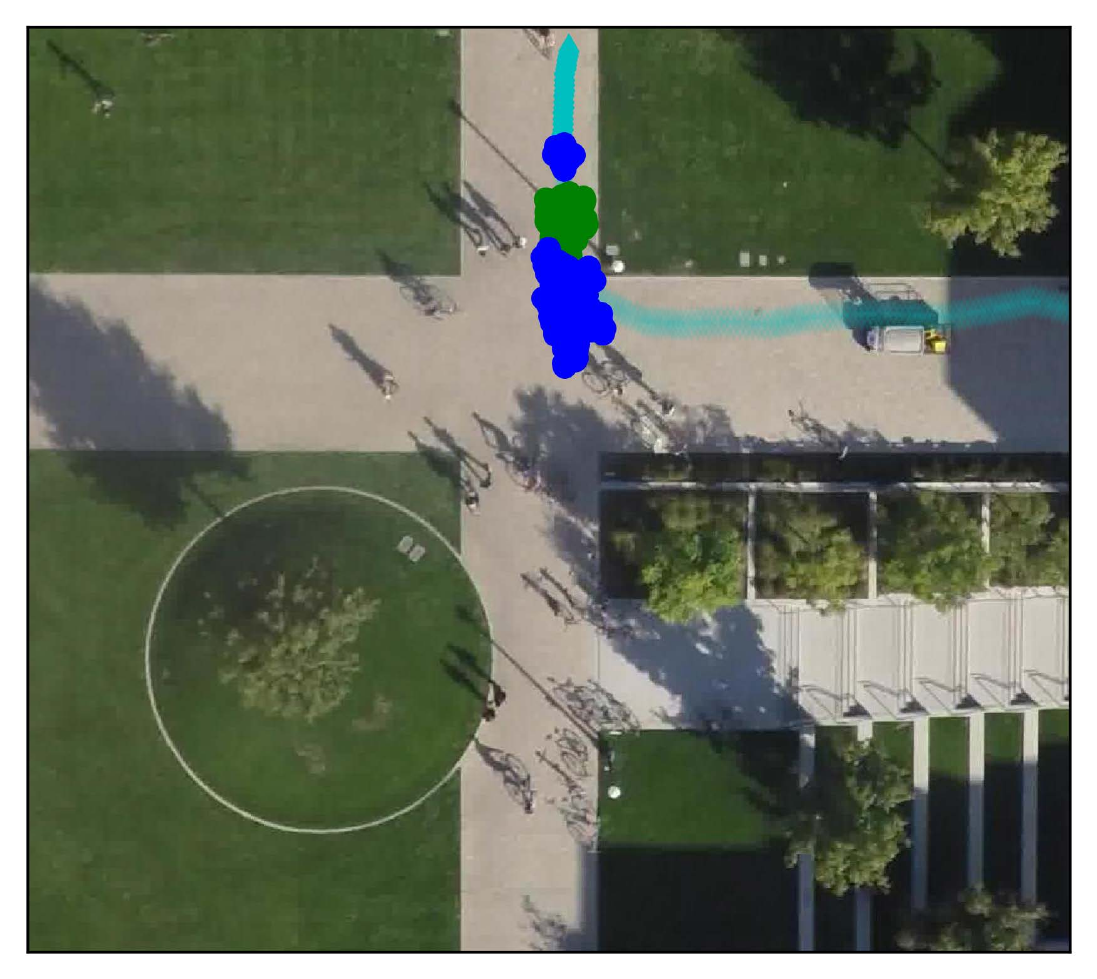}
		\includegraphics[width=0.45\textwidth]{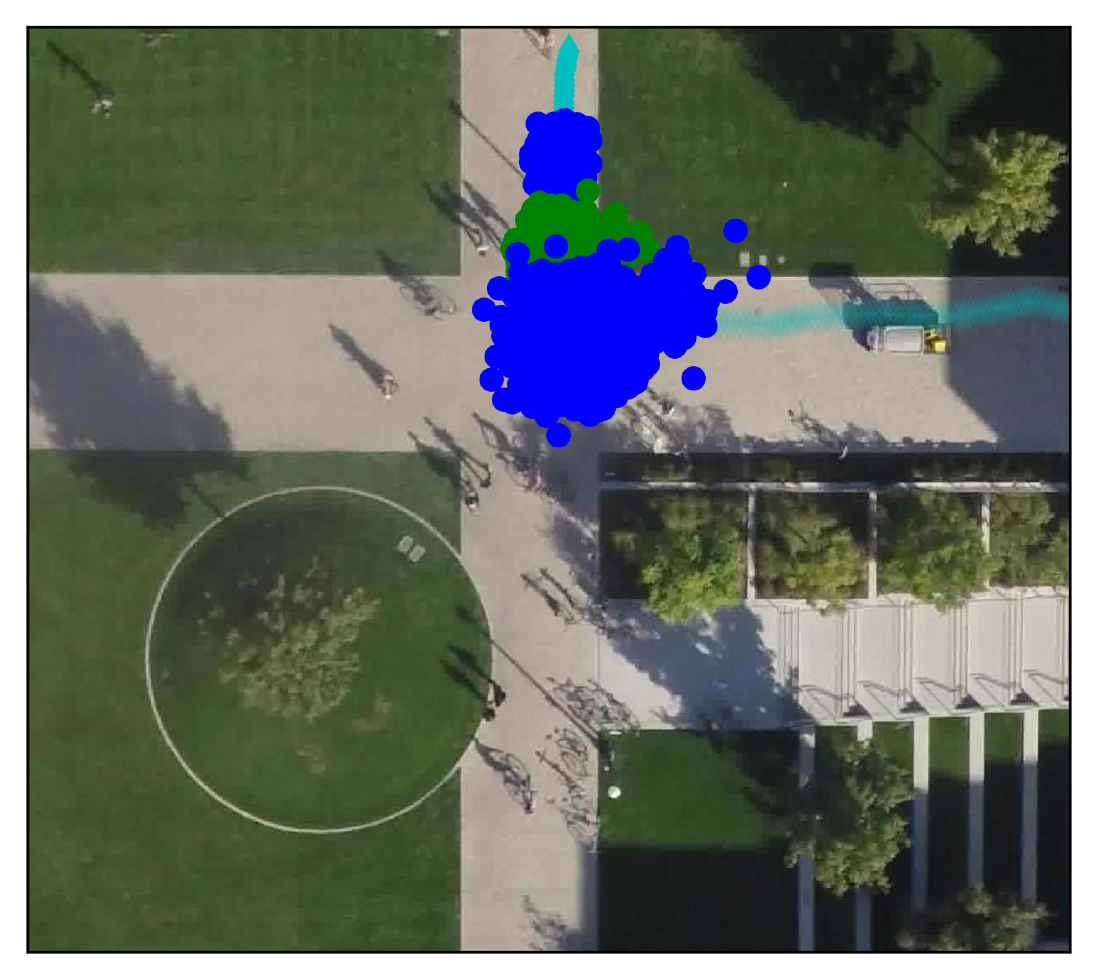}
	\end{center}
	\caption{Sample representation of Gaussian mixture distributions for trajectory points after $1$, $20$ and $40$ steps (alternating blue and green) generated by the Particle LSTM-MDN model (left) and the $\mathcal{N}$-Curve mixture model (right).}
	\label{fig:eval_smc}
\end{figure}

Finally, the findings of the toy example in \ref{sss:smc} concerned with comparing resulting distributions generated by an $\mathcal{N}$-Curve model and a sequential monte carlo (SMC) approach are revised.
The toy example showed that, in essence, both approaches capture changes in variance.
In that, the $\mathcal{N}$-Curve model tends to over-estimate uniformly distributed data, while the SMC approach tends to under-estimate the data after a few simulation steps, due to the use of Gaussian distributions during simulation.
The SMC approach, here the Particle LSTM-MDN model, generating predictions with smaller variance is already indicated by larger \emph{NLL} values during quantitative evaluation in \ref{subsec:traj_pred}.
This behavior can also be seen in the left image of figure \ref{fig:eval_smc}.
Here, simulated particles indicate two possible movement directions: straight and towards the right side of the scene.
Due to narrower predictions, the possibility of movement to the left is possibly omitted, as there is a bias in the data, that pedestrians moving closer to the right side of the pathway are more likely to move straight or turn to the right and not to the left.
As the $\mathcal{N}$-Curve mixture model produces higher variance output, the possibility of moving towards the left side of the scene becomes relevant.
For the sample representation of the distributions depicted in figure \ref{fig:eval_smc}, the $\mathcal{N}$-Curve mixture model component heading to the left side is left out for better comparison of the two approaches.
Again, the findings of the toy example are confirmed in the setting of trajectory prediction on real world data.

\section{Conclusions and Future Works}
\label{sec:conclusions}
In this paper, the $\mathcal{N}$-Curve mixture model, an approach for learning the model of a continuous-time stochastic processes defined by Gaussian mixture distributions, has been presented.
The approach is based on Bézier curves with Gaussian control points, thus a respective stochastic process is represented by a mixture of parametric, probabilistic curves, termed $\mathcal{N}$-Curves.
By using parametric curves and optimizing in function space rather than the $d$-dimensional space of sequence values, the proposed model is able to generate smooth continuous predictions in a single inference step.
Using the presented training approach, the $\mathcal{N}$-Curve mixture model can be learned from discrete sequence data.
Initial experiments show that the presented model is viable for $n$-step sequence prediction and achieves state-of-the-art performance on the tasks of trajectory prediction and human motion modeling.
Further, several toy examples and a qualitative evaluation show, that the proclaimed properties of the proposed approach hold true.

Future work mainly focuses on developing the $\mathcal{N}$-Curve mixture model into a recurrent system, in order to process sequences of variable length (i.e. allow index sets like $t \in \mathbb{R}^+_0$).
Further, the possibilities of incorporating regularizations and correlations between control points should be explored, to allow, for example, biased curve sampling for data generation.
Lastly, the training approach could be adapted to prevent generation of multiple similar predictions with non-zero weights, thus putting higher emphasize on the suppression of superfluous mixture components.

%
%
%
\bibliographystyle{splncs04}  
\bibliography{bibliography}

\end{document}